\documentclass[12pt, draftclsnofoot, onecolumn]{IEEEtran}
\usepackage{amsmath,amsfonts}
\usepackage{array}
\usepackage{textcomp}
\usepackage{xcolor}
\usepackage{stfloats}
\usepackage{url}
\usepackage{verbatim}
\usepackage{graphicx}
\usepackage{cite}
\usepackage{makecell}
\usepackage[ruled,linesnumbered]{algorithm2e}
\usepackage{setspace}
\usepackage{subfigure}
\usepackage{caption}
\begin{document}

\title{FedAQ: Communication-Efficient Federated Edge Learning via Joint Uplink and Downlink Adaptive Quantization}

\author{Linping Qu,~\IEEEmembership{Graduate Student Member,~IEEE,} Shenghui Song,~\IEEEmembership{Senior Member,~IEEE,} and Chi-Ying Tsui,~\IEEEmembership{Senior Member,~IEEE}
\thanks{The work described in this paper was supported by a grant from the Research Grants Council of the Hong Kong Special Administrative Region, China (Project Reference Number: AoE/E-601/22-R). Part of this work was presented at the IEEE Global Communications Conference (GLOBECOM), Rio de Janeiro, Brazil, Dec. 2022 \cite{qu2022feddq}. Linping Qu, Shenghui Song, and Chi-Ying Tsui are with the Department of Electronic and Computer Engineering, Hong Kong University of Science and Technology, Hong Kong (e-mail: lqu@connect.ust.hk; eeshsong@ust.hk; eetsui@ust.hk). }
}

\maketitle

\begin{abstract}
Federated learning (FL) is a powerful machine learning paradigm which leverages the data as well as the computational resources of clients, while protecting clients' data privacy. However, the substantial model size and frequent aggregation between the server and clients result in significant communication overhead, making it challenging to deploy FL in resource-limited wireless networks. In this work, we aim to mitigate the communication overhead by using quantization. Previous research on quantization has primarily focused  on the uplink communication, employing either fixed-bit quantization or adaptive quantization methods. In this work, we introduce a holistic approach by joint uplink and downlink adaptive quantization to reduce the communication overhead. In particular, we optimize the learning convergence by determining the optimal uplink and downlink quantization bit-length, with a communication energy constraint. Theoretical analysis shows that the optimal quantization levels depend on the range of model gradients or weights. Based on this insight, we propose a decreasing-trend quantization for the uplink and an increasing-trend quantization for the downlink, which aligns with the change of the model parameters during the training process. Experimental results show that, the proposed joint uplink and downlink adaptive quantization strategy can save up to 66.7\% energy compared with the existing schemes.
\end{abstract}


\section{Introduction}
\label{Introduction}
\subsection{Background}
Federated learning (FL) \cite{konevcny2015federated,yang2019federated} is an emerging technique to tackle the data isolation problem by \textcolor{black}{distributed learning technology}. Instead of sharing \textcolor{black}{the data of different clients}, FL \textcolor{black}{only shares model parameters such that the clients' privacy can be protected}\cite{li2020federated}.
However, FL faces a significant challenge in terms of communication overhead\cite{zhang2021fedpd,zhang2016parallel,konevcny2016federated,cui2018mqgrad}, \textcolor{black}{which} arises from three aspects\cite{kairouz2021advances,konevcny2016federated}. Firstly, numerous communication rounds between the clients and the parameter server \textcolor{black}{may be required to achieve a satisfactory performance}. Secondly, each communication round involves a potentially large number of participating clients, which can range from thousands to millions. Lastly, the \textcolor{black}{size of} deep neural network (DNN) model \textcolor{black}{keeps increasing, e.g.,} from megabytes to gigabyte size. Consequently, the clients need to transmit and receive a substantial amount of data to complete an FL task, resulting in high energy consumption and long latency. This poses a challenge for resource-limited \textcolor{black}{wireless networks and clients, calling for a communication-efficient design}.

\subsection{Related Works}
Several methods have been proposed in the literature to tackle the communication overhead \textcolor{black}{issue} in FL. These methods can be categorized into three \textcolor{black}{lines of research}: infrequent communication, \textcolor{black}{model sparsification}, and quantization. 

\textit{Infrequent Communication:} One approach to reduce communication overhead is to accelerate training and reduce the number of 
communication rounds between clients and the server. Local SGD and FedAvg algorithms, as proposed in \cite{mcmahan2017communication} and \cite{dean2012large}, perform multiple steps of local updating before \textcolor{black}{model aggregation}, reducing the \textcolor{black}{number of} communication rounds at the expense of local computation. \textcolor{black}{The authors of \cite{zhang2021fedpd} and  \cite{mills2019communication} considered the communication design for the scenario of non-identically independent distribution (I.I.D).} The number of communication round is further reduced by one-shot training \cite{guha2019one}. Theoretical analyses of periodic averaging \textcolor{black}{were given} in \cite{stich2018local,wang2021cooperative,yu2019parallel,haddadpour2019local,khaled2020tighter}, while \cite{zhang2016parallel} and \cite{wang2019adaptive} \textcolor{black}{analyzed} the impact of communication frequency on convergence rate and \textcolor{black}{proposed} to adaptively modify the frequency during training, aiming to save communication overhead and keep a low error floor. Additionally, \cite{reisizadeh2020fedpaq} and \cite{konevcny2016federated} \textcolor{black}{combined} infrequent communication with other strategies such as \textcolor{black}{model sparsification} and quantization to further save communication.

\textcolor{black}{\textit{Model sparsification:} Model sparsification techniques aim to reduce the number of DNN model parameters}. These techniques involve selecting the most significant gradients based on a predefined percentage or threshold\cite{aji2017sparse,lindeep,wang2018atomo,ji2021dynamic,yan2020dual}. Other techniques like low rank\cite{konevcny2016federated,vogels2019powersgd} and the lottery ticket hypothesis\cite{li2021lotteryfl} \textcolor{black}{were} adopted in FL to reduce the dimensions of gradients. \textcolor{black}{Aggressive} reduction on parameters can be achieved by these methods, but they often require empirical knowledge and manual efforts. Flexible gradient compression techniques have also been proposed \textcolor{black}{where the sparsity rate is determined automatically}\cite{lu2018multi,phuong2020distributed}.
Metrics based on gradient variance \cite{tsuzuku2018variance}, standard deviation \cite{chen2020standard}, or entropy \cite{xiao2021egc} \textcolor{black}{were heuristically} used to determine the sparsity. Theoretical convergence analyses for model sparsification have also been studied in \cite{wangni2018gradient,alistarh2018convergence,stich2018sparsified}.

\textit{Quantization:} Quantization is a lossy compression method that is orthogonal to infrequent communication and model sparsification. It aims to reduce the bit length of transmitted parameters \cite{shlezinger2020federated,alistarh2017qsgd}. Shorter bit-length quantization can significantly reduce communication overhead \textcolor{black}{at the cost of communication} errors that affect learning accuracy. Finding a suitable bit-length \textcolor{black}{with} a good trade-off between the communication overhead and learning accuracy is crucial in quantized FL \cite{oland2015reducing,cui2018mqgrad}.

\textcolor{black}{The authors of \cite{alistarh2017qsgd} proposed} a fixed 8-bit quantization on gradients. \cite{wen2017terngrad} and \cite{xu2020ternary} \textcolor{black}{used} ternary values (+1, 0, -1) to represent all the gradients while \cite{seide20141,strom2015scalable,bernstein2018signsgd} \textcolor{black}{proposed} 1-bit quantization. Besides the pursuit of shorter bit-length, other approaches \textcolor{black}{used} different types of quantizers to replace the classical stochastic uniform quantizer. Vector quantization \cite{yu2018gradiveq,shlezinger2020uveqfed,gandikota2021vqsgd} and clustering-based quantization \cite{cui2020clustergrad} \textcolor{black}{were} proposed to better fit the characteristics of FL. 
To mitigate quantization errors, methods like random rotation \cite{konevcny2016federated}, \cite{suresh2017distributed}, error feedback \cite{wu2018error}, \cite{karimireddy2019error}, and lazily aggregated gradients \cite{sun2019communication} have been proposed. 
Theoretical analyses of convergence rate for quantized FL \textcolor{black}{were} investigated in \cite{reisizadeh2020fedpaq},\cite{tang2018communication} and \cite{jiang2018linear}.
However, the optimum bit-length may vary for different models and tasks, \textcolor{black}{and thus requires} significant manual effort to determine the best bit-length. Additionally, fixed-bit quantization may not adapt well to the dynamic training process. To address these issues, adaptive quantization has gained \textcolor{black}{much} attention.

Existing adaptive quantization schemes in FL mainly focused on the uplink communication and used an increasing-trend quantization approach, where a small bit length is used in the early training stages, while a longer bit length is employed in the late training stages \cite{oland2015reducing,cui2018mqgrad,guo2020accelerating,jhunjhunwala2021adaptive}. Criteria for changing the bit-length have been proposed. For example, \cite{oland2015reducing} \textcolor{black}{used} the root mean square (RMS) value of gradients to evaluate the impact of quantization on training loss. If the RMS value is large, a \textcolor{black}{large number of} quantization bit will be assigned, and vice versa. \textcolor{black}{Reinforcement learning was proposed in \cite{cui2018mqgrad} to determine} the quantization bit \textcolor{black}{based on training loss}. 
Other approaches, such as \cite{jhunjhunwala2021adaptive} and \cite{guo2020accelerating}, \textcolor{black}{utilized} training loss or the mean to standard deviation ratio (MSDR) of gradients as the criteria \textcolor{black}{for selecting the quantization bits}. 
However, they mainly focused on the uplink transmission \textcolor{black}{without theoretical support}. Downlink quantization also holds significance for energy-sensitive clients. Reducing the bit length of model parameters in the downlink not only saves receiving energy but also decreases channel decoding energy.

\subsection{Contributions and Organization}
In this work, we propose a novel approach to decrease energy consumption \textcolor{black}{and communication latency} by joint uplink and downlink adaptive quantization. 

To determine the dynamic quantization bit-length in both the uplink and downlink, \textcolor{black}{we formulate an optimization problem}. \textcolor{black}{For a given energy constraint, we optimize the learning convergence by allocating the bit-length in the uplink and downlink during various training stages.
The solution of the optimization problem reveals that the optimal quantization bit exhibits a positive correlation with the range of the model weights or gradients. }

The main contributions of this work are outlined as follows:
\begin{itemize}
    \item Joint Uplink and Downlink \textcolor{black}{Quantization}: Unlike existing studies that solely focus on a single link, this work emphasizes the joint consideration of both the uplink and downlink to achieve more substantial reduction in energy consumption.
    \item Rigorous Theoretical Analysis: The design of our approach is rooted in rigorous theoretical analysis, distinguishing it from  intuitive or heuristic approach. \textcolor{black}{The numerical range of model updates and model weights are found to be critical in making quantization decisions.}
    \item Based on the theoretical analyses and the characteristics of FL, decreasing-trend quantization in the uplink and increasing-trend quantization in the downlink is motivated, whose superiority is demonstrated by extensive experiments.
\end{itemize}

The remainder of the paper is organized as follows. In Section \ref{System Model}, the system model is described. In Section \ref{Theoretical Analysis and Design}, comprehensive theoretical analysis of joint uplink and downlink quantized FL is provided. The formulation of the optimization problem is discussed and the algorithm based on the theoretical analysis is then described. In Section \ref{Special Cases Analysis}, the theoretical analysis and the corresponding optimization solutions for two special cases: uplink-only quantization and downlink-only quantization, are given. In Section \ref{Expirement}, experimental results are povided and discussed. Finally, the conclusion is drawn in Section \ref{Conclusion}.

\textcolor{black}{\textbf{Notations}. The notations used in this article are listed in Table \ref{tab:notations}.}
\section{System Model}
\label{System Model}
In this section, we will introduce the concerned FL system with joint uplink and downlink quantization.
\begin{table}[t]\color{black}
\small
\caption{\textcolor{black}{Key Notations.}}
\centering
\begin{tabular}{|c|c|c|c|}
\hline
{\textbf{Notation}}&{\textbf{Definition}}&{\textbf{Notation}}&{\textbf{Definition}}\\
\hline
{$\mathbf{w}$}&{DNN model.}&{$\Delta \mathbf{w}$}&{Local model update.}\\
\hline
{$f$}&{Loss function of training.}&{${\nabla} f$}&{The gradient.}\\
\hline
{$m$}&{Index of the communication round.}&{$K$}&{Total communication rounds.}\\
\hline
{$t$}&{Index of local training step.}&{$\tau$}&{Local training steps.}\\
\hline
{$i$}&{Index of the mobile device.}&{$n$}&{The number of total mobile devices.}\\
\hline
{$Q()$}&{Quantization operator.}&{$s$}&{The number of quantization bins.}\\
\hline
{$L$}&{The L-smooth parameter.}&{$\eta$}&{Learning rate.}\\
\hline
{$d$}&{Dimensions of the DNN model.}&{$R$}&{Range of model.}\\
\hline
{$\sigma^2$}&{The variance of estimated local gradients.}&{$p_i$}&{The data size ratio of the $i$-th client.}\\
\hline
\end{tabular}
\label{tab:notations}
\end{table}
\subsection{\textcolor{black}{Federated Learning}}
The goal of FL is to train a global model by utilizing the data residing on different clients. The problem can be formulated as
\begin{align}
{\underset{\mathbf{w}\in {\mathbb{R}^d}}{min}\ f(\mathbf{w}) = \sum\limits^n_{i=1}p_if_i(\mathbf{w})},
\end{align}
where $n$ denotes the number of clients, $p_i$ is the ratio between the data size of the $i^{th}$ client to that of all  clients, and $f_i(\mathbf{w})$ \textcolor{black}{represents} the training loss \textcolor{black}{of} the $i^{th}$ \textcolor{black}{client} on its local datasets. In the $m^{th}$ round of communication, the server broadcasts the global model $\mathbf{w}_m$ to all the selected clients. After receiving the global model, each client performs $\tau$ steps of local training by stochastic gradient descent (SGD) algorithm with a learning rate of $\eta$. The local updating rule is given by
\begin{align}
{\mathbf{w}^i_{m,t+1}=\mathbf{w}^i_{m,t}-\eta\tilde{\nabla} f_i(\mathbf{w}^i_{m,t})},
\end{align}
where $t=0,...,\tau-1$, $\tilde{\nabla} f_i(\mathbf{w}^i_{m,t})$ denotes the stochastic gradient computed from local datasets, and $\mathbf{w}^i_{m,0}=\mathbf{w}_m$. 

After $\tau$ steps of local training, the $i^{th}$ client obtains a new model $\mathbf{w}^i_{m,\tau}$, \textcolor{black}{and} the local model update is \textcolor{black}{given} by 
\begin{align}
\Delta \mathbf{w}^i_m=\mathbf{w}^i_{m,\tau}-\mathbf{w}_m.
\end{align} 
Then, each client uploads the local model update to the server, and the server will update the global model by
\begin{align}
\mathbf{w}_{m+1}=\mathbf{w}_{m}+\sum\limits_{i=1}^np_i\Delta \mathbf{w}^i_m.
\end{align}
The new global model will be broadcast to selected clients for \textcolor{black}{next} round of communication. 
\subsection{Quantization Model}
\begin{figure}[t]
    \begin{center}
    \includegraphics[width=0.5\linewidth]{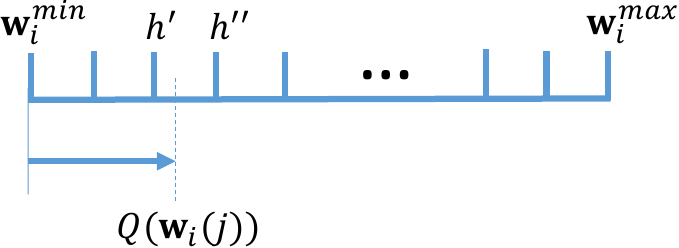}
    \end{center}
    \caption{Illustration of quantization.}
    \label{fig:quant}
\end{figure}

In practice, both the model update $\Delta \mathbf{w}^i_m$ and global model $\mathbf{w}_m$ will be quantized during the uplink and downlink communication. In this paper, we adopt a widely utilized stochastic uniform quantizer\cite{konevcny2016federated}\cite{suresh2017distributed}. In particular, for a given model $\mathbf{w}_i \in {\mathbb{R}^d}$, we first compute its range, i.e., $\mathbf{w}_i^{max}- \mathbf{w}_i^{min}$, where $\mathbf{w}_i^{max}=max_{1\leq j\leq d}\mathbf{w}_i(j)$ and $\mathbf{w}_i^{min}=min_{1\leq j\leq d}\mathbf{w}_i(j)$ denote the maximum and minimum value of the model elements, respectively.
With $N$ bit quantization, the range is divided into $2^N-1$ bins. As shown in Figure~\ref{fig:quant}, if $\mathbf{w}_i(j)$ is located in one bin whose lower bound and upper bound are $h'$ and $h''$, respectively, then the quantized value will be calculated as
\begin{align}
Q(\mathbf{w}_i(j))=
\begin{cases}
h', & \mbox{with probability  }\frac{h''-\mathbf{w}_i(j)}{h''-h'}, \\
h'', & \mbox{otherwise}.
\end{cases}
\end{align}

\subsection{FL with joint uplink and downlink quantization}
Let $Q_1(\cdot)$ and $Q_2(\cdot)$ denote the quantization function in the uplink and downlink \textcolor{black}{communication}, respectively. In the $m^{th}$ round of communication, the client will receive a quantized global model $Q_2(\mathbf{w}_m)$ from the server. After $\tau$ steps of local training based on the received quantized model, the client will \textcolor{black}{obtain the} local model update
\begin{align}
\Delta \mathbf{w}^i_m=\mathbf{w}^i_{m,\tau}-Q_2(\mathbf{w}_m),
\end{align}
\textcolor{black}{which} will be quantized by
\begin{align}
Q_1(\Delta \mathbf{w}^i_m)=Q_1(\mathbf{w}^i_{m,\tau}-Q_2(\mathbf{w}_m)). 
\end{align}
Finally, the server will \textcolor{black}{update the global model} by
\begin{align}
\mathbf{w}_{m+1}=\mathbf{w}_{m}+\sum\limits_{i=1}^np_iQ_1(\Delta \mathbf{w}^i_m).
\end{align}

\section{Theoretical Analysis and Design}
\label{Theoretical Analysis and Design}
In this section we will jointly design the uplink and downlink quantization to save the overall communication overhead. 
For that purpose, we first \textcolor{black}{derive the convergence bound for FL with quantization error in the uplink and downlink. Then, we} formulate the adaptive quantization issue into an optimization problem, whose objective is to minimize the learning convergence bound by tuning the quantization level in the uplink and downlink under an energy consumption constraint. Finally we solve the optimization problem theoretically and propose an algorithm based on the theoretical conclusion.
\subsection{\textcolor{black}{Convergence Analysis}}
In this subsection, we derive the FL convergence bound with quantization. \textcolor{black}{For that purpose, we first} introduce three commonly-used assumptions.\\
\textbf{Assumption 1}\ \textit{(Unbiased random quantizer)} The quantization operation Q($\cdot$) is unbiased and its variance is bounded by the quantization level and range of the parameters\cite{suresh2017distributed,jhunjhunwala2021adaptive}:
\begin{align*}
&\mathbb{E}[Q(\mathbf{w})|\mathbf{w}]=\mathbf{w} \\
&\mathbb{E}[||Q(\mathbf{w})-\mathbf{w}||^2|\mathbf{w}]\leq q_sR^2(\mathbf{w}),
\end{align*}
where $R(\mathbf{w})=\mathbf{w}_{max}-\mathbf{w}_{min}$, $q_s=\frac{d}{s^2}$, and $s$ \textcolor{black}{denotes the total number of} quantization bins.\\
\textbf{Assumption 2}\ \textit{(L-smoothness)} The loss function $f$ is L-smooth with respect to $\mathbf{w}$. For any $\mathbf{w}$, $\hat{\mathbf{w}} \in \mathbb{R}^d$, we have \cite{jhunjhunwala2021adaptive, reisizadeh2020fedpaq}
\begin{align*}
||\nabla f(\mathbf{w})-\nabla f(\hat{\mathbf{w}})||\leq L||\mathbf{w}-\hat{\mathbf{w}}||.  
\end{align*}
\textbf{Assumption 3}\ \textit{(Variance of SGD)} \textcolor{black}{Locally estimated} stochastic gradient $\tilde{\nabla} f(\mathbf{w})$ is unbiased and \textcolor{black}{with bounded} variance:
\begin{align*}
&\mathbb{E}[\tilde{\nabla} f(\mathbf{w})|\mathbf{w}]=\nabla f(\mathbf{w}) \\
&\mathbb{E}[||\tilde{\nabla} f(\mathbf{w})-\nabla f(\mathbf{w})||^2] \leq \sigma^2.
\end{align*}
With Assumptions 1--3, the convergence bound of quantized FL can be \textcolor{black}{given} by Theorem 1. \\
\noindent\textbf{Theorem 1.} \textit{\textcolor{black}{(Convergence of FedAQ for non-convex loss functions) Consider the FL system introduced in Section II.} After a given K communication rounds, the convergence bound is
\begin{align}
\frac{1}{K\tau}\sum_{m=0}^{K-1}\sum_{t=0}^{\tau -1}\mathbb{E}\Vert \nabla f(\bar{\mathbf{w}}_{m,t})\Vert^2 &\leq  \frac{Ld}{n^2K\tau\eta}\sum_{m=0}^{K-1}\sum_{i\in [n]}(\frac{R_m^i(\Delta \mathbf{w})}{s_m^i})^2 + \frac{2Ld}{K\tau\eta}\sum_{m=0}^{K-1}(\frac{R_m(\mathbf{w})}{s_m})^2 \nonumber\\
&+\frac{2(f(\mathbf{w}_0)-f^*)}{K\tau\eta} +\frac{L\eta\sigma^2+ L^2\eta^2(n+1)(\tau-1)\sigma^2}{n}, 
\label{Theorem1}
\end{align}
where $\bar{\mathbf{w}}_{m,t}$ is the averaged model, $f^*$ denotes the lower bound of training loss, \textcolor{black}{and the other symbols are defined in Table~\ref{tab:notations}.} }

\textit{Proof.} The proof is given in Appendix.\\
The first and second terms on the right-hand side (RHS) of Eq.(\ref{Theorem1}) capture the impact of uplink and downlink quantization, respectively. The third term captures the difference between the initial and final training loss. The fourth term comes from the stochastic gradient descending.
\subsection{\textcolor{black}{Joint Uplink and Downlink Adaptive Quantization}}
Our target is to minimize the right-hand side of (\ref{Theorem1}) under the total uplink and downlink energy \textcolor{black}{constraint} $E$, by \textcolor{black}{optimizing} both the uplink quantization level $s_m^i$ and the downlink quantization level $s_m$. \textcolor{black}{Given} the other terms on the right-hand side of (\ref{Theorem1}) are constant, the problem can be simplified \textcolor{black}{as} 
\begin{align}
&\mathop{min}\limits_{s_m^i,s_m} \ \frac{Ld}{n^2K\tau\eta}\sum_{m=0}^{K-1}\sum_{i\in [n]}(\frac{R_m^i(\Delta \mathbf{w})}{s_m^i})^2 + \frac{2Ld}{K\tau\eta}\sum_{m=0}^{K-1}(\frac{R_m(\mathbf{w})}{s_m})^2 \nonumber\\
&s.t. \quad e_1\sum_{m=0}^{K-1}\sum_{i\in [n]}d\lceil log_2s^i_m \rceil+e_2\sum_{m=0}^{K-1}nd\lceil log_2s_m \rceil= E,
\label{optimization}
\end{align}
where $e_1$ and $e_2$ represent the \textcolor{black}{per bit energy for uplink and downlink transmission, respectively.}

To solve the above optimization problem, we can use the variant of the Cauchy-Schwarz inequality:
\begin{align}
&\sum_{i=1}^nx_i^2 \geq \frac{1}{n}\left(\sum_{i=1}^nx_i\right)^2,
\label{cauchy}
\end{align}
\textcolor{black}{with equality if and only if $x_1 = x_2=...=x_n$.
By applying (\ref{cauchy}) to the} first term in the objective function (\ref{optimization}), we \textcolor{black}{can obtain}
\begin{align}
\sum_{m=0}^{K-1}\sum_{i\in [n]}(\frac{R_m^i(\Delta \mathbf{w})}{s_m^i})^2 \geq \sum_{m=0}^{K-1}\frac{1}{n}(\sum_{i\in [n]}\frac{R_m^i(\Delta \mathbf{w})}{s_m^i})^2,
\label{uplink_opt1}
\end{align}
and the condition for equality is 
\begin{align}
\frac{R^0_m(\Delta \mathbf{w})}{s^0_m}=\frac{R^1_m(\Delta \mathbf{w})}{s^1_m}=...=\frac{R^n_m(\Delta \mathbf{w})}{s^n_m}.
\label{uplink_condition1}
\end{align}
\textcolor{black}{By applying (\ref{cauchy})} to (\ref{uplink_opt1}), we get
\begin{align}
\sum_{m=0}^{K-1}\frac{1}{n}(\sum_{i\in [n]}\frac{R_m^i(\Delta \mathbf{w})}{s_m^i})^2 \geq \frac{1}{Kn}(\sum_{m=0}^{K-1}\sum_{i\in [n]}\frac{R_m^i(\Delta \mathbf{w})}{s_m^i})^2,
\label{uplink_opt2}
\end{align}
and the condition for equality is 
\begin{align}
\sum_{i\in [n]}\frac{R^i_0(\Delta \mathbf{w})}{s^i_0}=\sum_{i\in [n]}\frac{R^i_1(\Delta \mathbf{w})}{s^i_1}=...=\sum_{i\in [n]}\frac{R^i_{K-1}(\Delta \mathbf{w})}{s^i_{K-1}}.
\label{uplink_condition2}
\end{align}
Combining (\ref{uplink_opt1}) and (\ref{uplink_opt2}), we get
\begin{align}
\sum_{m=0}^{K-1}\sum_{i\in [n]}(\frac{R_m^i(\Delta \mathbf{w})}{s_m^i})^2 \geq \frac{1}{Kn}(\sum_{m=0}^{K-1}\sum_{i\in [n]}\frac{R_m^i(\Delta \mathbf{w})}{s_m^i})^2,
\label{uplink_opt}
\end{align}
and the condition for equality is
\begin{align}
\frac{R^i_m(\Delta \mathbf{w})}{s^i_m}=\alpha, 
\label{uplink_condition}
\end{align}
\textcolor{black}{where $i = 0,1,...,n$, $m=0,1,...,K-1$, and} $\alpha$ is a constant. This shows that to minimize the uplink term, the ratio between the uplink model update range and quantization level, i.e. $\frac{R^i_m(\Delta \mathbf{w})}{s^i_m}$, should be \textcolor{black}{identical for all clients} in every communication round.

\textcolor{black}{According to (\ref{cauchy}),} the second term in the objective function (\ref{optimization}) \textcolor{black}{can be minimized as}
\begin{align}
\sum_{m=0}^{K-1}(\frac{R_m(\mathbf{w})}{s_m})^2 \geq \frac{1}{K}(\sum_{m=0}^{K-1}\frac{R_m(\mathbf{w})}{s_m})^2,
\label{downlink_opt}
\end{align}
and the condition for equality is
\begin{align}
\frac{R_0(\mathbf{w})}{s_0}=\frac{R_1(\mathbf{w})}{s_1}=...=\frac{R_{K-1}(\mathbf{w})}{s_{K-1}}=\beta,
\label{downlink_condition}
\end{align}
 where $\beta$ is a constant. This shows that to minimize the downlink term, the ratio between the range of the downlink model and downlink quantization level, i.e., $\frac{R_m(\mathbf{w})}{s_m}$, is expected to be the same for every communication round.

\textcolor{black}{By substituting} (\ref{uplink_opt}) and (\ref{downlink_opt}) into the objective function in (\ref{optimization}), we have
\begin{align}
&\frac{2Ld}{K\tau\eta}\sum_{m=0}^{K-1}(\frac{R_m(\mathbf{w})}{s_m})^2 + \frac{Ld}{n^2K\tau\eta}\sum_{m=0}^{K-1}\sum_{i\in [n]}(\frac{R_m^i(\Delta \mathbf{w})}{s_m^i})^2 \nonumber\\
& \geq \frac{2Ld}{K^2\tau\eta}(\sum_{m=0}^{K-1}\frac{R_m(\mathbf{w})}{s_m})^2 + \frac{Ld}{n^3K^2\tau\eta}(\sum_{m=0}^{K-1}\sum_{i\in [n]}\frac{R_m^i(\Delta \mathbf{w})}{s_m^i})^2 \nonumber\\
&\geq \frac{2\sqrt{2}Ld}{n\sqrt{n}K^2\tau\eta}(\sum_{m=0}^{K-1}\frac{R_m(\mathbf{w})}{s_m})(\sum_{m=0}^{K-1}\sum_{i\in [n]}\frac{R_m^i(\Delta \mathbf{w})}{s_m^i}),
\label{dnup_opt}
\end{align}
and the condition for equality is 
\begin{align}
\sqrt{\frac{2Ld}{K^2\tau\eta}(\sum_{m=0}^{K-1}\frac{R_m(\mathbf{w})}{s_m})^2} = \sqrt{\frac{Ld}{n^3K^2\tau\eta}(\sum_{m=0}^{K-1}\sum_{i\in [n]}\frac{R_m^i(\Delta \mathbf{w})}{s_m^i})^2}.
\label{dnup_condition}
\end{align}
Introducing the other two conditions for equality, (\ref{uplink_condition}) and (\ref{downlink_condition}), into (\ref{dnup_condition}), we can simplify (\ref{dnup_condition}) as 
\begin{align}
\beta=\frac{\alpha}{\sqrt{2n}}.
\label{dnup_condition1}
\end{align}
Concluding the optimization steps, the problem in (\ref{optimization}) can be finally optimized as
\begin{align}
&\frac{2Ld}{K\tau\eta}\sum_{m=0}^{K-1}(\frac{R_m(\mathbf{w})}{s_m})^2 + \frac{Ld}{n^2K\tau\eta}\sum_{m=0}^{K-1}\sum_{i\in [n]}(\frac{R_m^i(\Delta \mathbf{w})}{s_m^i})^2 \nonumber\\
&\geq \frac{2\sqrt{2}Ld}{n\sqrt{n}K^2\tau\eta}(\sum_{m=0}^{K-1}\frac{R_m(\mathbf{w})}{s_m})(\sum_{m=0}^{K-1}\sum_{i\in [n]}\frac{R_m^i(\Delta \mathbf{w})}{s_m^i}),
\label{final_opt}
\end{align}
and the condition for equality is 
\begin{eqnarray}
\begin{cases}
\frac{R^i_m(\Delta \mathbf{w})}{s^i_m}=\alpha, \\
\frac{R_m(\mathbf{w})}{s_m}=\beta=\frac{\alpha}{\sqrt{2n}}.
\end{cases}
\label{final_condition}
\end{eqnarray}
Introducing (\ref{final_condition}) into (\ref{optimization}), we get
\begin{align}
\alpha=2^{\frac{e_1\sum_{m=0}^{K-1}\sum_{i\in [n]}log_2R^i_m(\Delta\mathbf{w})}{Kn(e_1+e_2)}+\frac{e_2\sum_{m=0}^{K-1}log_2\sqrt{2n}R_m(\mathbf{w})}{K(e_1+e_2)}-\frac{E}{Knd(e_1+e_2)}}.
\label{alpha}
\end{align}

\begin{figure}[t]
\centering
\subfigure[Range of local model updates.]{
\begin{minipage}[b]{0.4\linewidth}
\includegraphics[width=1\linewidth]{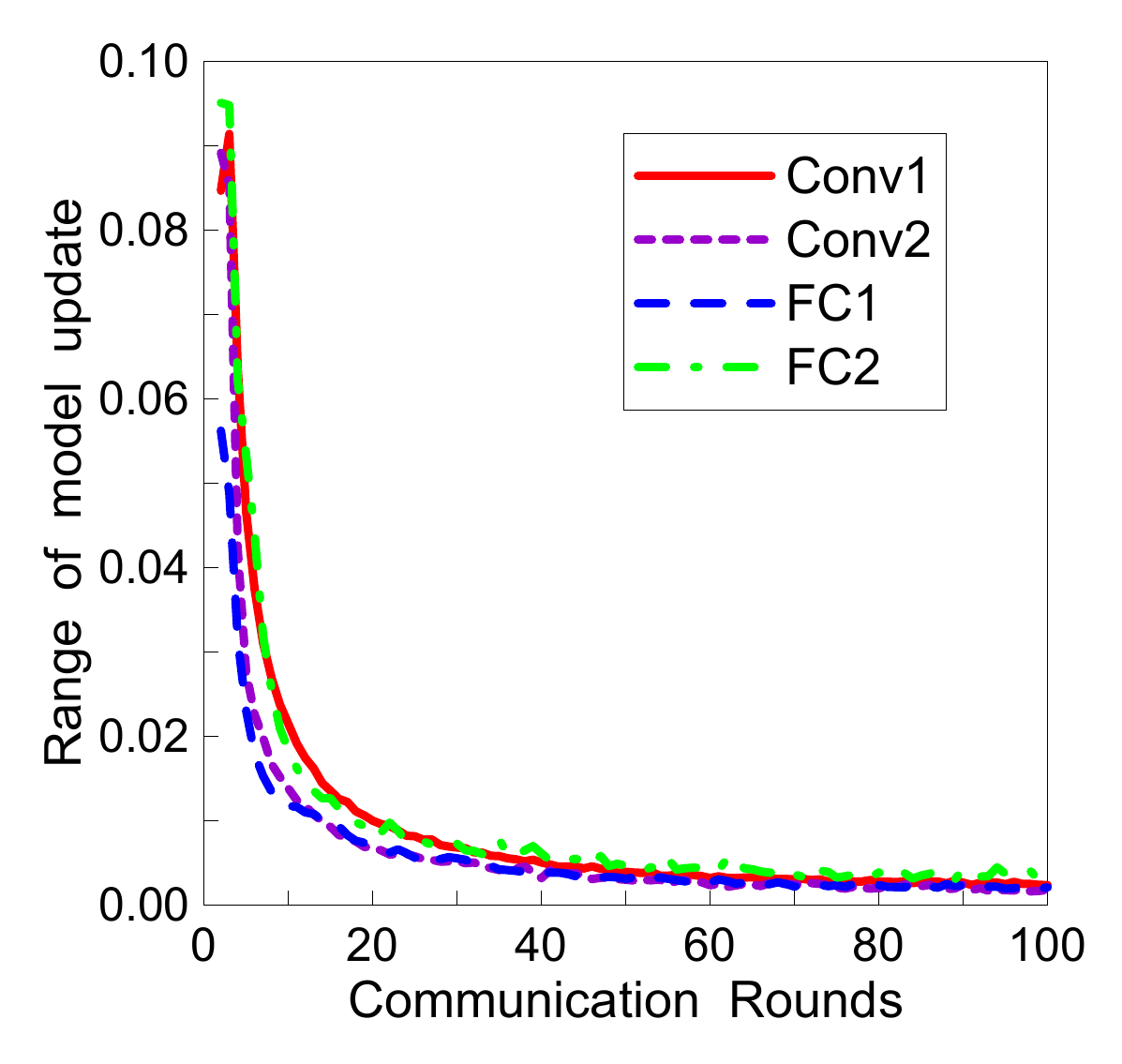}
\end{minipage}}
\qquad
\qquad
\subfigure[Range of the global model.]{
\begin{minipage}[b]{0.4\linewidth}
\includegraphics[width=1\linewidth]{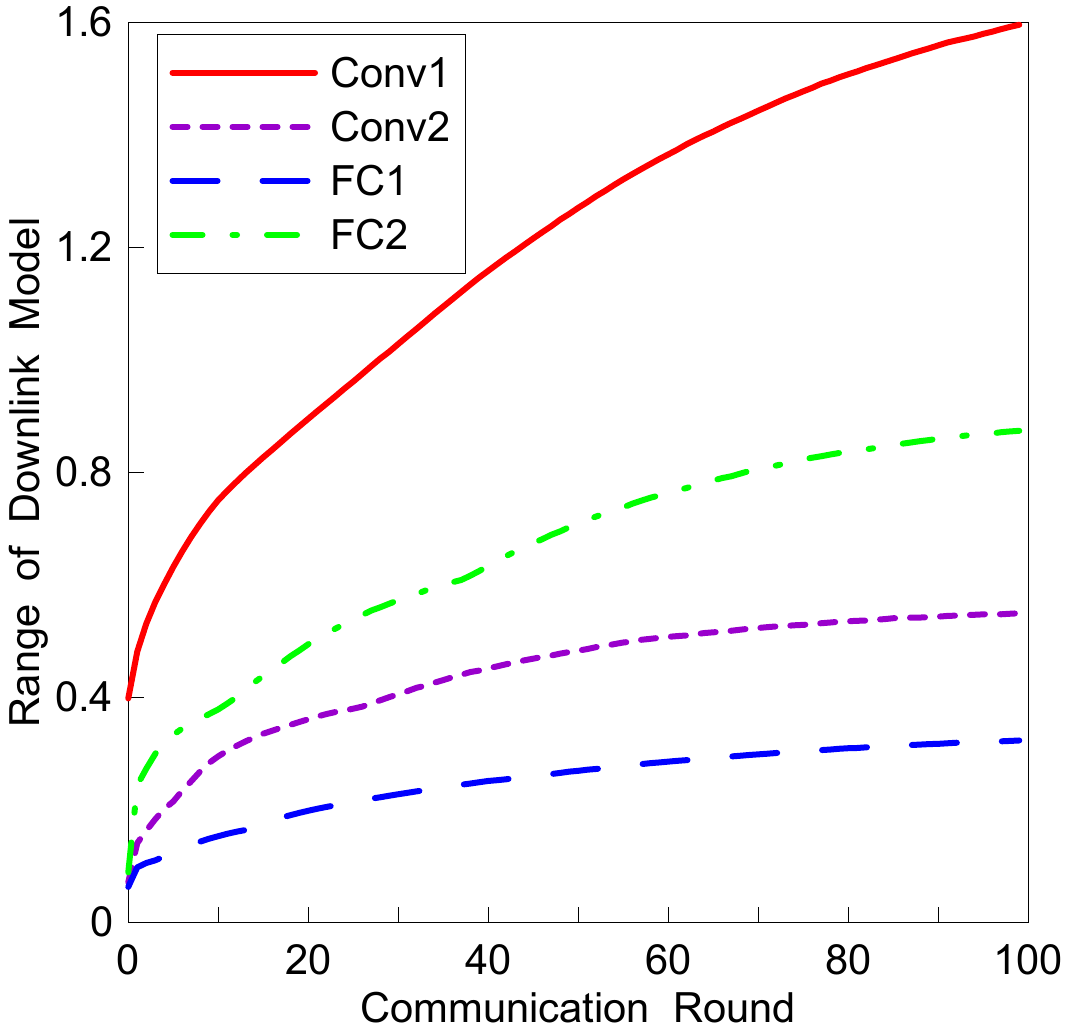}
\end{minipage}}
\caption{The range of local model updates in the uplink is decreasing with the communication round, while the range of the global model in the downlink is increasing. This is an example by the Fashion-MNIST experiment.}
\label{fig: model range}
\end{figure}

Finally, to achieve the best learning performance, for a given $K$ rounds of communication with the energy budget of $E$, the optimal quantization level for the uplink and downlink in the $m^{th}$ communication round is 
\begin{eqnarray}
\begin{cases}
uplink:\ bit_m^i= \lceil log_2(\frac{R_m^i(\Delta \mathbf{w})}{\alpha}) \rceil, \\
downlink:\ bit_m= \lceil log_2(\frac{R_m(\mathbf{w})}{\beta}) \rceil= \lceil log_2(\frac{\sqrt{2n}\ R_m(\mathbf{w})}{\alpha}) \rceil.
\end{cases}
\label{finalbit}
\end{eqnarray}
This shows that the optimal quantization bit depends on the range of communicated parameters. For the uplink, the optimal quantization bit depends on the range of the local model updates which has a decreasing trend, as shown in Fig.~\ref{fig: model range}(a). For the downlink, the optimal quantization bit depends on the range of the global model which has an increasing trend, as shown in Fig.~\ref{fig: model range}(b). This supports our quantization scheme using descending quantization in the uplink and ascending quantization in the downlink.

\subsection{Algorithm}
Based on the theoretical analysis, we propose a joint uplink and downlink adaptive quantization scheme. The detailed bit-changing rule has been given in (\ref{finalbit}). The algorithm is summerized in Algorithm~\ref{algo}. \textcolor{black}{Notice that in the $m^{th}$ communication round, the global model weights are quantized before broadcasting, where the bit-length can be decided by the function of $bit_m$ in (\ref{finalbit}). Before uplink transmission, for the $i^{th}$ client, the model updates are quantized by following the function of $bit_m^i$ in (\ref{finalbit}).}

\begin{algorithm}[t]
\setstretch{1.0}
\caption{System algorithm}
\label{algo}
\textbf{Input: }{The initial global model $\mathbf{w}_0$}\\
\textbf{Output: }{The final global model $\mathbf{w}_{K-1}$}\\
\For{Communication round $m=0,1,2,...,K-1$}
{downlink quantization: $\mathbf{w}_m \rightarrow Q_2(\mathbf{w}_m)$, with\\
$bit_m=\lceil log_2(\frac{R_m(\mathbf{w})}{\beta}) \rceil= \lceil log_2(\frac{\sqrt{2n}\cdot R_m(\mathbf{w})}{\alpha}) \rceil$\\    
    \For{Client $i$ in $\{1,2,...,n\}$}
    { Receive quantized global model: $\mathbf{w}^i_{m,0}=Q_2(\mathbf{w}_m)$\\ 
        \For{local training step $t=0,1,...,\tau$}
        {
        $\mathbf{w}^i_{m,t+1}=\mathbf{w}^i_{m,t}-\eta_m\nabla f(\mathbf{w}^i_{m,t})$\
        }
    $\Delta \mathbf{w}_m^i=\mathbf{w}^i_{m,\tau}-Q_2(\mathbf{w}_m^i)$\\
    uplink quantization: $\Delta\mathbf{w}_m^i \rightarrow Q_1(\Delta \mathbf{w}_m^i)$, with\\
    $bit_m^i= \lceil log_2(\frac{R_m^i(\Delta \mathbf{w})}{\alpha}) \rceil$\
    }
$\mathbf{w}_{m+1}=\mathbf{w}_m+\sum_{i=1}^np_iQ_1(\Delta \mathbf{w}_m^i)$
}
\end{algorithm}

\section{Special Cases Analysis}
\label{Special Cases Analysis}
In this section, we analyze the performance of two special cases: uplink-only quantization and downlink-only quantization.
\subsection{Uplink-only Quantization}
For uplink-only quantization, the downlink communication is assumed to be perfect. The problem can be formulated as, how to optimize the learning performance by tuning the uplink quantization level, for a given constraint on the uplink transmission energy. Because of page limitation, we here omit the detailed derivation process which is easy to obtain by making certain changes in our full-link theoretical analysis. After the derivation process, we get the following Theorem 2 for uplink-only quantization.\\
\noindent\textbf{Theorem 2.} \textit{ For a given K communication rounds, the convergence bound is}
\begin{align}
\frac{1}{K\tau}\sum_{m=0}^{K-1}\sum_{t=0}^{\tau -1}\mathbb{E}\Vert \nabla f(\bar{\mathbf{w}}_{m,t})\Vert^2 &\leq  \frac{Ld}{n^2K\tau\eta}\sum_{m=0}^{K-1}\sum_{i\in [n]}(\frac{R_m^i(\Delta \mathbf{w})}{s_m^i})^2 \nonumber\\
&+\frac{2(f(\mathbf{w}_0)-f^*)}{K\tau\eta} +\frac{L\eta\sigma^2+ L^2\eta^2(n+1)(\tau-1)\sigma^2}{n}, 
\label{Theorem2}
\end{align}
which exactly matches the derivation result in our previous conference paper \cite{qu2022feddq}.
To minimize the convergence upper bound for a given uplink energy constraint, we formulate this problem into
\begin{align}
&\mathop{min}\limits_{s_m^i} \ \frac{Ld}{n^2K\tau\eta}\sum_{m=0}^{K-1}\sum_{i\in [n]}(\frac{R_m^i(\Delta \mathbf{w})}{s_m^i})^2 \nonumber\\
&s.t. \quad e_1\sum_{m=0}^{K-1}\sum_{i\in [n]}d\lceil log_2s^i_m \rceil= E_1.
\label{objective2}
\end{align}
\textcolor{black}{Similar to (\ref{uplink_opt}), we can get the solution of (\ref{objective2}) as
\begin{align}
&s^i_m=\frac{R^i_m(\Delta \mathbf{w})}{\alpha} \nonumber\\
&bit_m^i= \lceil log_2(\frac{R_m^i(\Delta \mathbf{w})}{\alpha}) \rceil \nonumber\\
&\alpha=2^{\frac{1}{Kn}\sum_{m=0}^{K-1}\sum_{i\in [n]}log_2R^i_m(\Delta\mathbf{w})-\frac{E_1}{Knde_1}}
\label{uplinkonly_bit}
\end{align}
}
Since the range of local model updates has a decreasing trend, as shown in Fig.\ref{fig: model range}(a), the optimized adaptive quantization bit in the uplink is expected to follow a descending trend. This conclusion is consistent with the results in our conference paper\cite{qu2022feddq}.

\subsection{Downlink-only Quantization}
For downlink-only quantization, the uplink communication is assumed to be perfect. The problem can be formulated as, how to optimize the learning performance by tuning the downlink quantization level, for a given constraint on the downlink receiving energy. After derivation, we get the following Theorem 3 for downlink-only quantization.\\
\noindent\textbf{Theorem 3.} \textit{ For a given K communication rounds, the convergence bound is}
\begin{align}
\frac{1}{K\tau}\sum_{m=0}^{K-1}\sum_{t=0}^{\tau -1}\mathbb{E}\Vert \nabla f(\bar{\mathbf{w}}_{m,t})\Vert^2 &\leq \frac{2Ld}{K\tau\eta}\sum_{m=0}^{K-1}(\frac{R_m(\mathbf{w})}{s_m})^2 \nonumber\\
&+\frac{2(f(\mathbf{w}_0)-f^*)}{K\tau\eta} +\frac{L\eta\sigma^2+ L^2\eta^2(n+1)(\tau-1)\sigma^2}{n}.
\label{Theorem3}
\end{align}
To minimize the convergence upper bound for a given downlink energy constraint, we formulate this problem into
\begin{align}
&\mathop{min}\limits_{s_m} \ \frac{2Ld}{K\tau\eta}\sum_{m=0}^{K-1}(\frac{R_m(\mathbf{w})}{s_m})^2 \nonumber\\
&s.t. \quad e_2\sum_{m=0}^{K-1}nd\lceil log_2s_m \rceil= E_2.
\label{objective3}
\end{align}
\textcolor{black}{Similar to (\ref{downlink_opt}), we solve this optimization problem as
\begin{align}
&s_m=\frac{R_m(\mathbf{w})}{\beta} \nonumber\\
&bit_m= \lceil log_2(\frac{R_m(\mathbf{w})}{\beta}) \rceil \nonumber\\
&\beta=2^{\frac{1}{K}\sum_{m=0}^{K-1}log_2R_m(\mathbf{w})-\frac{E_2}{Knde_2}}
\label{downlinkonly_bit}
\end{align}
}
Since the range of the global model has an increasing trend, as shown in Fig.\ref{fig: model range}(b), the optimized adaptive quantization bit in the downlink is expected to follow an increasing trend.

\section{Experiment}
\label{Expirement}
In this section, we perform experiments on open datasets to validate the effectiveness of the proposed adaptive quantization method. The experiments will be performed in three scenarios: joint uplink and downlink quantization, uplink-only quantization, and downlink-only quantization. 

\subsection{\textcolor{black}{Datasets and DNN Models}}
Three datasets are used in the experiments: \textcolor{black}{MNIST\cite{lecun1998mnist}, Fashion-MNIST\cite{xiao2017fashion}, and CIFAR-10\cite{krizhevsky2010cifar}.}
The training datasets of MNIST, Fashion-MNIST, and CIFAR-10 are split among all clients in an I.I.D manner, and the test datasets are used to perform validation on the server side.

The experiments utilize four DNN networks:
\begin{itemize}
\item{LeNet-300-100\cite{han2015deep}: This is a fully connected neural network with two hidden layers, with 300 and 100 neurons for each, and a final softmax output layer with around 0.27M parameters in total.}
\item{Vanilla CNN \cite{mcmahan2017communication}: This network consists of two 5$\times$5 convolution layers (the first has 32 filters and the second has 64 filters, each followed by 2$\times$2 max pooling and ReLu activation), one fully connected layer with 512 unities followed by ReLu activation, and a final softmax output layer with around 0.58M parameters in total.}
\item{7-layer CNN: This network consists of four 3$\times$3 convolutional layers with padding 1 (the first with 48 filters followed by ReLu activation, the second with 96 filters followed by ReLu activation, 2$\times$2 max pooling, and 0.25 dropout, the third with 192 filters followed by ReLu activation, and the fourth with 256 filters followed by ReLu activation, 2$\times$2 max pooling, and 0.25 dropout), two fully connected layers (the first with 512 unities followed by ReLu activation, the second with 64 unities followed by ReLu activation and 0.25 dropout), and a final softmax output layer with around 9.07M parameters in total.}
\item{ResNet-18\cite{he2016deep}: ResNet is a very famous neural network for image recognition tasks. The detailed network structure can be found at https://pytorch.org/hub/pytorch\_vision\_resnet. In this work, we choose ResNet-18, which contains around 11.6M parameters in total.}
\end{itemize}

We conduct four sets of experiments: LeNet-300-100 on MNIST, vanilla CNN on Fashion-MNIST, 7-layer CNN on CIFAR-10, and ResNet-18 on CIFAR-10. All experiments are conducted on a computer with a 6th Gen Intel(R) Xeon(R) W-2245 @3.90GHz CPU and NVIDIA GeForce RTX 3090 GPU.

\textit{Hyper-parameters}: In this work, we suggest to set $\alpha$ between 0.003--0.005. The transmitting and receiving energy units are assumed to be $e_1=e_2=1pJ/b$. For the local training, the batch size is set to 64, and the momentum is 0.5 in the SGD optimizer. We set the learning rate at 0.01 and local update steps at 5. Ten local clients are used in the LeNet-300-100 on MNIST, vanilla CNN on Fashion-MNIST, and 7-layer CNN on CIFAR-10 experiments. Because the model size of ResNet-18 is much larger than the above three models, considering about the simulation time, four local clients are used in the ResNet-18 on CIFAR-10 experiment. We compare both the training loss and test accuracy for all the experiments. In this paper, the training loss is calculated as the average loss of all clients' local training \cite{haddadpour2021federated}.

\subsection{Joint Uplink and Downlink Adaptive Quantization}

\begin{figure*}
\centering
\subfigure[LeNet-300-100 on MNIST]{
\begin{minipage}[b]{0.3\linewidth}
\includegraphics[width=1\linewidth]{imgs/full_lenet_mnist_energy.pdf}
\end{minipage}}
\quad
\subfigure[LeNet-300-100 on MNIST]{
\begin{minipage}[b]{0.3\linewidth}
\includegraphics[width=1\linewidth]{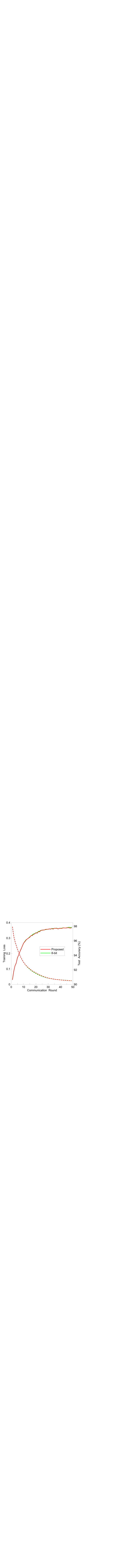}
\end{minipage}}
\quad
\subfigure[LeNet-300-100 on MNIST]{
\begin{minipage}[b]{0.3\linewidth}
\includegraphics[width=0.9\linewidth]{imgs/full_lenet_mnist_bit.pdf}
\end{minipage}}
\quad
\subfigure[vanilla-CNN on Fashion-MNIST]{
\begin{minipage}[b]{0.3\linewidth}
\includegraphics[width=1\linewidth]{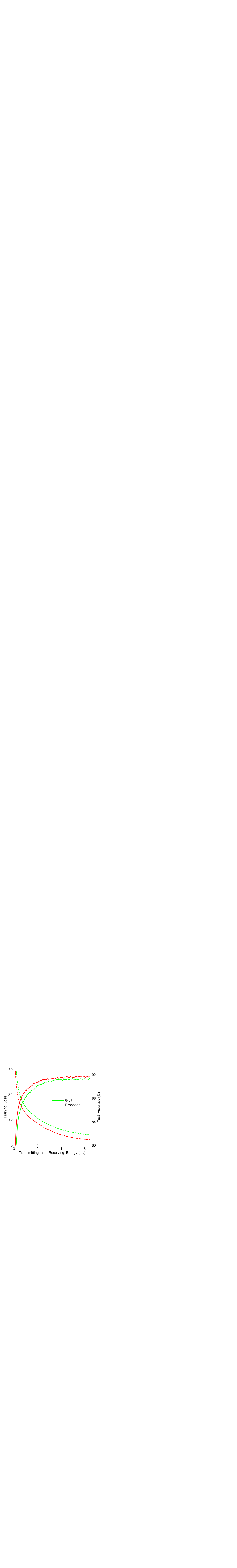}
\end{minipage}}
\quad
\subfigure[vanilla-CNN on Fashion-MNIST]{
\begin{minipage}[b]{0.3\linewidth}
\includegraphics[width=1\linewidth]{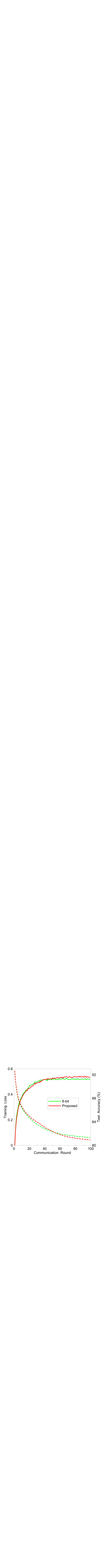}
\end{minipage}}
\quad
\subfigure[vanilla-CNN on Fashion-MNIST]{
\begin{minipage}[b]{0.3\linewidth}
\includegraphics[width=0.9\linewidth]{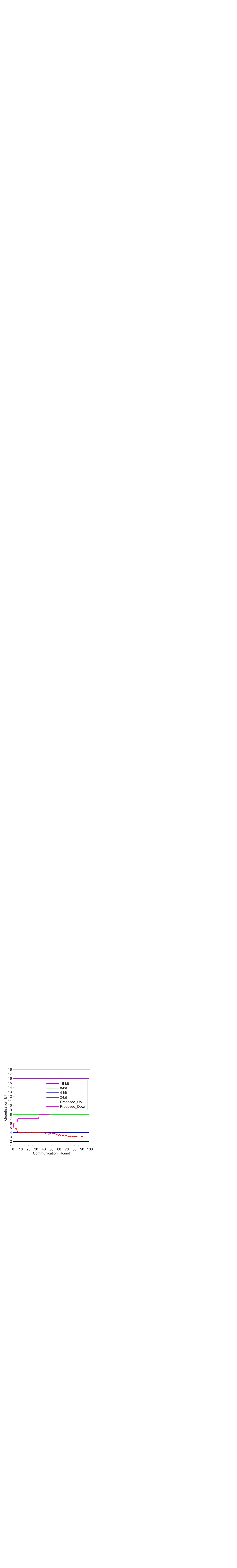}
\end{minipage}}
\quad
\subfigure[7-layer CNN on CIFAR-10]{
\begin{minipage}[b]{0.3\linewidth}
\includegraphics[width=1\linewidth]{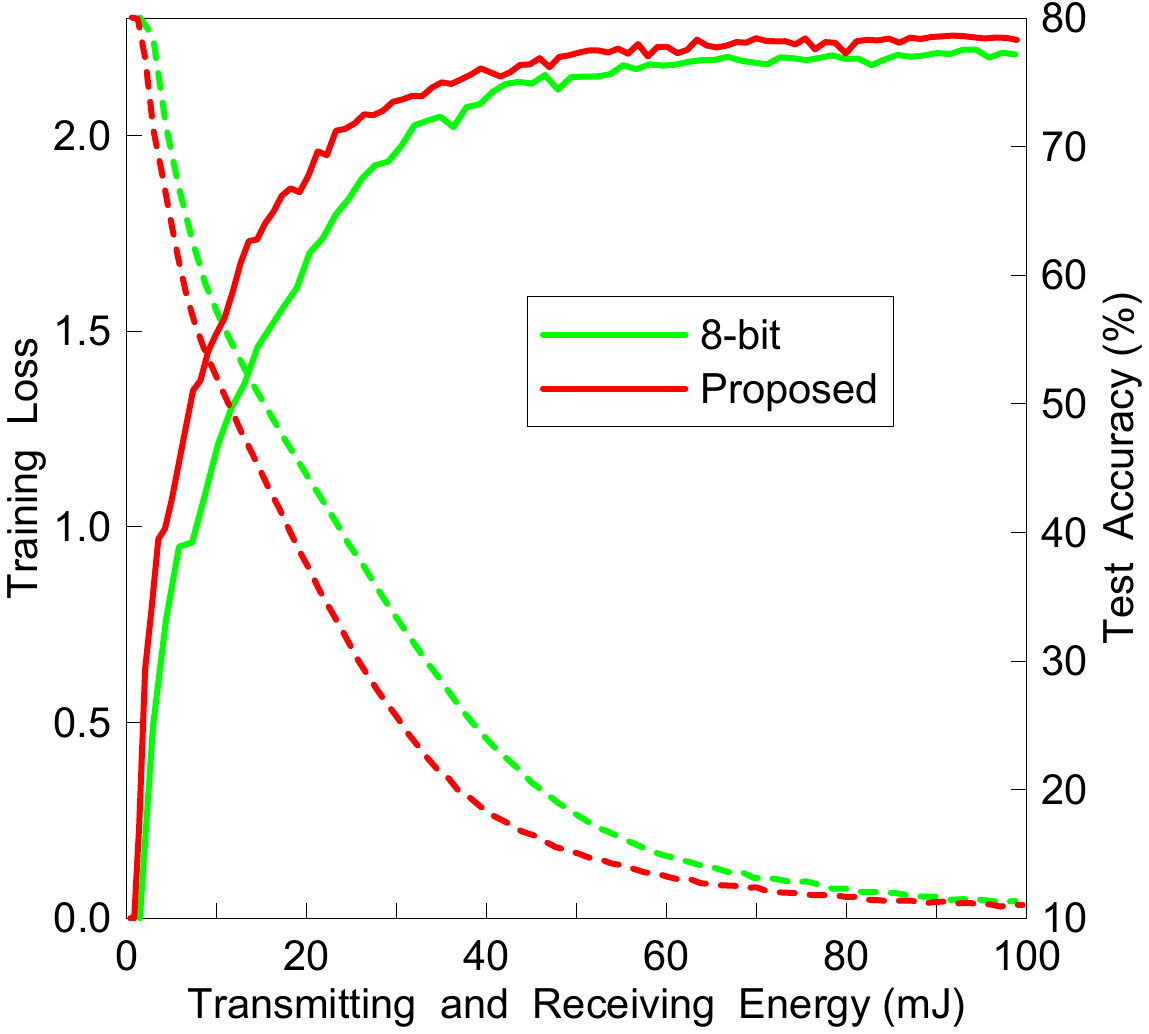}
\end{minipage}}
\quad
\subfigure[7-layer CNN on CIFAR-10]{
\begin{minipage}[b]{0.3\linewidth}
\includegraphics[width=1\linewidth]{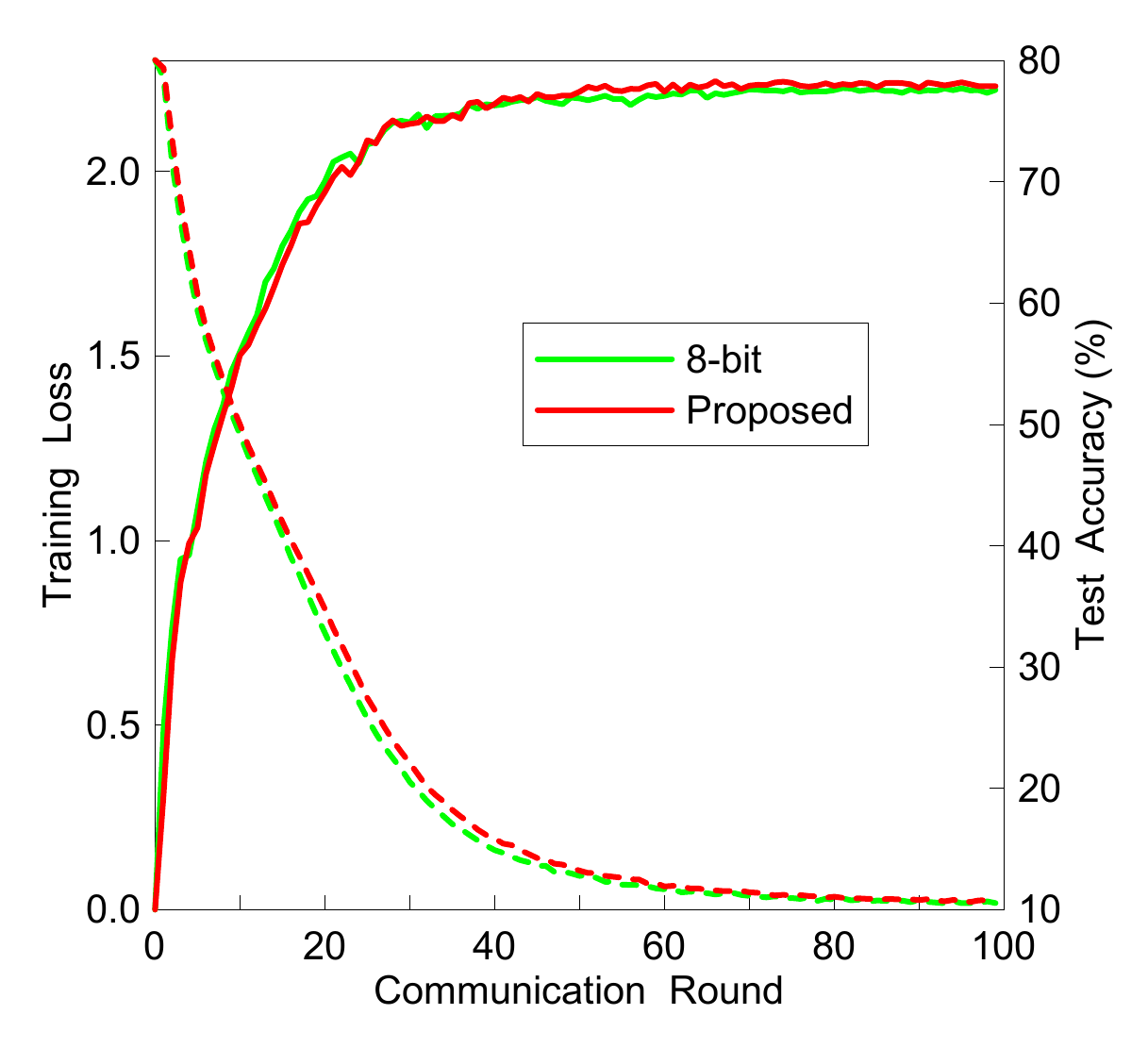}
\end{minipage}}
\quad
\subfigure[7-layer CNN on CIFAR-10]{
\begin{minipage}[b]{0.3\linewidth}
\includegraphics[width=0.9\linewidth]{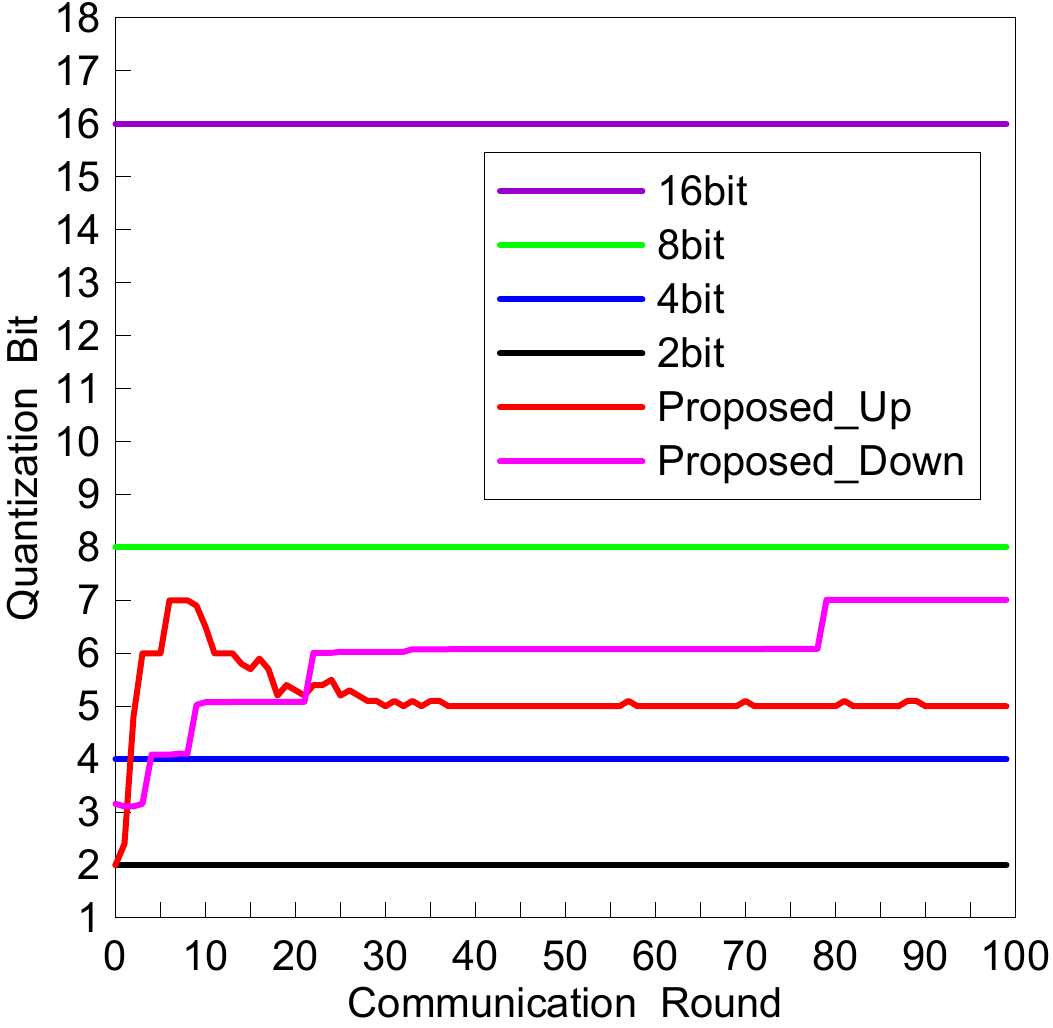}
\end{minipage}}
\quad
\subfigure[ResNet-18 on CIFAR-10]{
\begin{minipage}[b]{0.3\linewidth}
\includegraphics[width=1\linewidth]{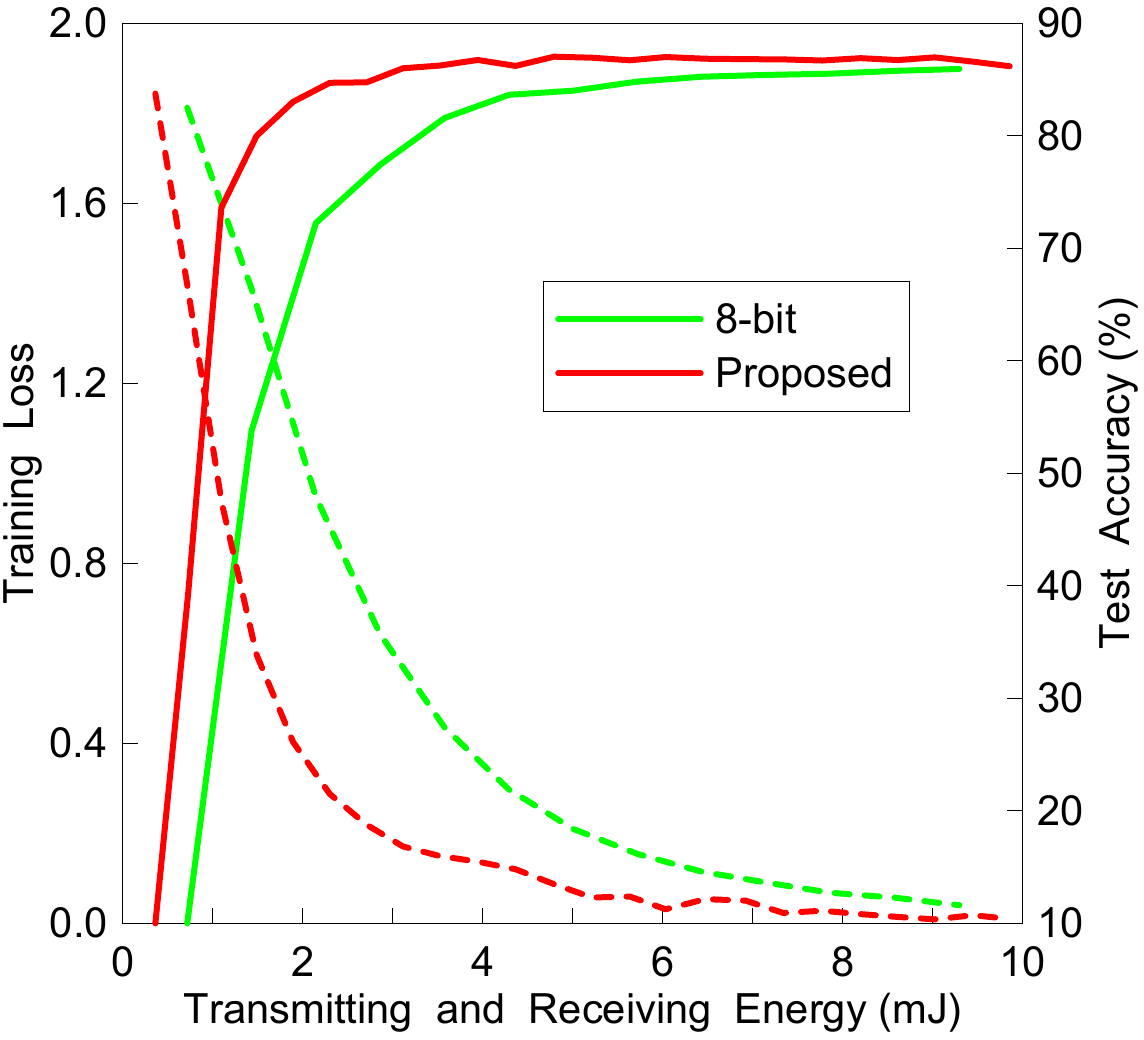}
\end{minipage}}
\quad
\subfigure[ResNet-18 on CIFAR-10]{
\begin{minipage}[b]{0.3\linewidth}
\includegraphics[width=1\linewidth]{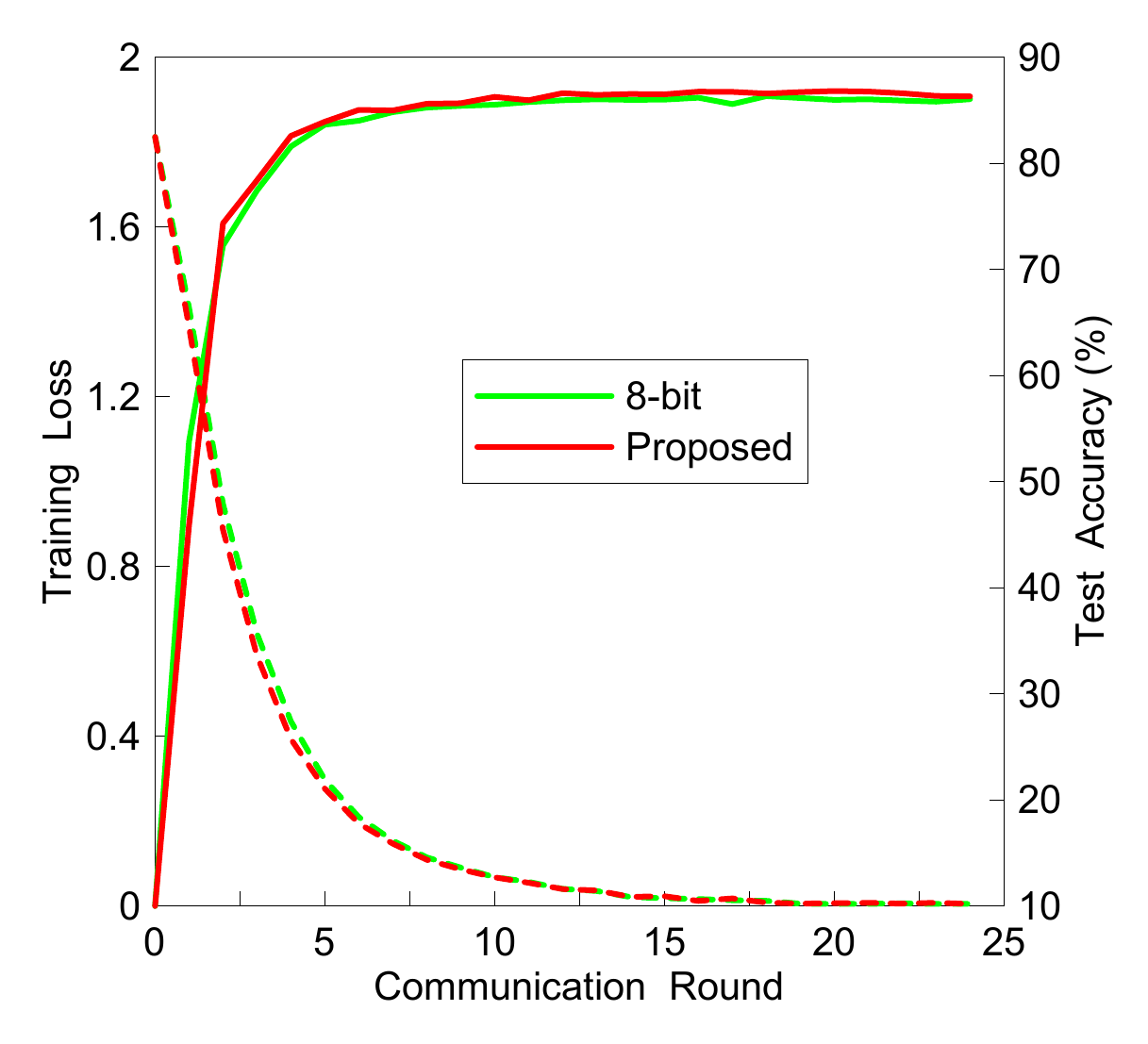}
\end{minipage}}
\quad
\subfigure[ResNet-18 on CIFAR-10]{
\begin{minipage}[b]{0.3\linewidth}
\includegraphics[width=0.9\linewidth]{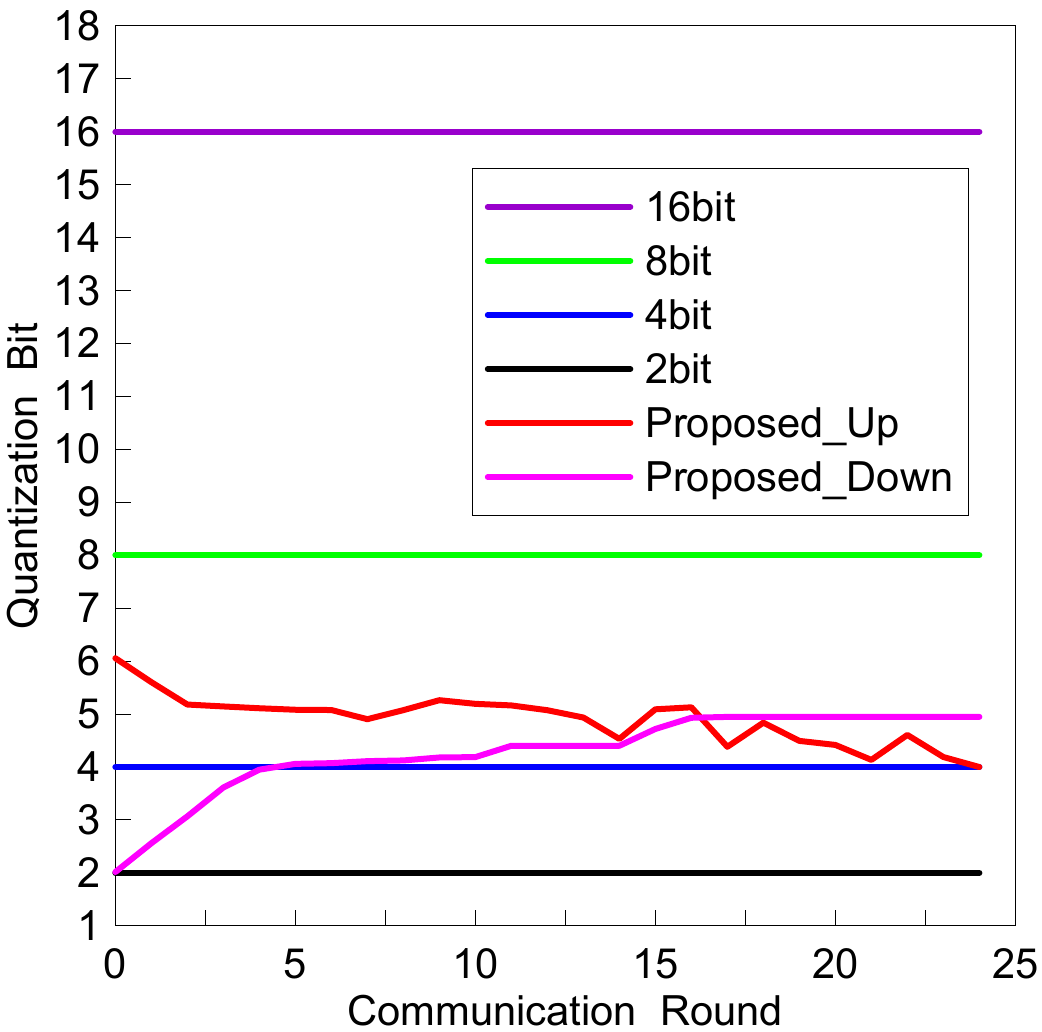}
\end{minipage}}
\caption{Results of joint uplink and downlink quantization experiments.}
\label{fig:full}
\end{figure*}

\begin{table}[t]
\caption{Joint uplink and downlink quantization}
\centering
\begin{tabular}{c|c|c|c}
\hline  
&\makecell[c]{Fixed 8-bit}&\makecell[c]{Proposed}&\makecell[c]{Energy Saving}\\
\hline
\makecell[c]{LeNet-300-100 on MNIST\\ (Test Acc.=97.0\%)}&0.90mJ&\makecell[c]{0.67mJ}&\makecell[c]{25.6\%}\\
\hline  
\makecell[c]{Vanilla CNN on Fashion-MNIST\\ (Test Acc.=91.3\%)}&5.86mJ&\makecell[c]{2.80mJ}&\makecell[c]{52.2\%}\\
\hline  
\makecell[c]{7-layer CNN on CIFAR-10\\ (Test Acc.=77.2\%)}&90.0mJ&\makecell[c]{50.2mJ}&\makecell[c]{44.2\%}\\
\hline 
\makecell[c]{ResNet-18 on CIFAR-10\\(Test Acc.=86.0\%)}&9.3mJ&\makecell[c]{3.1mJ}&\makecell[c]{66.7\%}\\
\hline
\end{tabular}
\label{tab:full}
\end{table}

To the best of our knowledge, most existing quantization works focus on only one-side communication and there is no joint uplink and downlink adaptive quantization work. So we compare the joint uplink and downlink adaptive quantization with the best fixed-bit quantization scheme. For the fixed quantization, we tried 2-bit, 4-bit, 8-bit, and 16-bit quantization, and 8-bit quantization is found to be the best quantization scheme. Thus, our adaptive quantization is compared with fixed 8-bit quantization.

The results of the four experiments are shown in Fig.~\ref{fig:full}, where (a)--(c) are the results of LeNet-300-100 on MNIST, (d)--(f) are the results of vanilla CNN on Fashion-MNIST, (g)--(i) are the results of 7-layer CNN on CIFAR-10, and (j)-(l) are the results of ResNet-18 on CIFAR-10. First, let us look at the energy efficiency. It can be observed from (a),(d),(g), and (j) that the proposed adaptive quantization method consumes less energy than the optimal fixed quantization, i.e., 8-bit quantization. For example, as shown in (d), to achieve the same test accuracy of 91.3\% on the Fashion-MNIST dataset, the fixed 8-bit quantization scheme consumes 5.86mJ, while our scheme only consumes 2.80mJ, which achieves a reduction of 52.2\% of the communication energy. The detailed comparison can be found in Table~\ref{tab:full}. Regarding convergence speed, Fig.\ref{fig:full} (b),(e),(h), and (k) show that achieving the same test accuracy or training loss, the proposed adaptive scheme consumes a similar number of communication rounds with the fixed 8-bit quantization. Thus, to achieve the same learning performance, our method can save energy consumption without sacrificing convergence rate. This superior performance results from the effective strategy for bit resource allocation. Furthermore, as shown in (c),(f),(i), and (l), the quantization level shows a decreasing trend in the uplink, and an increasing trend in the downlink, which is consistent with the theoretical analyses.

\subsection{Uplink Only Design}

\begin{figure*}
\centering
\subfigure[MNIST]{
\begin{minipage}[b]{0.21\linewidth}
\includegraphics[width=1.1\linewidth]{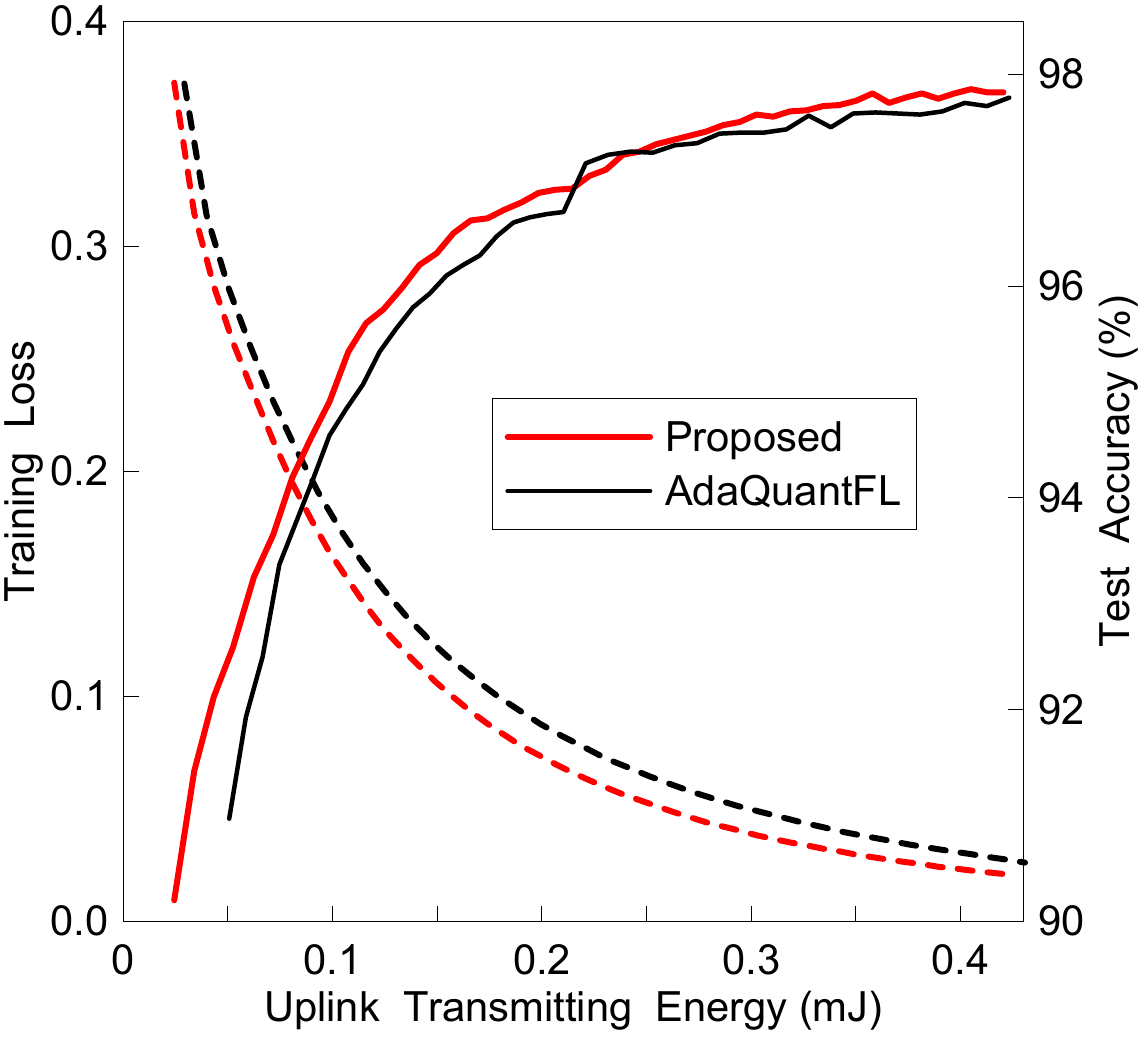}
\end{minipage}}
\quad
\subfigure[MNIST]{
\begin{minipage}[b]{0.21\linewidth}
\includegraphics[width=1.1\linewidth]{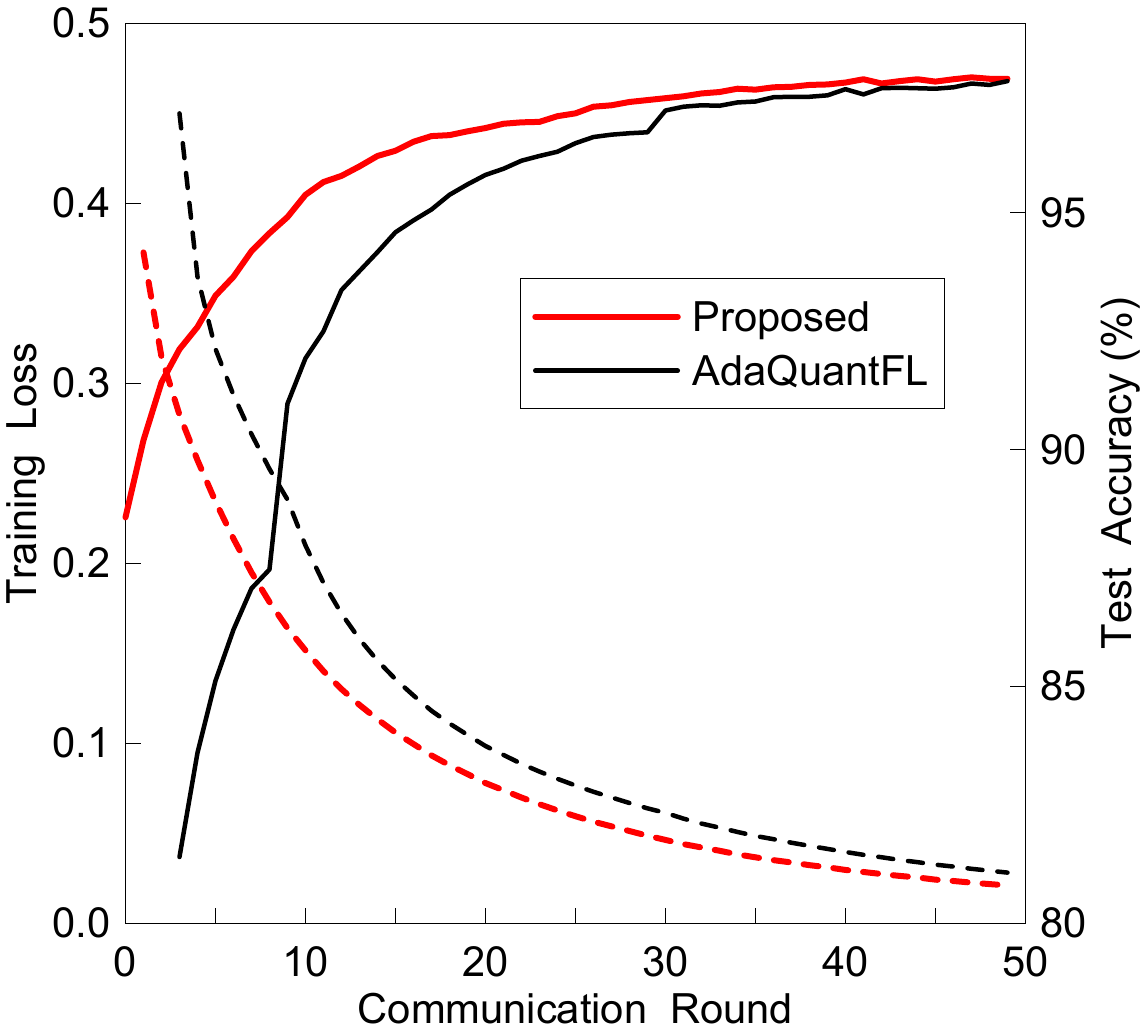}
\end{minipage}}
\quad
\subfigure[MNIST]{
\begin{minipage}[b]{0.21\linewidth}
\includegraphics[width=1\linewidth]{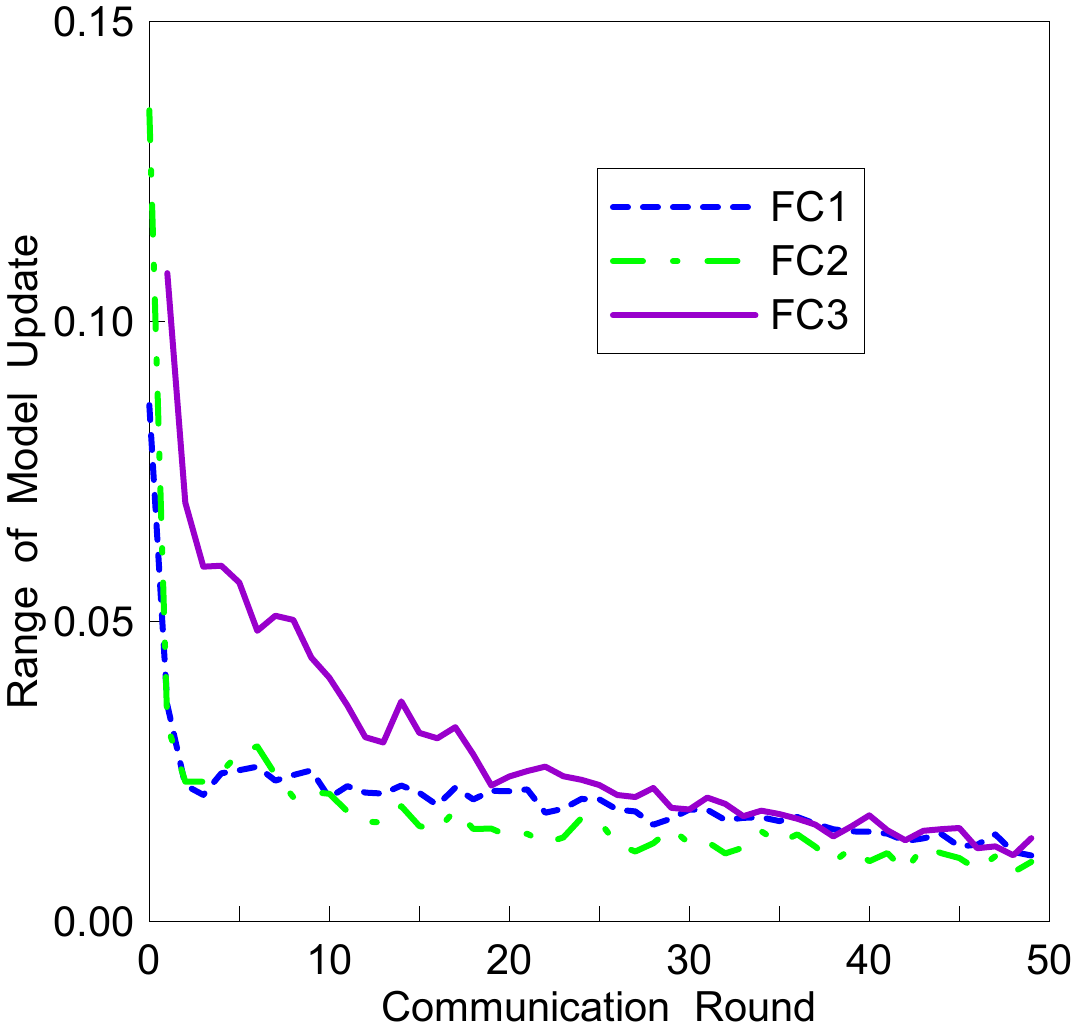}
\end{minipage}}
\quad
\subfigure[MNIST]{
\begin{minipage}[b]{0.21\linewidth}
\includegraphics[width=1\linewidth]{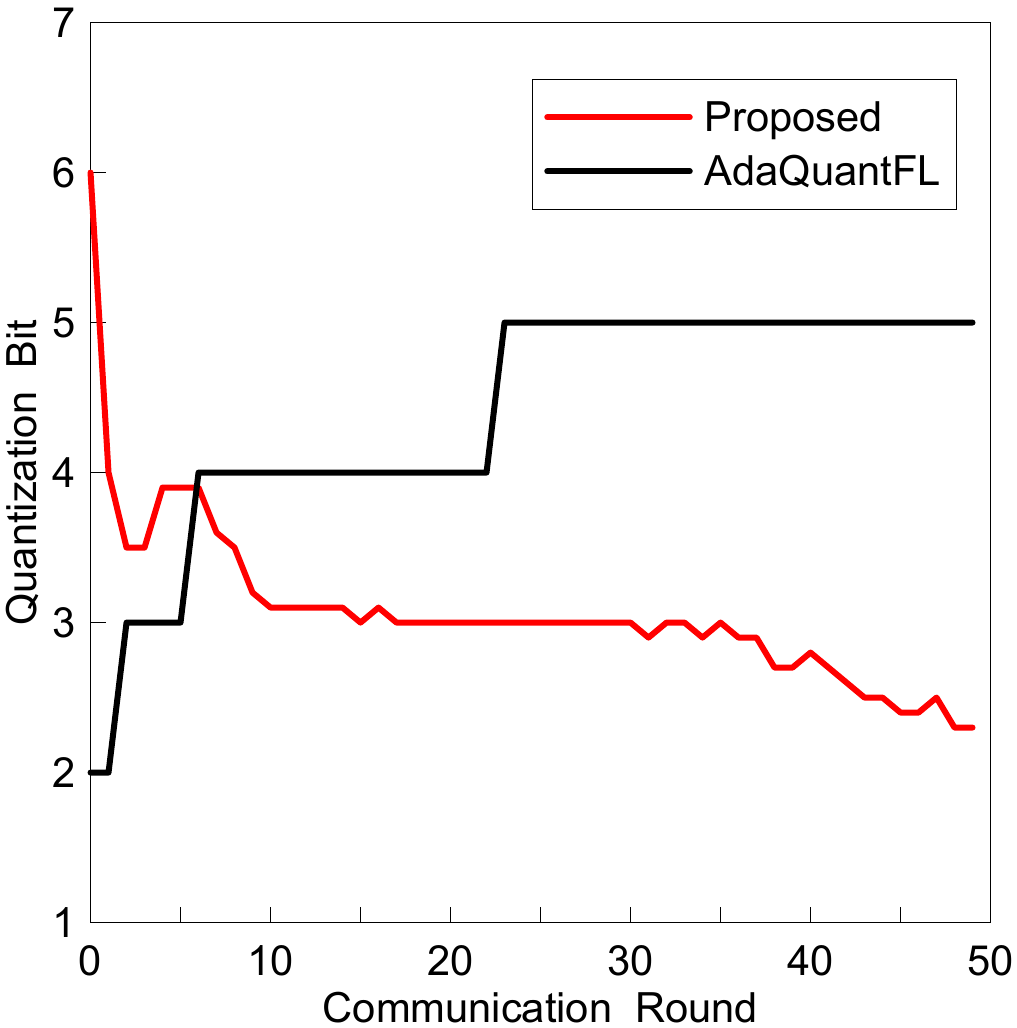}
\end{minipage}}
\quad
\subfigure[Fashion-MNIST]{
\begin{minipage}[b]{0.21\linewidth}
\includegraphics[width=1.1\linewidth]{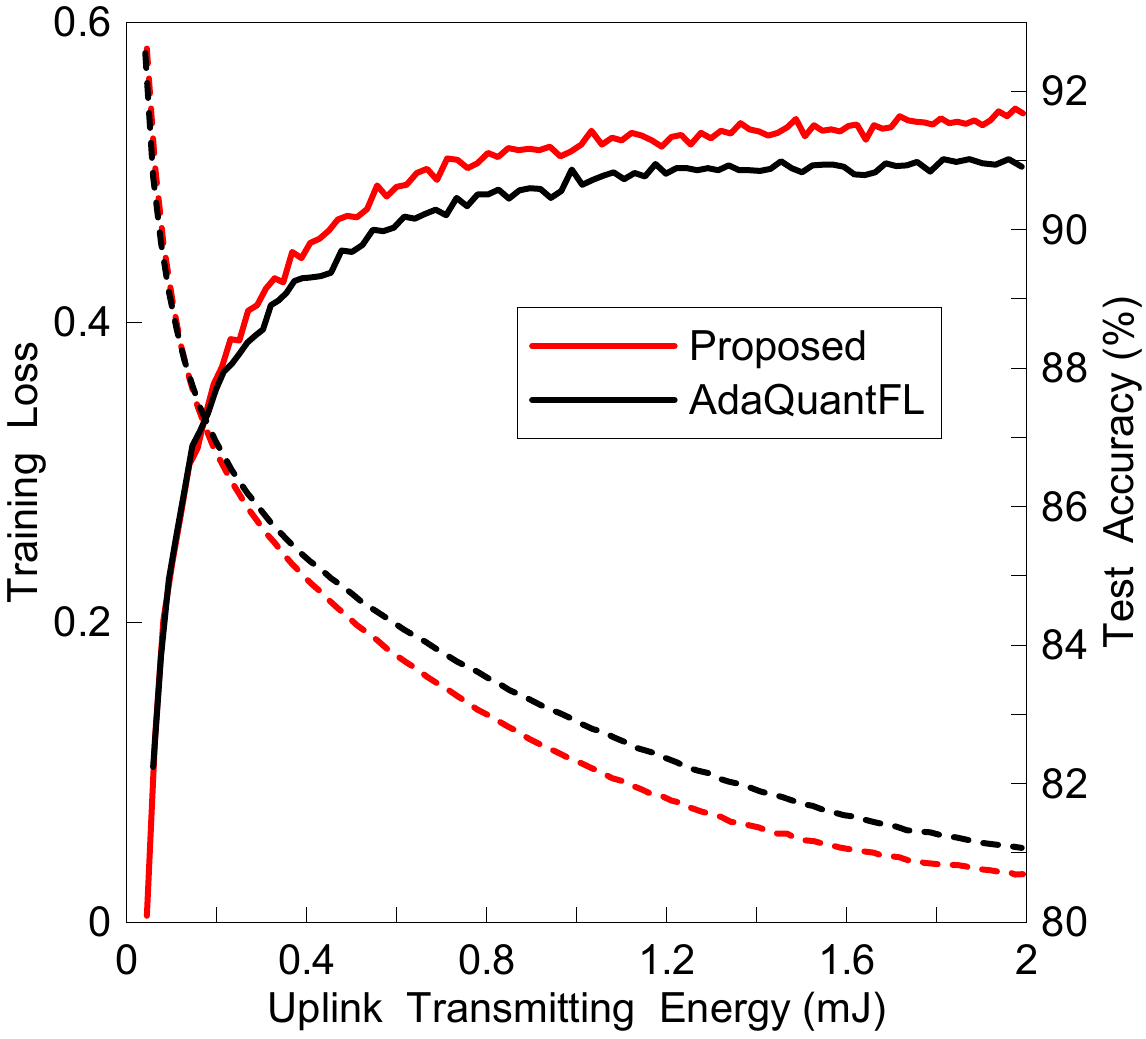}
\end{minipage}}
\quad
\subfigure[Fashion-MNIST]{
\begin{minipage}[b]{0.21\linewidth}
\includegraphics[width=1.1\linewidth]{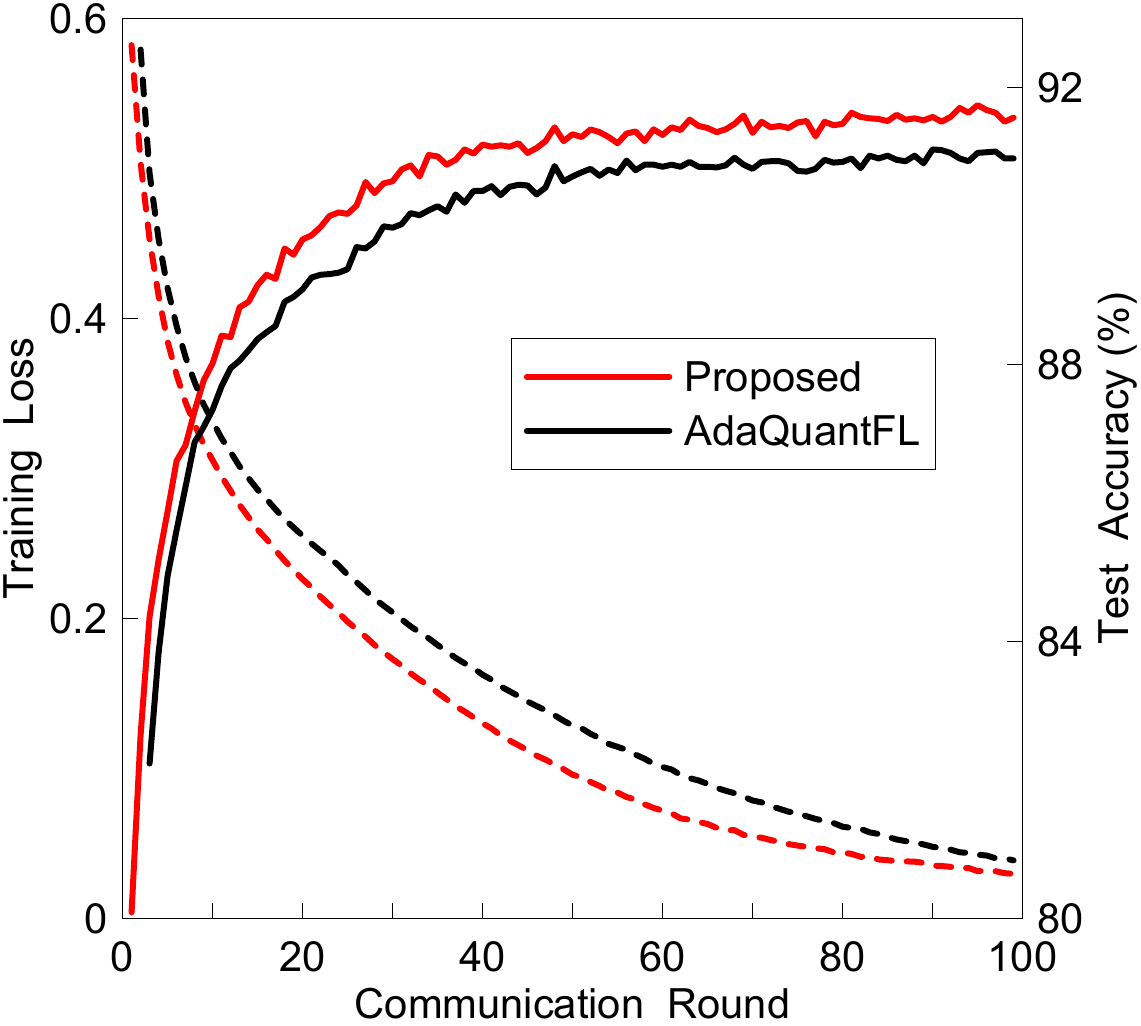}
\end{minipage}}
\quad
\subfigure[Fashion-MNIST]{
\begin{minipage}[b]{0.21\linewidth}
\includegraphics[width=1\linewidth]{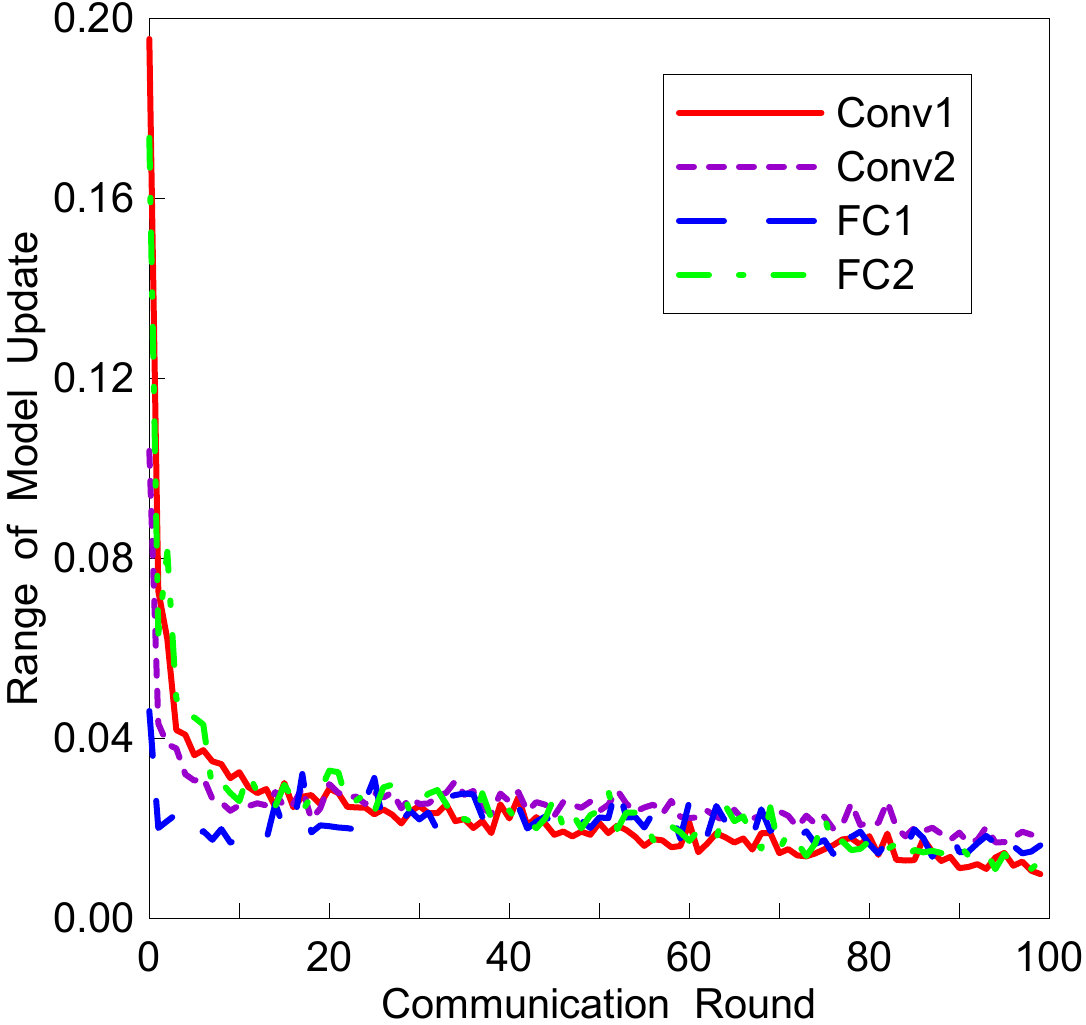}
\end{minipage}}
\quad
\subfigure[Fashion-MNIST]{
\begin{minipage}[b]{0.21\linewidth}
\includegraphics[width=1\linewidth]{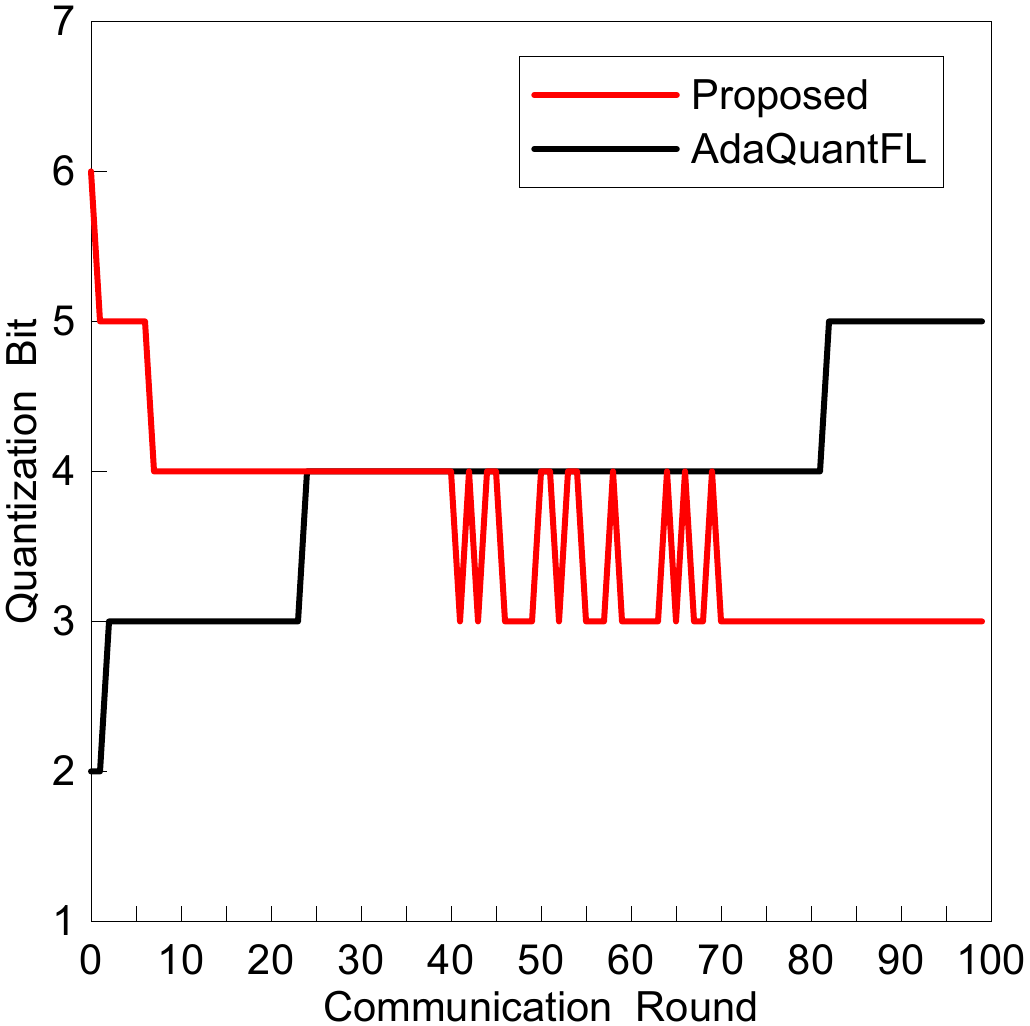}
\end{minipage}}
\quad
\subfigure[CNN on CIFAR-10]{
\begin{minipage}[b]{0.21\linewidth}
\includegraphics[width=1.1\linewidth]{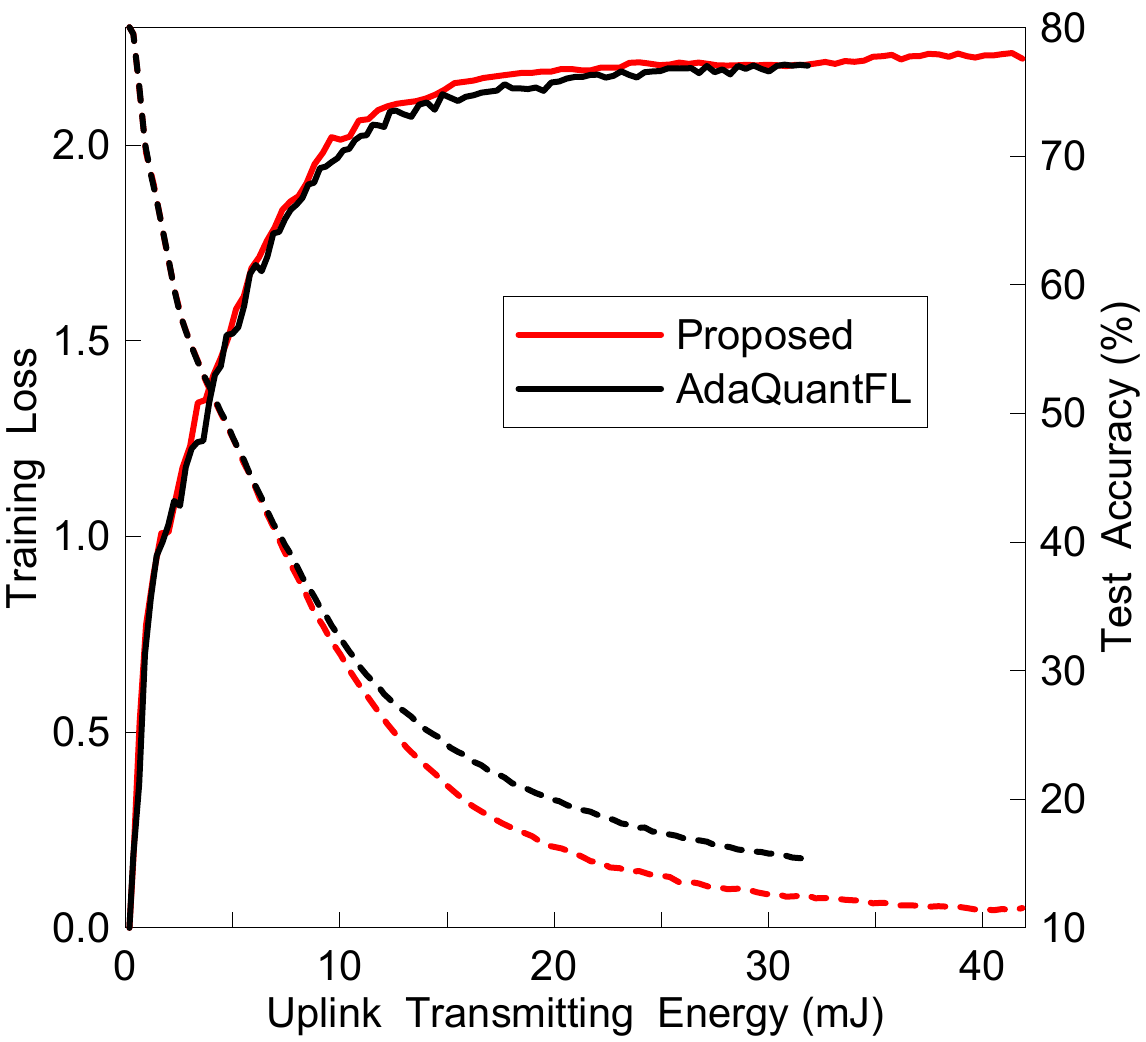}
\end{minipage}}
\quad
\subfigure[CNN on CIFAR-10]{
\begin{minipage}[b]{0.21\linewidth}
\includegraphics[width=1.1\linewidth]{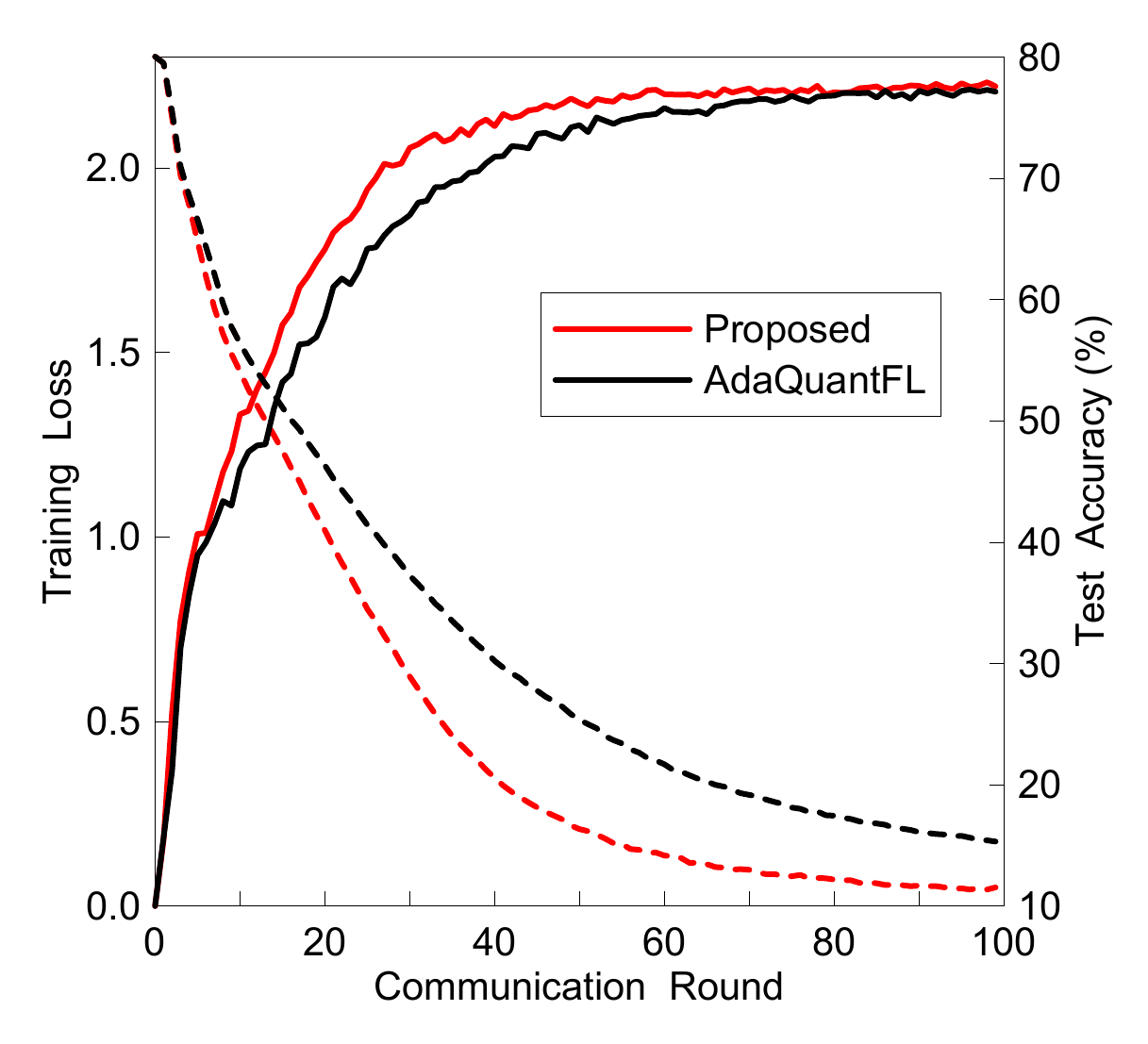}
\end{minipage}}
\quad
\subfigure[CNN on CIFAR-10]{
\begin{minipage}[b]{0.21\linewidth}
\includegraphics[width=1\linewidth]{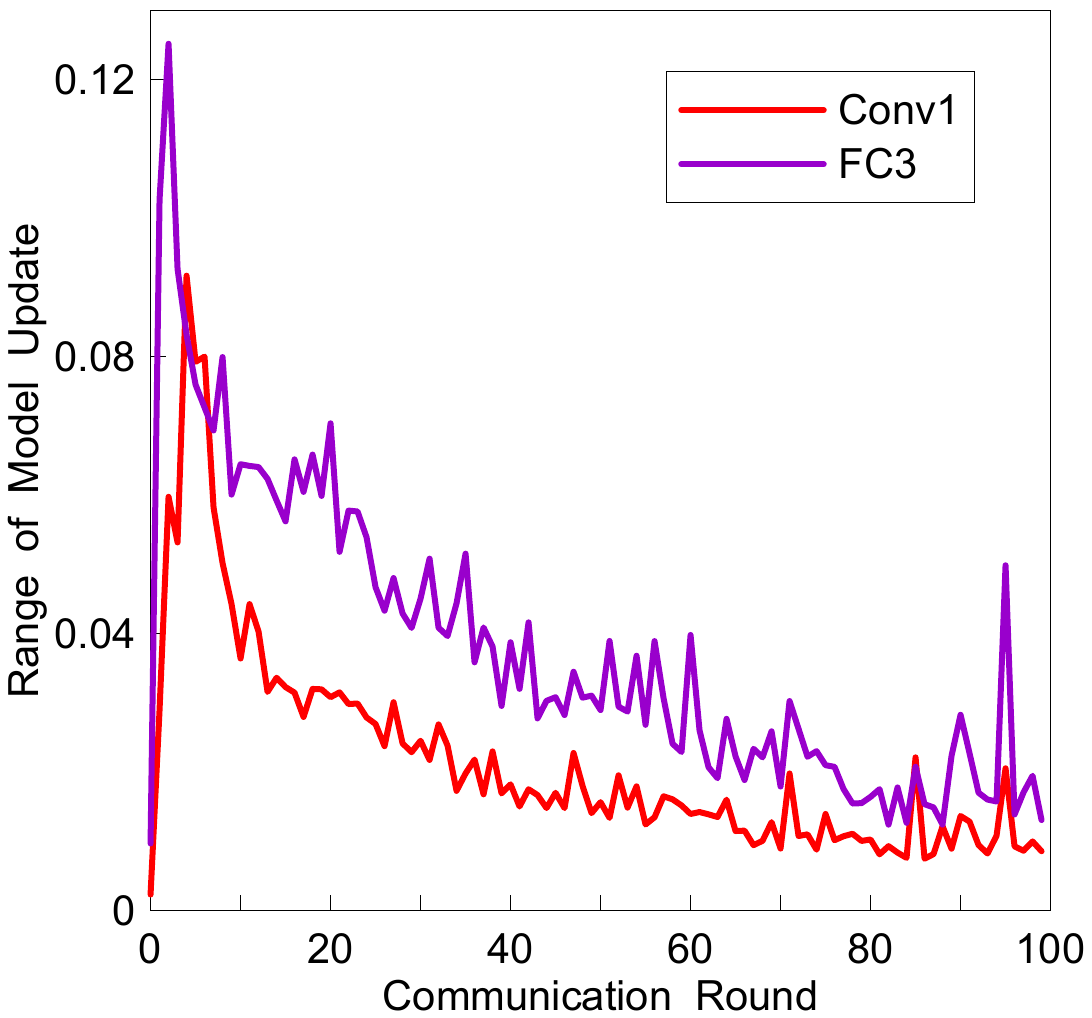}
\end{minipage}}
\quad
\subfigure[CNN on CIFAR-10]{
\begin{minipage}[b]{0.21\linewidth}
\includegraphics[width=1\linewidth]{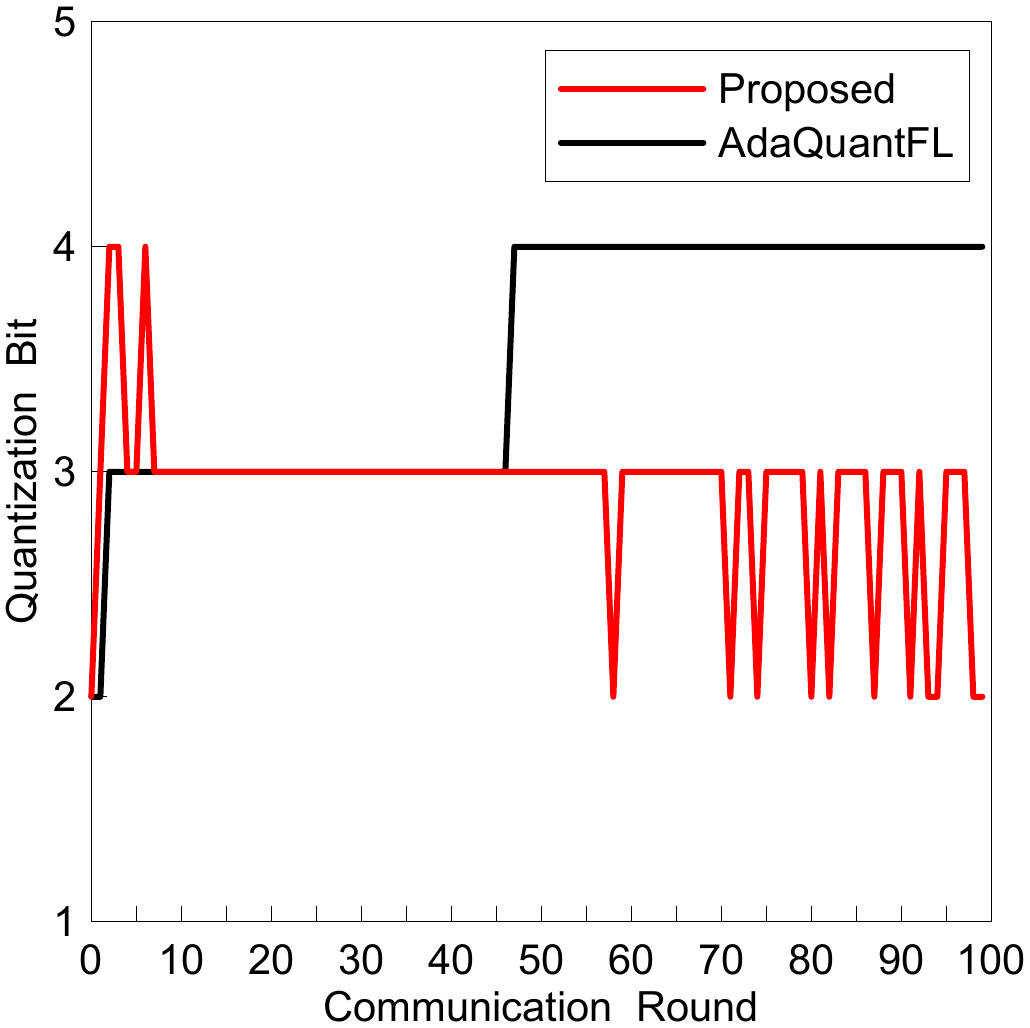}
\end{minipage}}
\quad
\subfigure[ResNet18 on CIFAR-10]{
\begin{minipage}[b]{0.21\linewidth}
\includegraphics[width=1.1\linewidth]{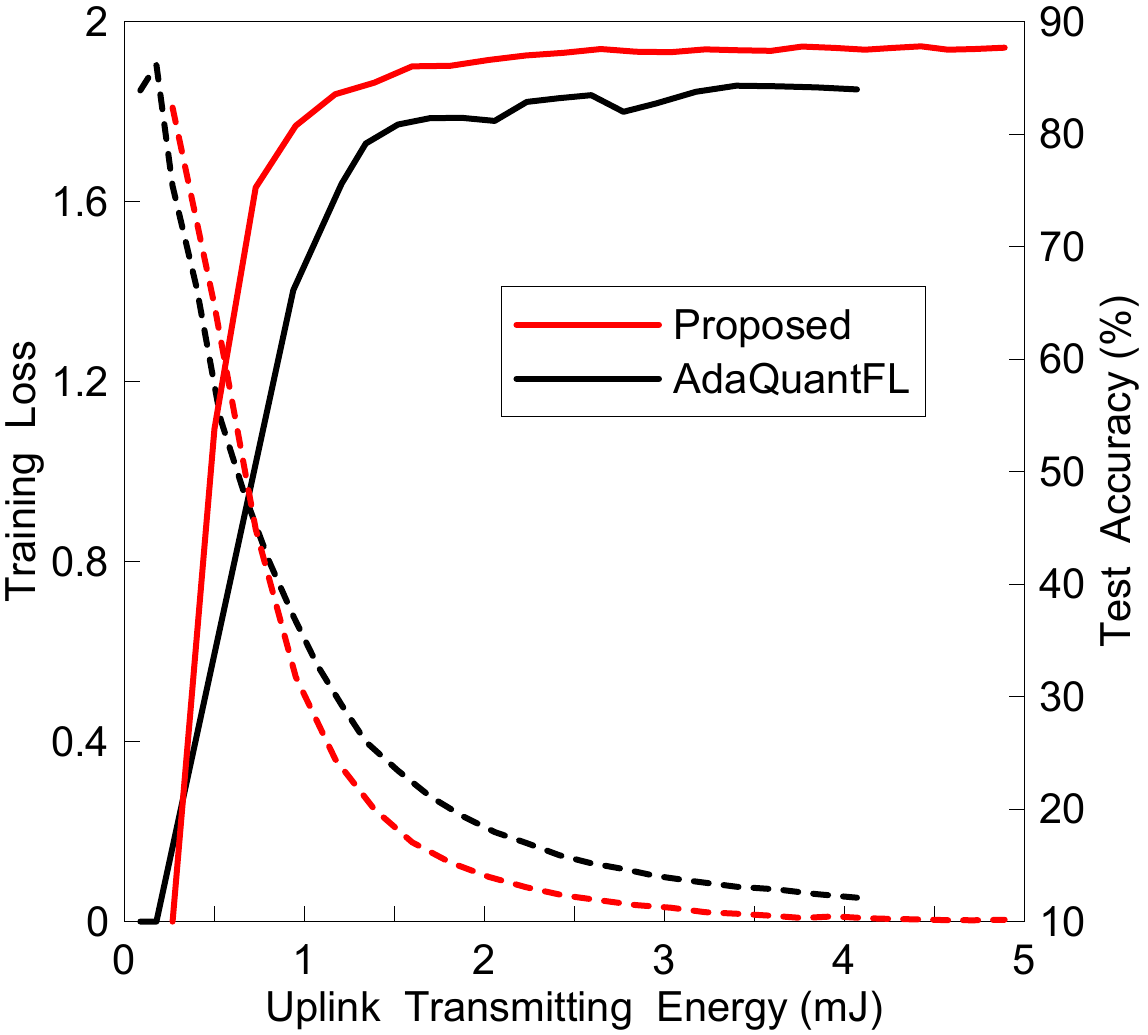}
\end{minipage}}
\quad
\subfigure[ResNet18 on CIFAR-10]{
\begin{minipage}[b]{0.21\linewidth}
\includegraphics[width=1.1\linewidth]{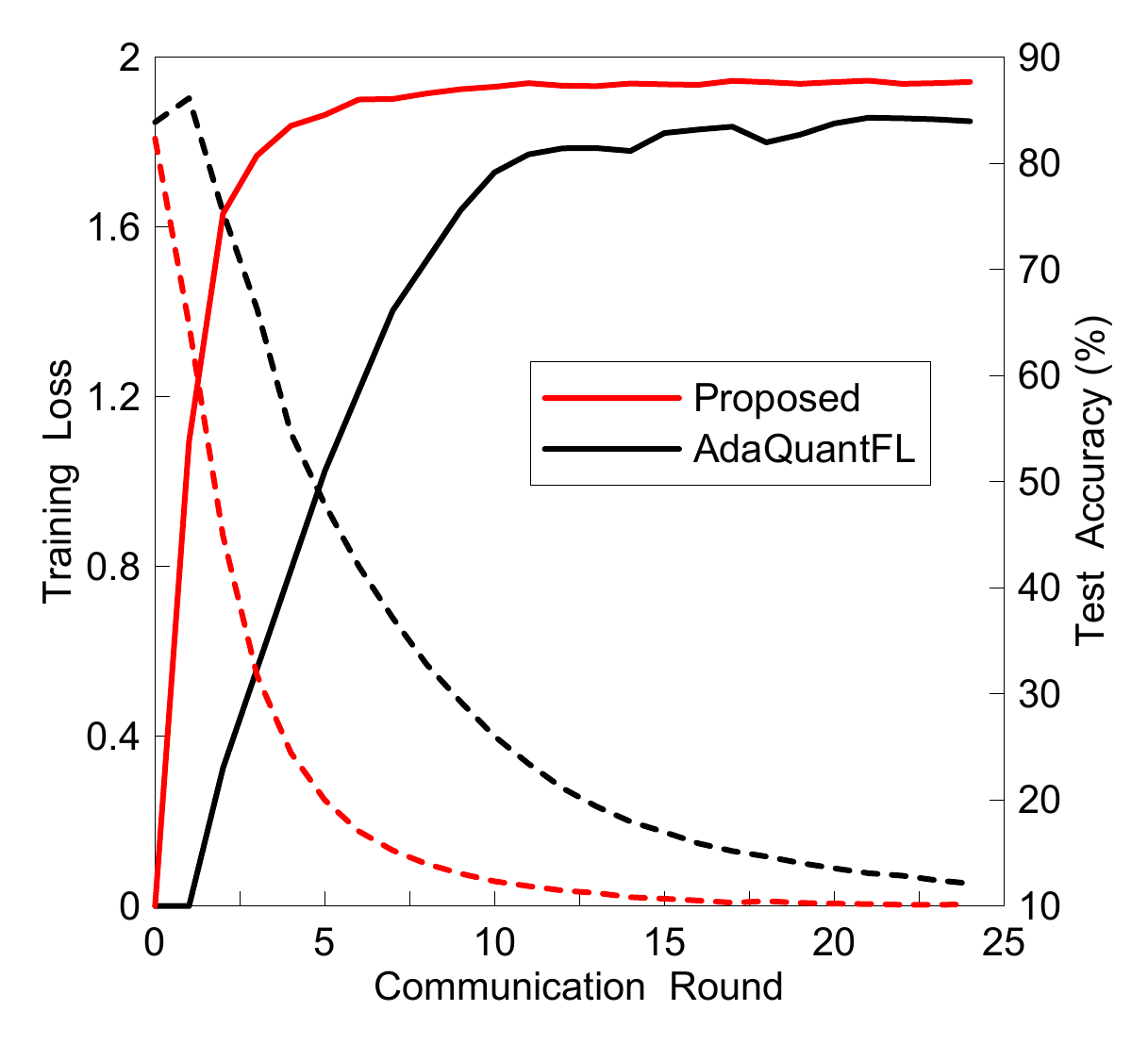}
\end{minipage}}
\quad
\subfigure[ResNet18 on CIFAR-10]{
\begin{minipage}[b]{0.21\linewidth}
\includegraphics[width=1\linewidth]{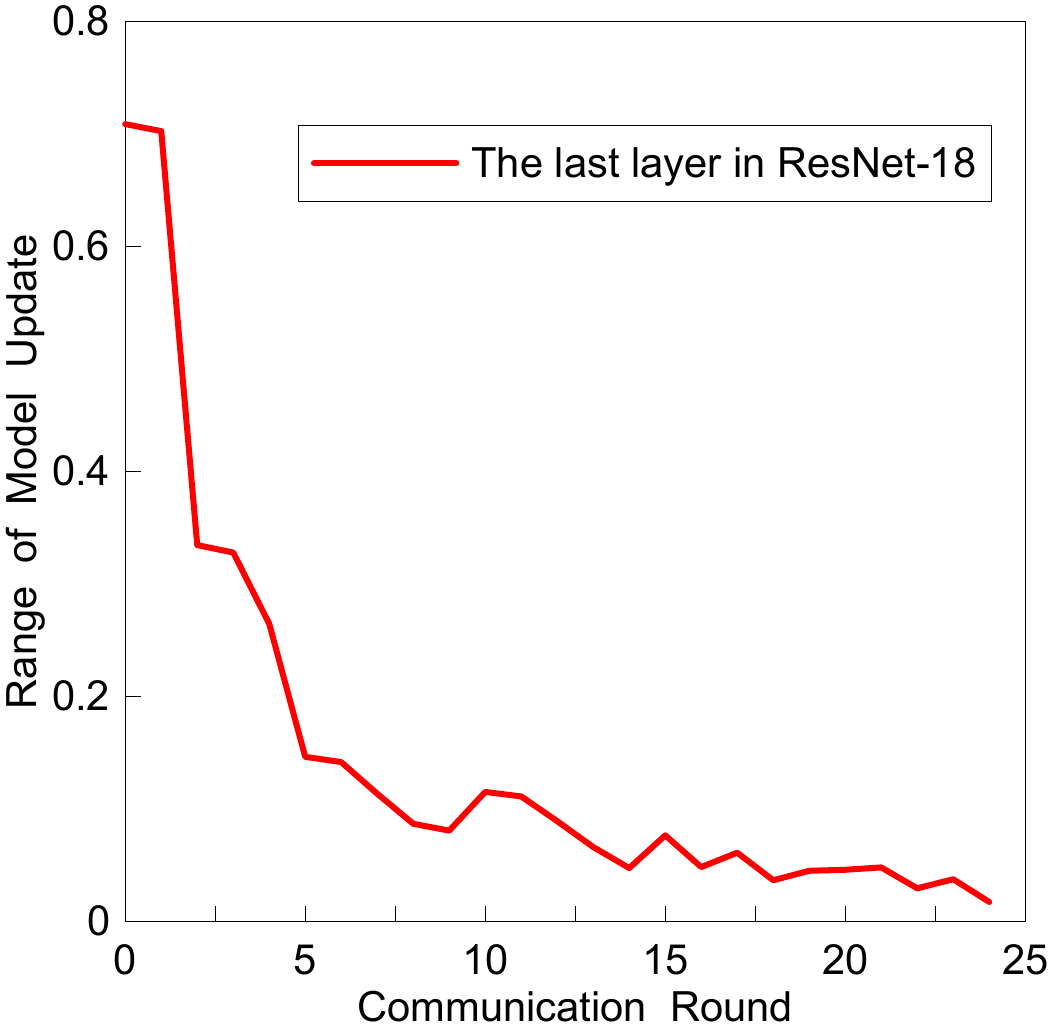}
\end{minipage}}
\quad
\subfigure[ResNet18 on CIFAR-10]{
\begin{minipage}[b]{0.21\linewidth}
\includegraphics[width=1\linewidth]{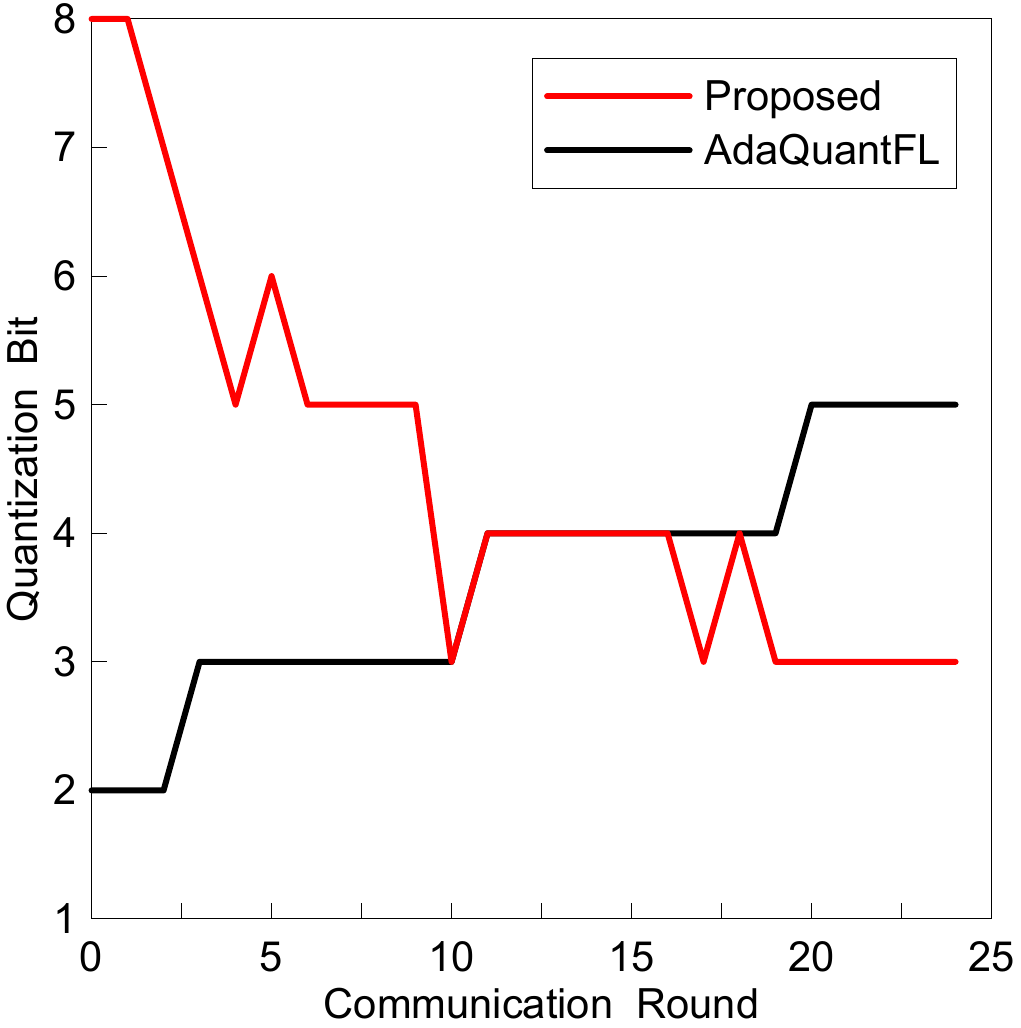}
\end{minipage}}
\caption{Results of uplink-only quantization experiments.}
\label{fig:up}
\end{figure*}

\begin{table}[t]
\caption{Uplink-only quantization}
\centering
\begin{tabular}{c|c|c|c}
\hline  
&\makecell[c]{AdaQuantFL}&\makecell[c]{Proposed}&\makecell[c]{Energy Saving}\\
\hline
\makecell[c]{LeNet-300-100 on MNIST\\ (Test Acc.=97.8\%)}&0.43mJ&\makecell[c]{0.35mJ}&\makecell[c]{18.6\%}\\
\hline  
\makecell[c]{Vanilla CNN on Fashion-MNIST\\ (Test Acc.=91.0\%)}&2.0mJ&\makecell[c]{0.71mJ}&\makecell[c]{64.5\%}\\
\hline  
\makecell[c]{7-layer CNN on CIFAR-10\\ (Test Acc.=77.0\%)}&30.4mJ&\makecell[c]{23.0mJ}&\makecell[c]{24.3\%}\\
\hline 
\makecell[c]{ResNet-18 on CIFAR-10\\(Test Acc.=84.0\%)}&3.4mJ&\makecell[c]{1.4mJ}&\makecell[c]{58.8\%}\\
\hline
\end{tabular}
\label{tab:uplink}
\end{table}

In this sub-section, we conduct experiments with adaptive quantization only in the uplink, and no quantization in the downlink. Many works have focused on uplink quantization, and we compare the proposed scheme with the state-of-the-art AdaQuantFL, which uses increasing-trend quantization\cite{jhunjhunwala2021adaptive}.

Fig.\ref{fig:up} provides detailed results of the energy efficiency, convergence rate, range of parameters, and bit-changing curves in the experiments. In Fig.~\ref{fig:up}, (a)--(d) in the first row are the results for LeNet-300-100 on MNIST, (e)--(h) in the second row are the results for vanilla CNN on Fashion-MNIST, (i)--(l) in the third row are the results for 7-layer CNN on CIFAR-10, and (m)--(p) in the last row are the results for ResNet-18 on CIFAR-10. The results for energy efficiency, in Fig.\ref{fig:up}, (a),(e),(i), and (m) show that the descending quantization can achieve the same test accuracy or training loss as AdQuantFL but with lower energy consumption. Take the experiment of the vanilla CNN on Fashion-MNIST as an example. To achieve the test accuracy of 91.0\%, AdaQuantFL consumes 2.0mJ of transmitting energy, while the proposed scheme only consumes 0.71mJ, indicating an energy saving ratio of 64.5\%. In Fig.\ref{fig:up}, (b),(f),(j), and (n) demonstrate that the proposed scheme can converge faster to the same test accuracy than AdaQuantFL. The main reason is that a longer bit-length at the early training stage can help to enhance the convergence speed. More detailed explanation is provided in \cite{qu2022feddq}. Fig.\ref{fig:up} (c),(g),(k), and (o) show that the range of local model updates always shows a decreasing trend because the model stabilizes as training goes on and the model updates become smaller and smaller. Finally, the quantization level curves in subfigures (d), (h), (l), and (p) show an increasing trend for AdaQuantFL and a decreasing trend for the proposed scheme. This agrees with the theoretical analysis. A detailed comparison of the four experiments can be found in Table~\ref{tab:uplink}.

\subsection{Downlink Only Design}

\begin{figure*}
\centering
\subfigure[MNIST]{
\begin{minipage}[b]{0.21\linewidth}
\includegraphics[width=1.1\linewidth]{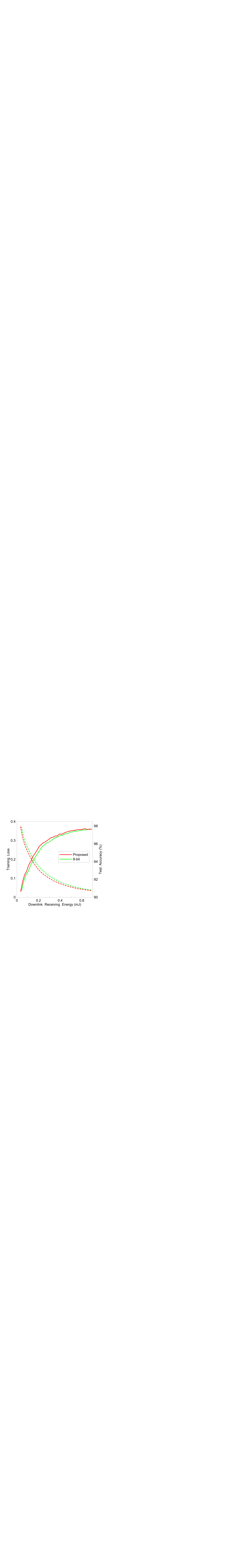}
\end{minipage}}
\quad
\subfigure[MNIST]{
\begin{minipage}[b]{0.21\linewidth}
\includegraphics[width=1.1\linewidth]{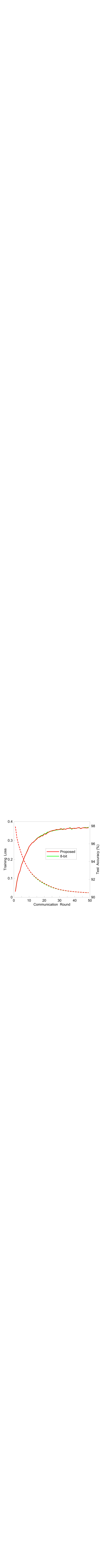}
\end{minipage}}
\quad
\subfigure[MNIST]{
\begin{minipage}[b]{0.21\linewidth}
\includegraphics[width=1\linewidth]{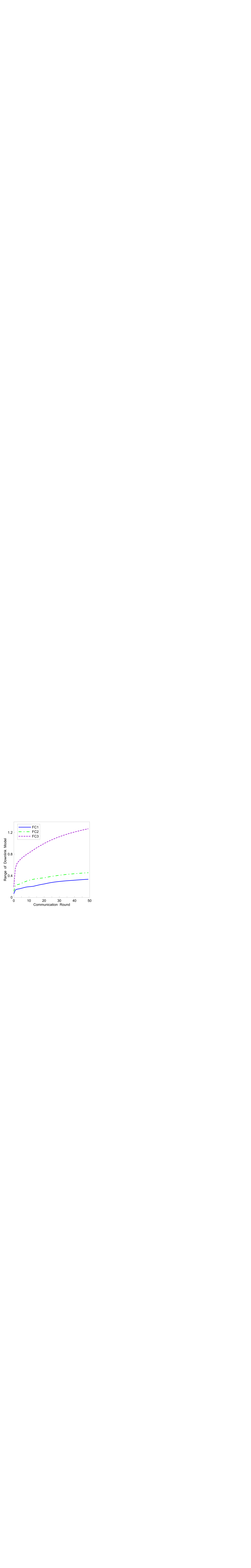}
\end{minipage}}
\quad
\subfigure[MNIST]{
\begin{minipage}[b]{0.21\linewidth}
\includegraphics[width=1\linewidth]{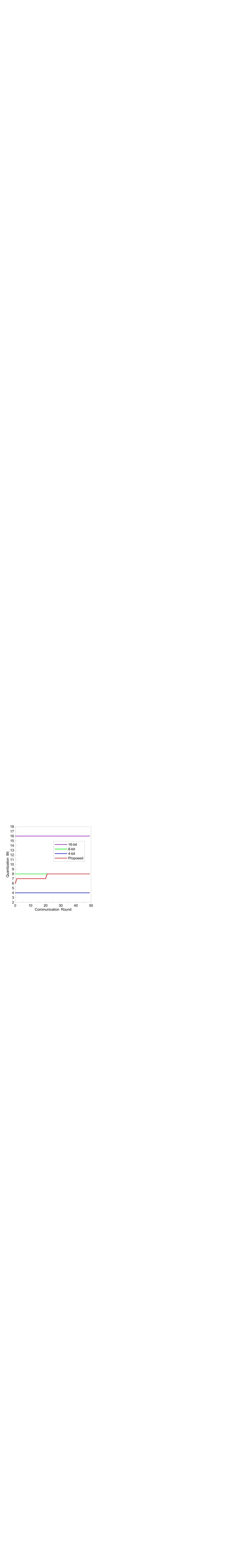}
\end{minipage}}
\quad
\subfigure[Fashion-MNIST]{
\begin{minipage}[b]{0.21\linewidth}
\includegraphics[width=1.1\linewidth]{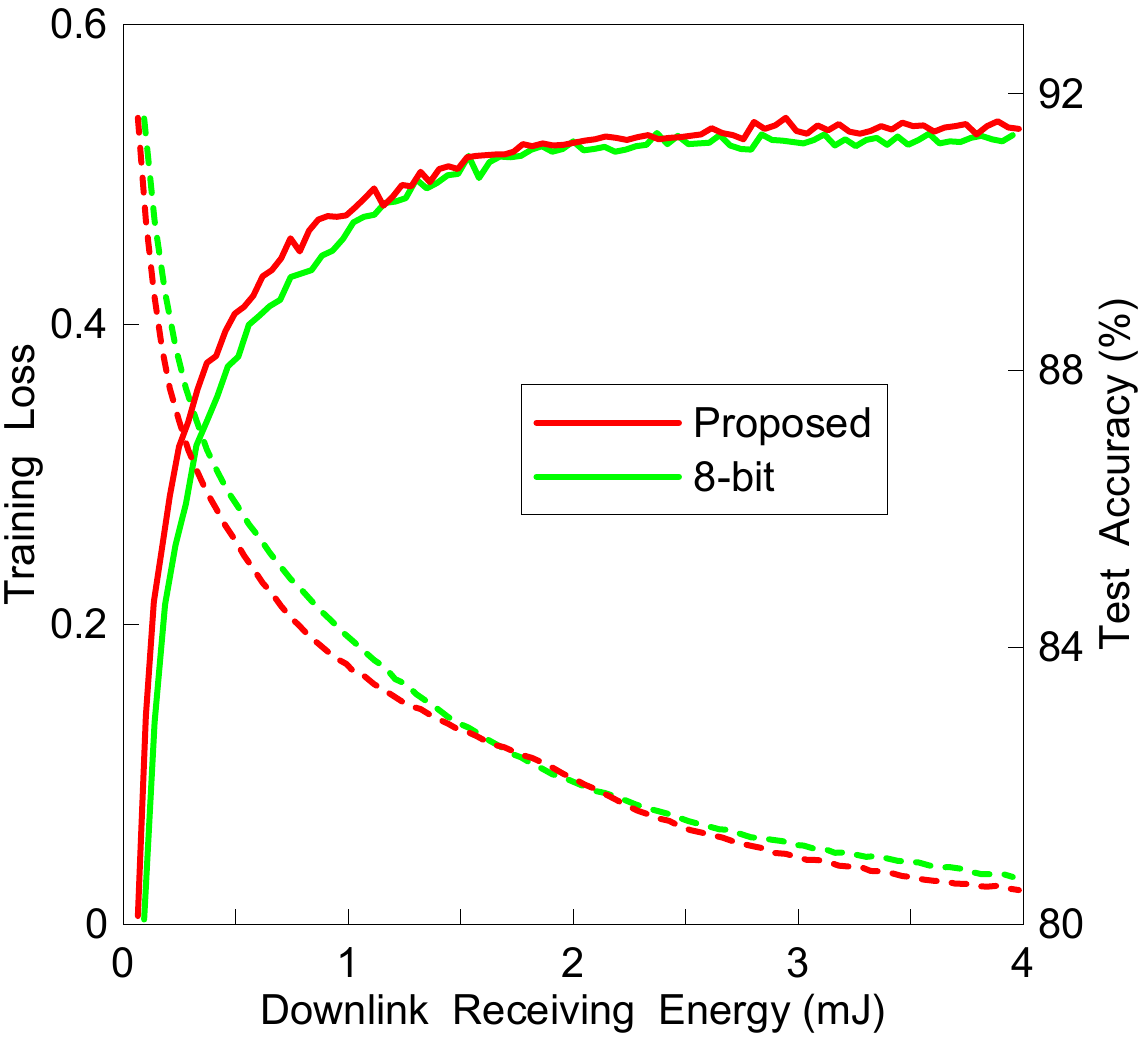}
\end{minipage}}
\quad
\subfigure[Fashion-MNIST]{
\begin{minipage}[b]{0.21\linewidth}
\includegraphics[width=1.1\linewidth]{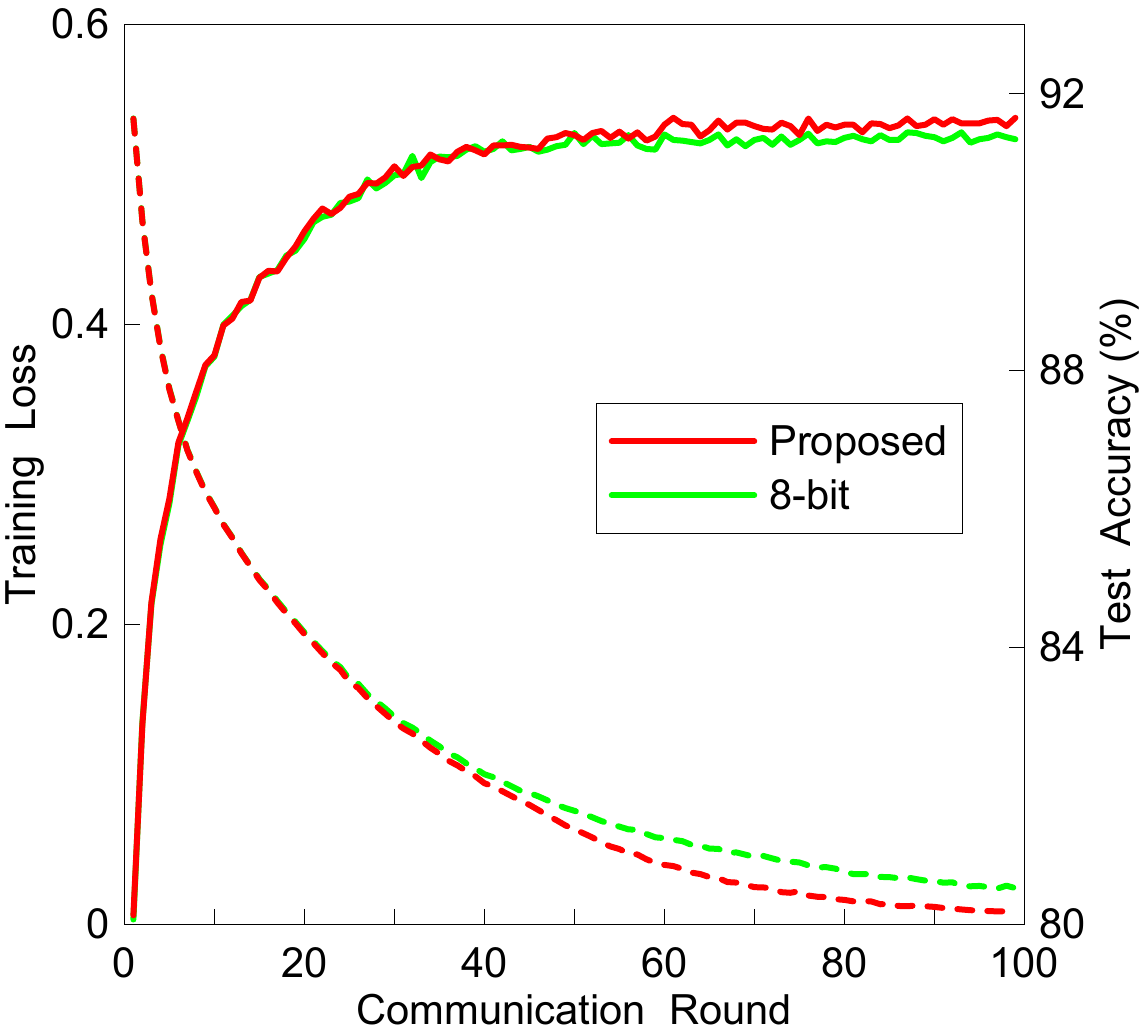}
\end{minipage}}
\quad
\subfigure[Fashion-MNIST]{
\begin{minipage}[b]{0.21\linewidth}
\includegraphics[width=1\linewidth]{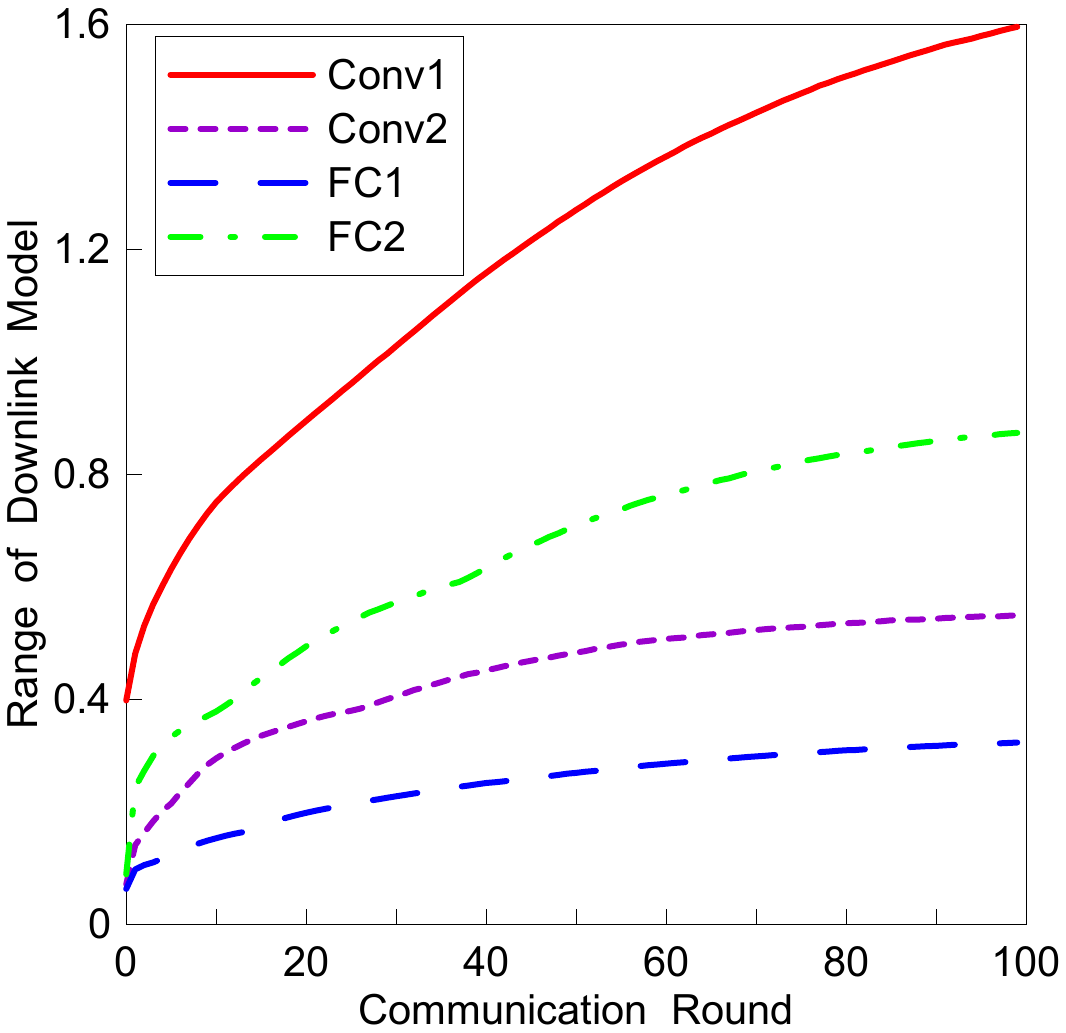}
\end{minipage}}
\quad
\subfigure[Fashion-MNIST]{
\begin{minipage}[b]{0.21\linewidth}
\includegraphics[width=1\linewidth]{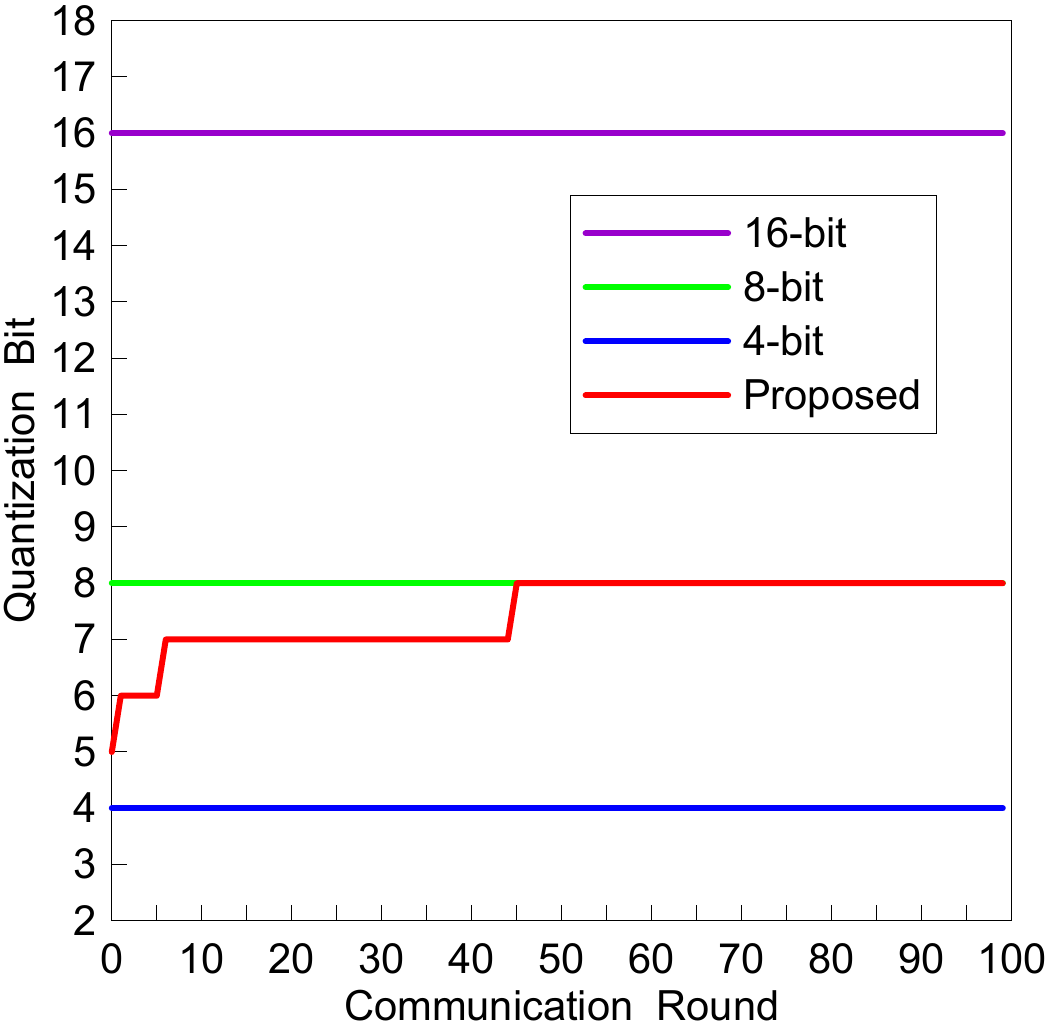}
\end{minipage}}
\quad
\subfigure[CNN on CIFAR-10]{
\begin{minipage}[b]{0.21\linewidth}
\includegraphics[width=1.1\linewidth]{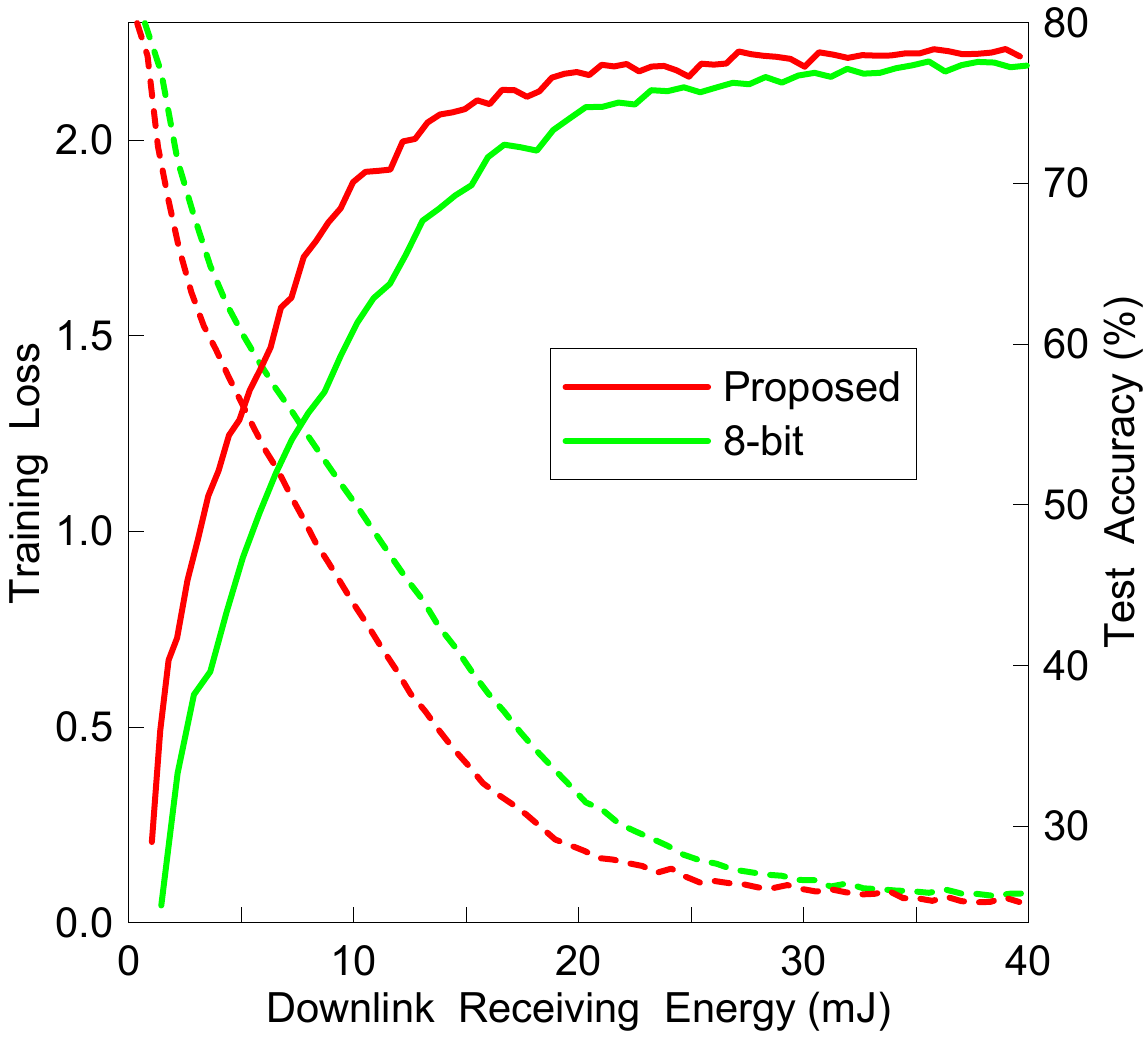}
\end{minipage}}
\quad
\subfigure[CNN on CIFAR-10]{
\begin{minipage}[b]{0.21\linewidth}
\includegraphics[width=1.1\linewidth]{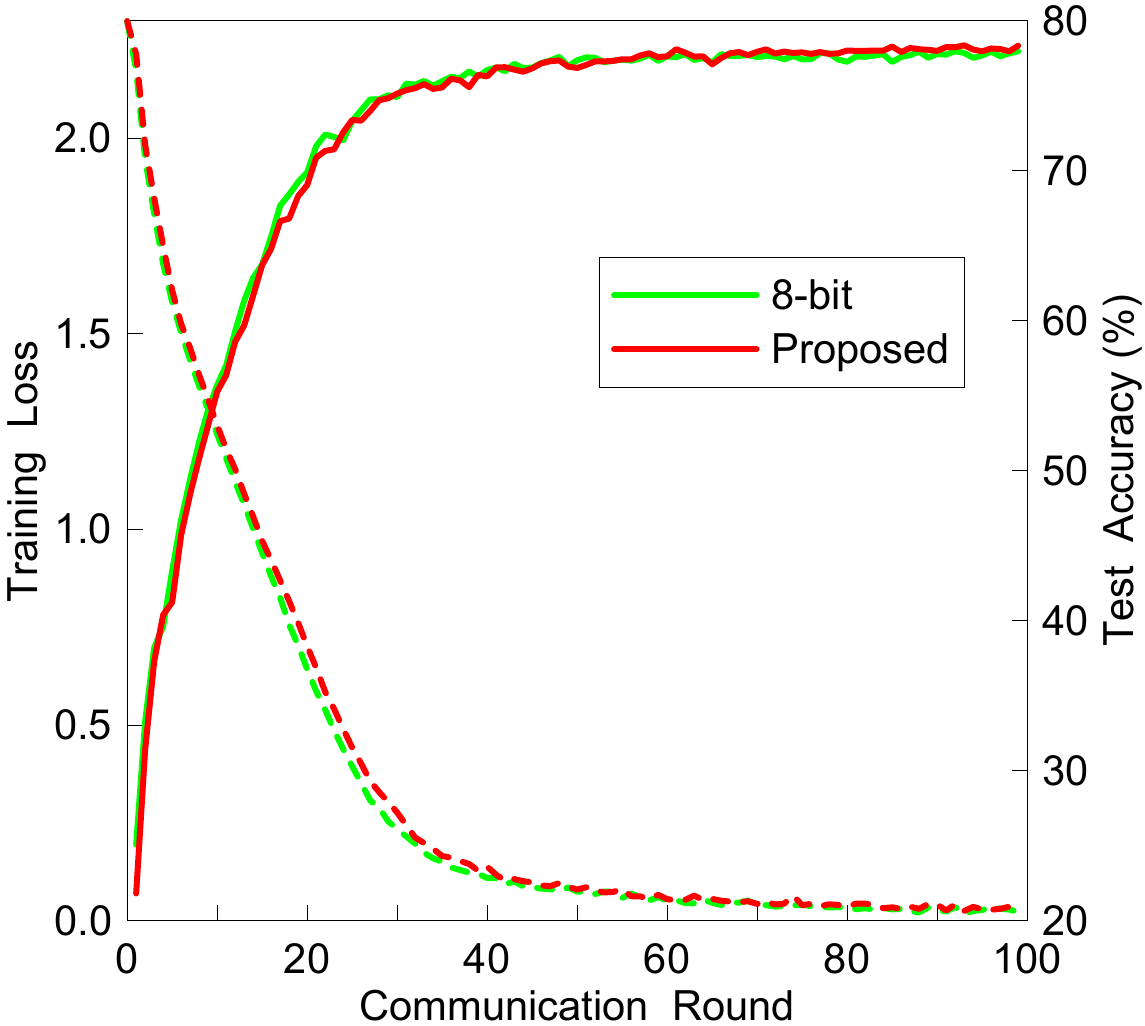}
\end{minipage}}
\quad
\subfigure[CNN on CIFAR-10]{
\begin{minipage}[b]{0.21\linewidth}
\includegraphics[width=1\linewidth]{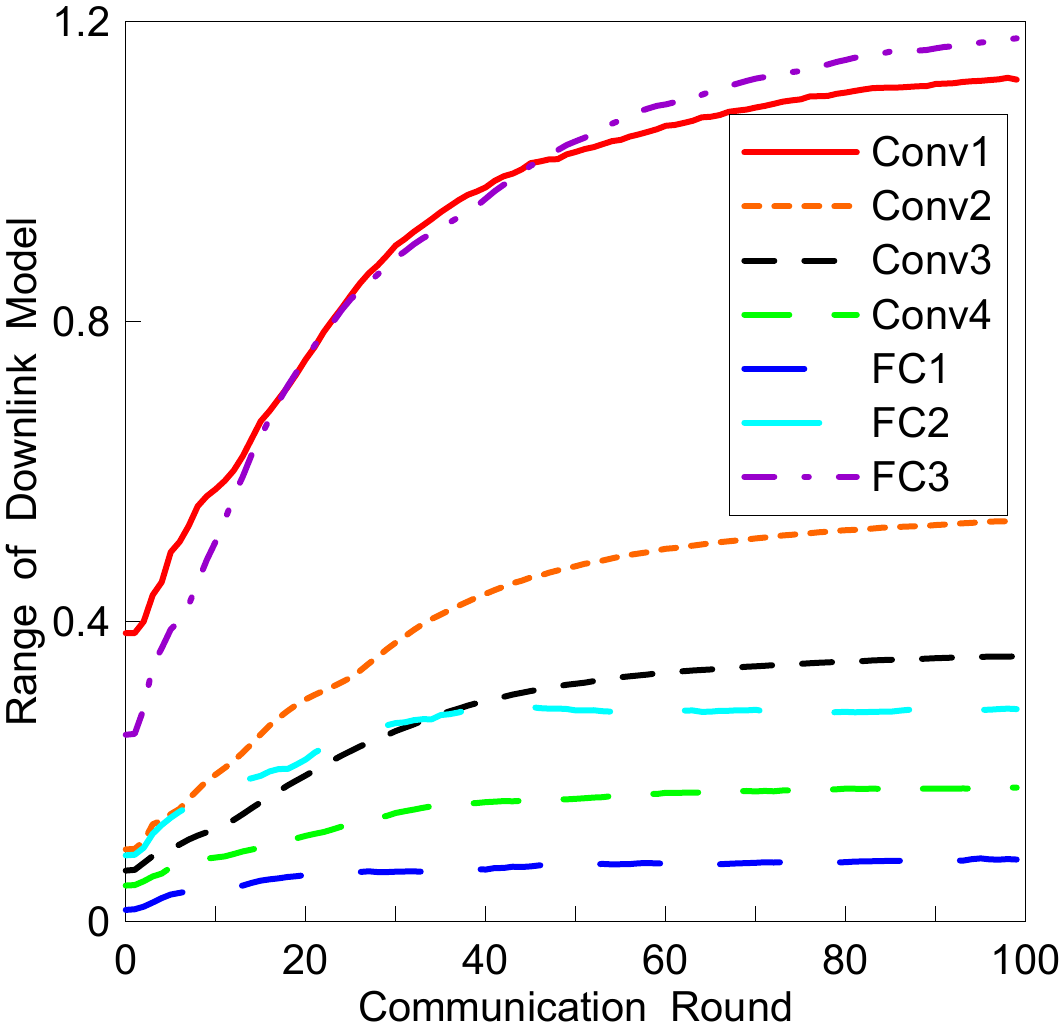}
\end{minipage}}
\quad
\subfigure[CNN on CIFAR-10]{
\begin{minipage}[b]{0.21\linewidth}
\includegraphics[width=1\linewidth]{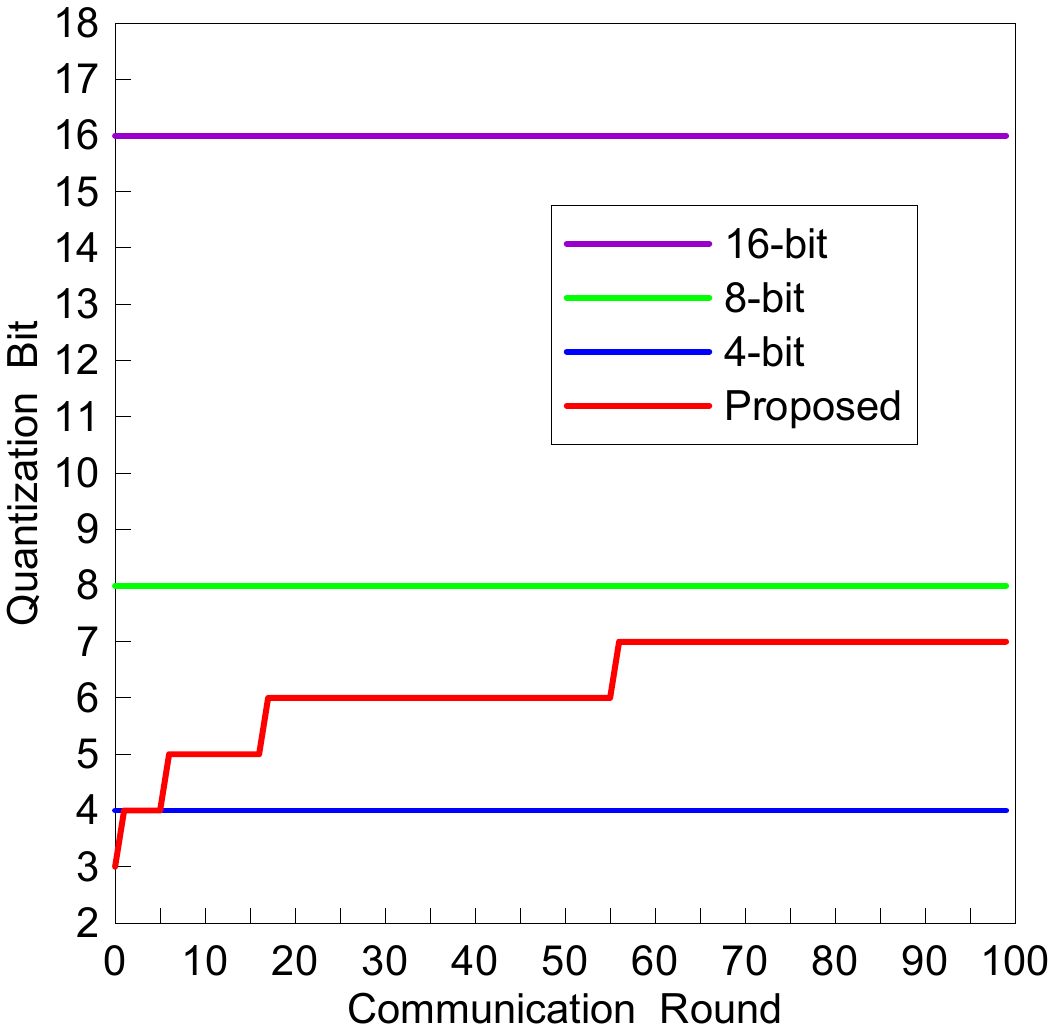}
\end{minipage}}
\quad
\subfigure[ResNet18 on CIFAR-10]{
\begin{minipage}[b]{0.21\linewidth}
\includegraphics[width=1.1\linewidth]{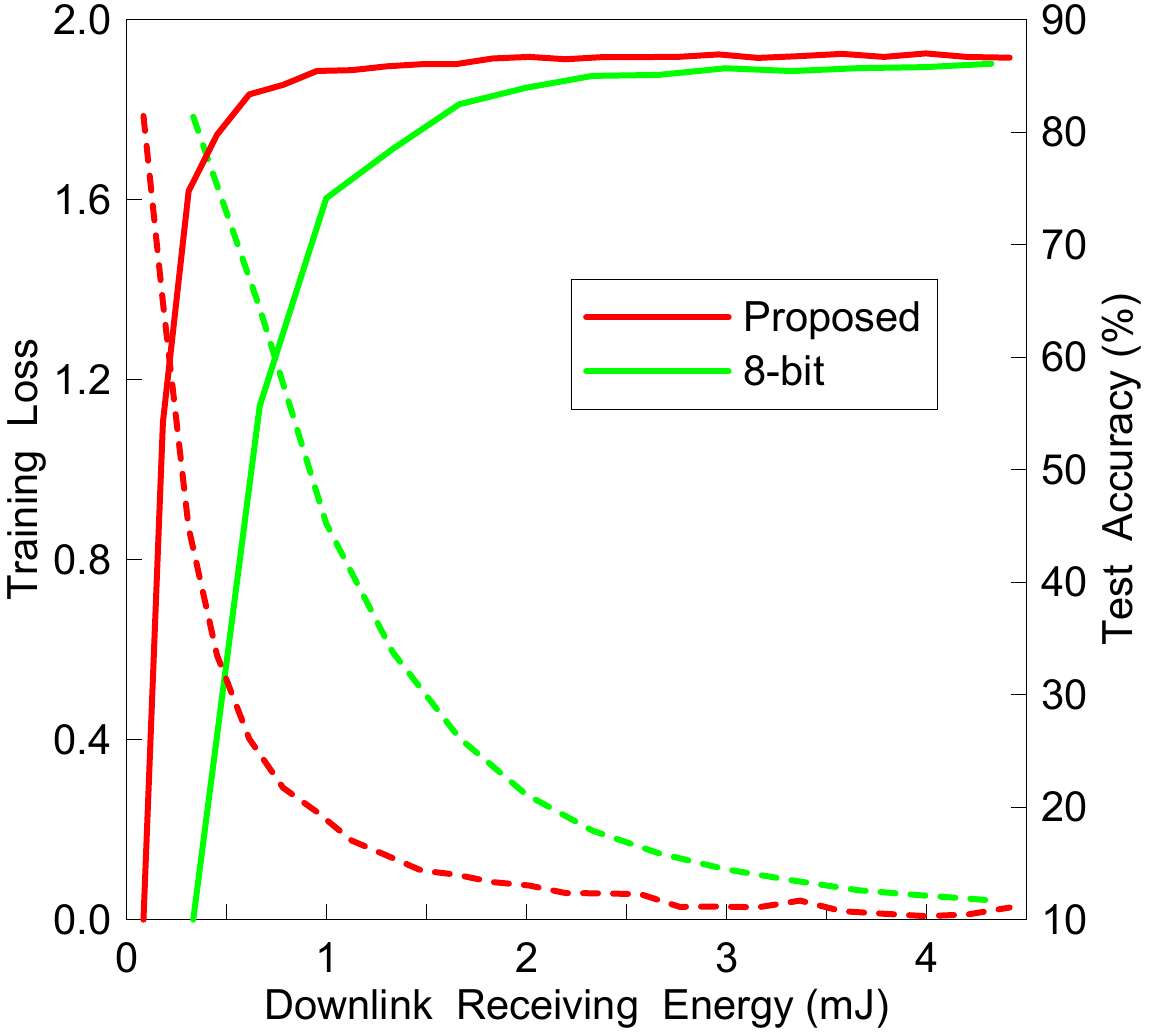}
\end{minipage}}
\quad
\subfigure[ResNet18 on CIFAR-10]{
\begin{minipage}[b]{0.21\linewidth}
\includegraphics[width=1.1\linewidth]{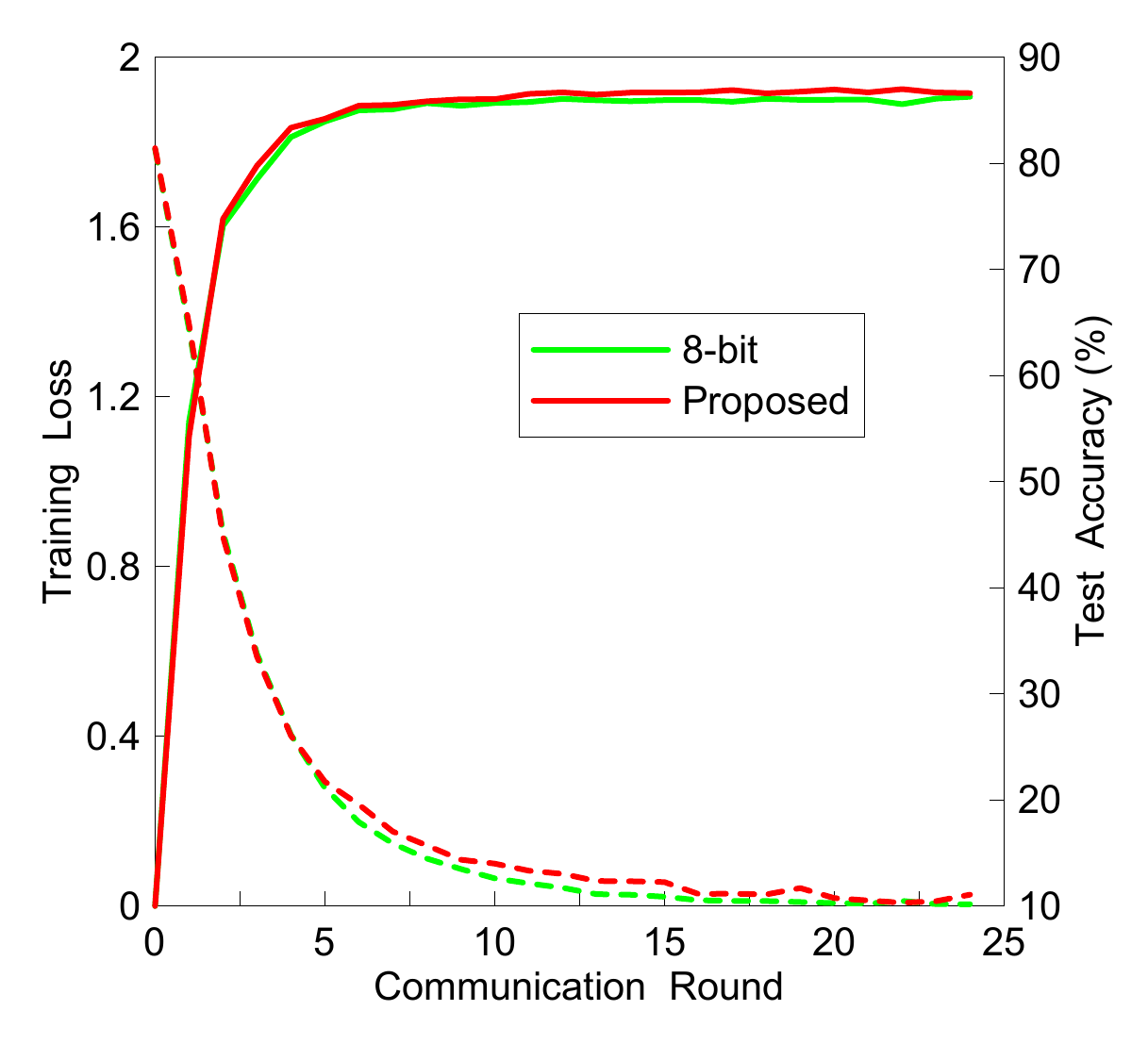}
\end{minipage}}
\quad
\subfigure[ResNet18 on CIFAR-10]{
\begin{minipage}[b]{0.21\linewidth}
\includegraphics[width=1\linewidth]{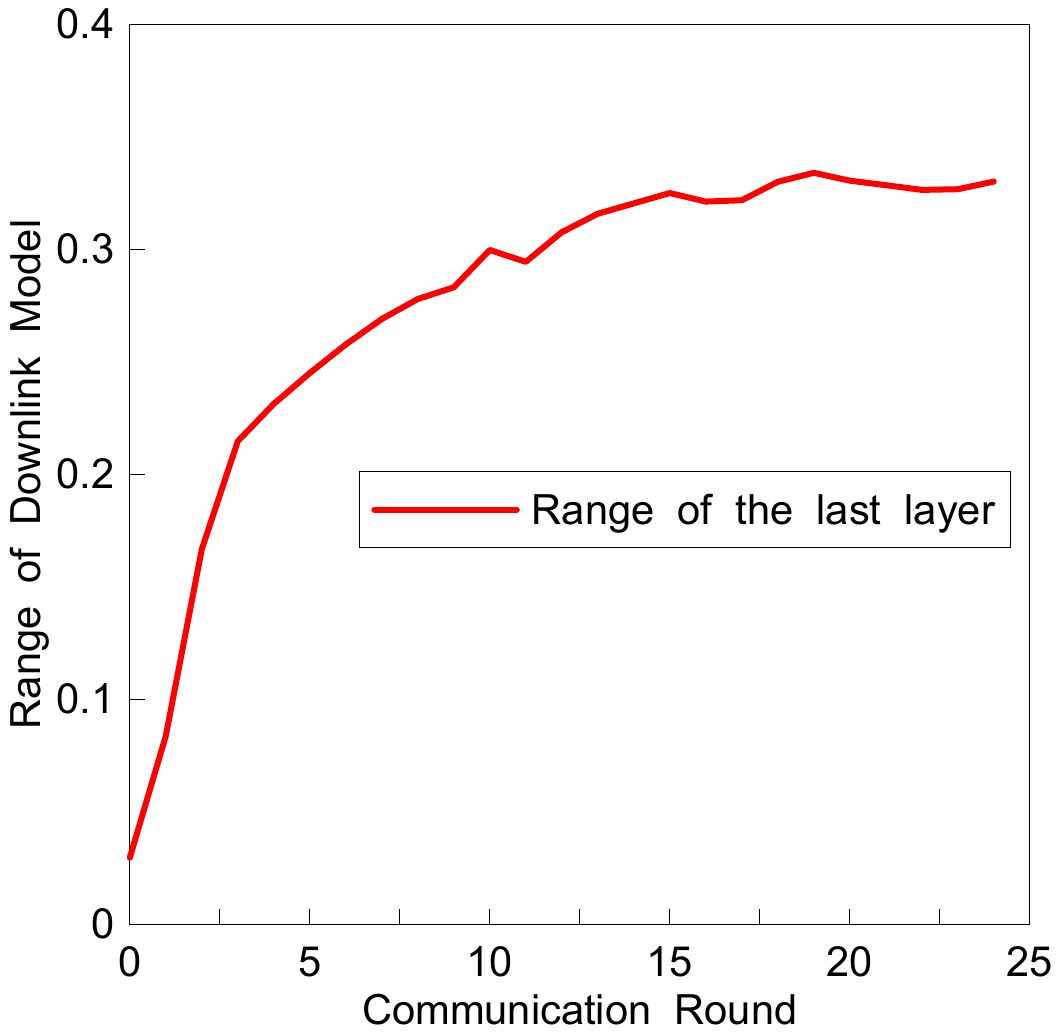}
\end{minipage}}
\quad
\subfigure[ResNet18 on CIFAR-10]{
\begin{minipage}[b]{0.21\linewidth}
\includegraphics[width=1\linewidth]{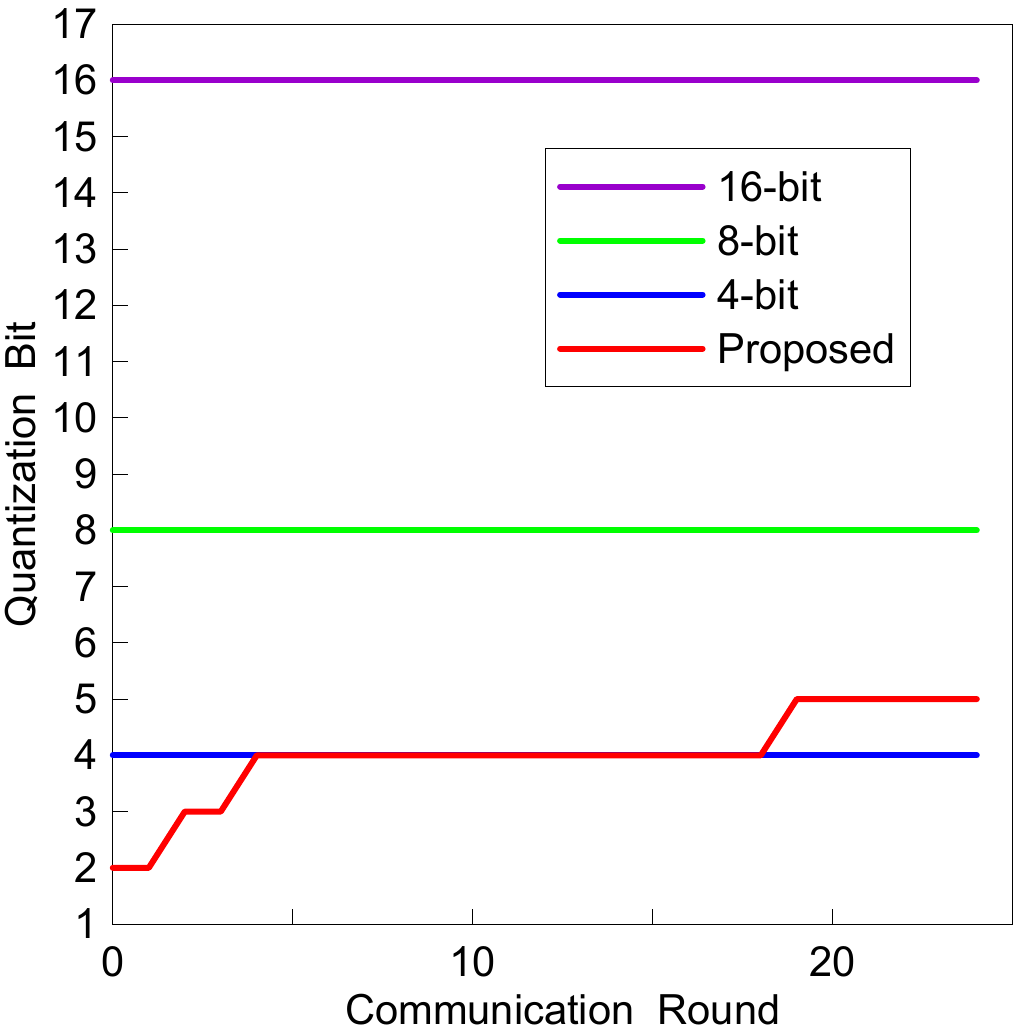}
\end{minipage}}
\caption{Results for downlink-only quantization experiments.}
\label{fig:down}
\end{figure*}

\begin{table}[t]
\caption{Downlink-only quantization}
\centering
\begin{tabular}{c|c|c|c}
\hline  
&\makecell[c]{Fixed 8-bit}&\makecell[c]{Proposed}&\makecell[c]{Energy Saving}\\
\hline
\makecell[c]{LeNet-300-100 on MNIST\\ (Test Acc.=97.5\%)}&0.60mJ&\makecell[c]{0.52mJ}&\makecell[c]{13.3\%}\\
\hline  
\makecell[c]{Vanilla CNN on Fashion-MNIST\\ (Test Acc.=91.0\%)}&2.10mJ&\makecell[c]{1.73mJ}&\makecell[c]{17.6\%}\\
\hline  
\makecell[c]{7-layer CNN on CIFAR-10\\ (Test Acc.=77.3\%)}&40.0mJ&\makecell[c]{21.6mJ}&\makecell[c]{46.0\%}\\
\hline 
\makecell[c]{ResNet-18 on CIFAR-10\\(Test Acc.=86.0\%)}&4.32mJ&\makecell[c]{1.47mJ}&\makecell[c]{66.0\%}\\
\hline
\end{tabular}
\label{tab:downlink}
\end{table}

In this section, we conduct experiments with adaptive quantization only in the downlink, and no quantization in the uplink. Few prior works have focused on downlink quantization. We compare the proposed increasing-trend quantization with the optimum fixed-bit quantization, i.e., 8-bit quantization, in the downlink.

Fig.\ref{fig:down} presents the results for LeNet-300-100 on MNIST, vanilla CNN on Fashion-MNIST, 7-layer CNN on CIFAR-10, and ResNet-18 on CIFAR-10 row by row. The first column of Fig.\ref{fig:down} shows that, for all experiments, the proposed quantization scheme can always achieve the same test accuracy or same training loss at lower downlink energy consumption compared with the optimum fixed-bit quantization scheme. Taking the vanilla CNN on Fashion-MNIST experiment as an example, to achieve the test accuracy of 91.0\%, the 8-bit quantization consumes 2.10mJ while the proposed scheme only consumes 1.73mJ, indicating an energy saving ratio of 17.6\%. As shown in the second column, the proposed method has a similar convergence speed to the fixed 8-bit quantization. So the proposed method saves downlink energy without any sacrifice of learning performance. The third column of Fig.\ref{fig:down} shows that different from the range of local model updates, the range of the global model weights shows an increasing trend. This is because as training goes on, the wider range of model parameters provides a better generalization capacity to capture different training data. The quantization level in the downlink is also designed based on the range of the global model, so it shows an increasing trend with training as shown in the last column of Fig.~\ref{fig:down}. The detailed comparison of the four experiments can be found in Table~\ref{tab:downlink}.

\section{Conclusion}
\label{Conclusion}
In this paper, we proposed a joint uplink and downlink adaptive quantization scheme to \textcolor{black}{save the communication overhead for} FL systems. For that purpose, we first \textcolor{black}{derived the convergence bound of FL with both uplink and downlink quantization errors. Then, we} formulated an optimization problem that is to achieve the best learning performance under energy constraints by tuning uplink and downlink quantization levels. The results showed that the optimal quantization level depends on the range of communicated parameters. Additionally, \textcolor{black}{it is} observed that the range of local model updates in the uplink has a decreasing trend, while the range of the global model weights has an increasing trend. Accordingly, we propose to adopt decreasing-trend quantization in the uplink and increasing-trend quantization in the downlink. Comprehensive experiments in three scenarios: joint uplink and downlink quantization, uplink-only quantization, and downlink-only quantization showed that the proposed method can save significant amount of communication energy, while maintaining the same or even better learning performance than the existing schemes.

\section*{Appendix}
We first define some notations which will be used throughout the appendix. For each communication round $m=0,1,..., K-1$ and local iteration $t=0,1,...,\tau-1$, we denote
\begin{align}
&\mathbf{w}_{m+1}=\mathbf{w}_m+\frac{1}{r}\sum_{i\in{S_m}}Q_1(\mathbf{w}_{m,\tau}^i-Q_2(\mathbf{w}_m)),\nonumber\\
&\hat{\mathbf{w}}_{m+1}=\mathbf{w}_m+\frac{1}{n}\sum_{i\in{[n]}}Q_1(\mathbf{w}_{m,\tau}^i-Q_2(\mathbf{w}_m)),\nonumber\\
&\bar{\mathbf{w}}_{m,t}=\frac{1}{n}\sum_{i\in{[n]}}\mathbf{w}_{m,t}^i,
\label{add_notation}
\end{align}
where $\mathbf{w}_{m+1}$ is the updated global model based on $r$ selected clients, $\hat{\mathbf{w}}_{m+1}$ is the updated global model based on all $n$ clients, and $\bar{\mathbf{w}}_{m,t}$ is the averaged model of $n$ clients.\\
\textbf{Lemma 1.} Considering the sequence of model ${\mathbf{w}_{m+1},\hat{\mathbf{w}}_{m+1},\bar{\mathbf{w}}_{m,\tau}}$, if Assumptions 1 and 2 hold, then we have
\begin{align}
\mathbb{E}f(\mathbf{w}_{m+1})&\leq \mathbb{E}f(\bar{\mathbf{w}}_{m,\tau})+\frac{L}{2}\mathbb{E}\Vert \hat{\mathbf{w}}_{m+1}-\bar{\mathbf{w}}_{m,\tau}\Vert^2 +\frac{L}{2}\mathbb{E}\Vert \hat{\mathbf{w}}_{m+1}-\mathbf{w}_{m+1}\Vert^2 
\label{lemma1}
\end{align}
for any period $m=0,...,K-1$. The proof is provided in Section \ref{proof of lemma 1}, and is a commonly used lemma for $L$-smooth functions.
In the following three lemmas, we derive the upper bound for the three terms on the right-hand side (RHS) of (\ref{lemma1}).\\
\textbf{Lemma 2.} Let Assumptions 2 and 3 hold, and consider the sequence of updates with stepsize $\eta$. Then we have 
\begin{align}
\mathbb{E}f(\bar{\mathbf{w}}_{m,\tau}) &\leq \mathbb{E}f(\mathbf{w}_m)+\frac{Ld\cdot R^2_m(\mathbf{w}_m)}{2s_m^2}-\frac{\eta}{2}\sum_{t=0}^{\tau -1}\mathbb{E}\Vert \nabla f(\bar{\mathbf{w}}_{m,t})\Vert^2 \nonumber\\
&-\eta(\frac{1}{2n}-\frac{L\eta}{2n}-\frac{L^2\eta^2\tau(\tau-1)}{n})\sum_{t=0}^{\tau -1}\sum_{i \in [n]}\mathbb{E}\Vert \nabla f(\mathbf{w}_{m,t}^i)\Vert^2 \nonumber\\
&+\frac{L\tau\eta^2\sigma^2}{2n}+\frac{L^2\eta^3\sigma^2(n+1)\tau(\tau-1)}{2n},
\label{lemma2}
\end{align}
for every round $m=0,...,K-1$. The proof can be found in Section \ref{proof of lemma 2}. This lemma shows that by receiving the global model $\mathbf{w}_m$, each client will execute $\tau$ steps of local update. Because of local updating, the training loss can be further decreased based on the loss of the original global model $\mathbf{w}_m$. \\
\textbf{Lemma 3.} With Assumption 1, for sequences ${\hat{\mathbf{w}}_{m+1},\bar{\mathbf{w}}_{m,\tau}}$ defined in (\ref{add_notation}), we have 
\begin{align}
\mathbb{E}\Vert \hat{\mathbf{w}}_{m+1}-\bar{\mathbf{w}}_{m,\tau}\Vert^2 &\leq d(\frac{R(\mathbf{w}_m)}{s_m})^2 +\frac{1}{n^2}\sum_{i \in [n]}d(\frac{R(\Delta \mathbf{w}_m^i)}{s_m^i})^2.
\end{align}
The proof can be found in Section \ref{proof of lemma 3}.\\
\textbf{Lemma 4.} If Assumptions 1 and 3 hold, then for the sequence of averages ${\hat{\mathbf{w}}_{m+1}}$ defined in (\ref{add_notation}), we have 
\begin{align}
\mathbb{E}\Vert \hat{\mathbf{w}}_{m+1}-&\mathbf{w}_{m+1}\Vert^2 \leq \frac{4(1+q_s)}{r(n-1)}(1-\frac{r}{n})(n\sigma^2\tau\eta^2+\tau\eta^2\sum_{i\in [n]}\sum_{t=0}^{\tau-1}\Vert \nabla f(\mathbf{w}_{m,t}^i)\Vert^2).
\end{align}
The proof can be found in \cite{reisizadeh2020fedpaq}, Section 8.5. This lemma is to calculate the error by client selection.

By combining Lemmas 1$\sim$4, we can get the following recursive inequality on the expected function value on the updated model at the cloud server, i.e., ${\mathbf{w}_m:m=1,..., K}$:
\begin{align}
\mathbb{E}f(\mathbf{w}_{m+1}) &\leq \mathbb{E}f(\mathbf{w}_m)-\frac{\eta}{2}\sum_{t=0}^{\tau -1}\mathbb{E}\Vert \nabla f(\bar{\mathbf{w}}_{m,t})\Vert^2 +\frac{Ld\cdot R_m^2(\mathbf{w}_m)}{s_m^2} + \frac{Ld}{2n^2}\sum_{i \in [n]}(\frac{R(\Delta \mathbf{w}_m^i)}{s_m^i})^2\nonumber\\
&-\frac{\eta}{2n} \left\{1-(1+\frac{4(n-r)(1+q_s)\tau}{r(n-1)})L\eta-2\tau(\tau-1)L^2\eta^2\right\} \sum_{t=0}^{\tau -1}\sum_{i \in [n]}\mathbb{E}\Vert \nabla f(\mathbf{w}_{m,t}^i)\Vert^2 \nonumber\\
&+\frac{L^2(n+1)\tau(\tau-1)\eta^3\sigma^2}{2n}  +\frac{L\tau\eta^2\sigma^2}{2n} +\frac{2L(1+q_s)(n-r)\tau\eta^2\sigma^2}{r(n-1)}.
\label{complicated}
\end{align}
In this work, we focus on quantization error rather than random selection error, so we can select all the clients in each round, i.e., $r=n$, and (\ref{complicated}) can be simplified into
\begin{align}
\mathbb{E}f(\mathbf{w}_{m+1}) \leq& \mathbb{E}f(\mathbf{w}_m)-\frac{\eta}{2}\sum_{t=0}^{\tau -1}\mathbb{E}\Vert \nabla f(\bar{\mathbf{w}}_{m,t})\Vert^2 +\frac{Ld\cdot R_m^2(\mathbf{w}_m)}{s_m^2} + \frac{Ld}{2n^2}\sum_{i \in [n]}(\frac{R(\Delta \mathbf{w}_m^i)}{s_m^i})^2\nonumber\\
&-\frac{\eta}{2n} (1-L\eta-2\tau(\tau-1)L^2\eta^2)\sum_{t=0}^{\tau -1}\sum_{i \in [n]}\mathbb{E}\Vert \nabla f(\mathbf{w}_{m,t}^i)\Vert^2  \nonumber\\
&+\frac{L^2(n+1)\tau(\tau-1)\eta^3\sigma^2}{2n} +\frac{L\tau\eta^2\sigma^2}{2n}.
\label{simplify}
\end{align}
When $\eta$ is small such that
\begin{align}
1-L\eta -2\tau(\tau-1)L^2\eta^2 \geq 0,
\label{smalleta}
\end{align}
we have
\begin{align}
\mathbb{E}f(\mathbf{w}_{m+1}) \leq& \mathbb{E}f(\mathbf{w}_m)-\frac{\eta}{2}\sum_{t=0}^{\tau -1}\mathbb{E}\Vert \nabla f(\bar{\mathbf{w}}_{m,t})\Vert^2 +\frac{Ld\cdot R_m^2(\mathbf{w}_m)}{s_m^2} + \frac{Ld}{2n^2}\sum_{i \in [n]}(\frac{R(\Delta \mathbf{w}_m^i)}{s_m^i})^2\nonumber\\
&+\frac{L^2(n+1)\tau(\tau-1)\eta^3\sigma^2}{2n} +\frac{L\tau\eta^2\sigma^2}{2n}.
\label{tosum}
\end{align}
Summing (\ref{tosum}) over $m=0,...,K-1$ and rearranging the terms produces that
\begin{align}
\frac{\eta}{2}\sum_{m=0}^{K-1}\sum_{t=0}^{\tau -1}\mathbb{E}\Vert \nabla f(\bar{\mathbf{w}}_{m,t})\Vert^2 &\leq Ld\sum_{m=0}^{K-1}(\frac{R_m(\mathbf{w})}{s_m})^2 + \frac{Ld}{2n^2}\sum_{m=0}^{K-1}\sum_{i \in [n]}(\frac{R_m^i(\Delta \mathbf{w})}{s_m^i})^2 \nonumber\\
&+f(\mathbf{w}_0)-f^* +\frac{KL^2(n+1)\tau(\tau-1)\eta^3\sigma^2}{2n} +\frac{KL\tau\eta^2\sigma^2}{2n}. 
\label{summed}
\end{align}
Multiplying $\frac{2}{K\tau\eta}$ on the LHS and RHS of (\ref{summed}), we get
\begin{align}
\frac{1}{K\tau}\sum_{m=0}^{K-1}\sum_{t=0}^{\tau -1}\mathbb{E}\Vert \nabla f(\bar{\mathbf{w}}_{m,t})\Vert^2 &\leq   \frac{2Ld}{K\tau\eta}\sum_{m=0}^{K-1}(\frac{R_m(\mathbf{w})}{s_m})^2 + \frac{Ld}{n^2K\tau\eta}\sum_{m=0}^{K-1}\sum_{i\in [n]}(\frac{R_m^i(\Delta \mathbf{w})}{s_m^i})^2 \nonumber\\
&+\frac{2(f(\mathbf{w}_0)-f^*)}{K\tau\eta} +\frac{L\eta\sigma^2+ L^2\eta^2(n+1)(\tau-1)\sigma^2}{n}, 
\label{summed2}
\end{align}
which completes the proof of Theorem 1.

\subsection{Proof of Lemma 1}
\label{proof of lemma 1}
In the $m^{th}$ communication round, the selected clients form a subset $S_m$. Since $S_m$ is uniformly picked from all clients set [n], we have
\begin{align}
\mathbb{E}_{S_m}\mathbf{w}_{m+1}&=\mathbf{w}_m + \mathbb{E}_{S_m}\frac{1}{r}\sum_{i\in{S_m}}Q_1(\mathbf{w}_{m,\tau}^i-Q_2(\mathbf{w}_m)) \nonumber\\
&=\mathbf{w}_m+\sum_{S\subseteq[n] \atop |S|=r}Pr[S_k=S]\frac{1}{r}\sum_{i\in{S_m}}Q_1(\mathbf{w}_{m,\tau}^i-Q_2(\mathbf{w}_m)) \nonumber\\
&=\mathbf{w}_m+\frac{1}{\binom{n}{r}}\frac{1}{r}\binom{n-1}{r-1}\sum_{i\in [n]} Q_1(\mathbf{w}_{m,\tau}^i-Q_2(\mathbf{w}_m)) \nonumber\\
&=\mathbf{w}_m+\frac{1}{n}\sum_{i\in[n]}Q_1(\mathbf{w}_{m,\tau}^i-Q_2(\mathbf{w}_m)) \nonumber\\
&=\hat{\mathbf{w}}_{m+1}.
\label{29}
\end{align}

According to Assumption 1 that the random quantizer Q($\cdot$) is unbiased, we have
\begin{align}
\mathbb{E}_{Q}\hat{\mathbf{w}}_{m+1}&=\mathbf{w}_m+\frac{1}{n}\sum_{i\in[n]}\mathbb{E}_{Q_1}Q_1(\mathbf{w}_{m,\tau}^i-\mathbb{E}_{Q_2}Q_2(\mathbf{w}_m)) \nonumber\\
&=\mathbf{w}_m+\frac{1}{n}\sum_{i\in[n]}(\mathbf{w}_{m,\tau}^i-\mathbb{E}_{Q_2}Q_2(\mathbf{w}_m)) \nonumber\\
&=\mathbf{w}_m+\frac{1}{n}\sum_{i\in[n]}(\mathbf{w}_{m,\tau}^i-\mathbf{w}_m) \nonumber\\
&= \frac{1}{n}\sum_{i\in[n]}\mathbf{w}_{m,\tau}^i \nonumber\\
&=\bar{\mathbf{w}}_{m,\tau}.
\label{30}
\end{align}

Since loss function $f$ is $L$-smooth, for any variables $\mathbf{x,y}$, we have
\begin{align}
f(\mathbf{w}) \leq f(\mathbf{y}) + \left \langle\nabla f(\mathbf{y}), \mathbf{w}-\mathbf{y}\right \rangle +\frac{L}{2}||\mathbf{w}-\mathbf{y}||^2.
\label{91}
\end{align}
Then, we can have
\begin{align}
f(\mathbf{w}_{m+1})&=f(\mathbf{\hat{w}}_{m+1}+\mathbf{w}_{m+1}-\mathbf{\hat{w}}_{m+1})  \nonumber\\
&\leq f(\mathbf{\hat{w}}_{m+1}) + \left \langle\nabla f(\mathbf{\hat{w}}_{m+1}), \mathbf{w}_{m+1}-\mathbf{\hat{w}}_{m+1}\right \rangle +\frac{L}{2}||\mathbf{w}_{m+1}-\mathbf{\hat{w}}_{m+1}||^2.
\label{92}
\end{align}
We then take the expectation of both sides of (\ref{92}), and since $\mathbb{E}_{S_m}\mathbf{w}_{m+1}=\hat{\mathbf{w}}_{m+1}$ (See (\ref{29})), we have
\begin{align}
\mathbb{E}f(\mathbf{w}_{m+1}) \leq \mathbb{E}f(\mathbf{\hat{w}}_{m+1})+\frac{L}{2}\mathbb{E}||\mathbf{\hat{w}}_{m+1}-\mathbf{w}_{m+1}||^2. 
\label{93}
\end{align}
Similarly, since loss function $f$ is $L$-smooth, we can have
\begin{align}
f(\mathbf{\hat{w}}_{m+1})&=f(\mathbf{\bar{w}}_{m,\tau}+\mathbf{\hat{w}}_{m+1}-\mathbf{\bar{w}}_{m,\tau})  \nonumber\\
&\leq f(\mathbf{\bar{w}}_{m,\tau}) + \left \langle\nabla f(\mathbf{\bar{w}}_{m,\tau}), \mathbf{\hat{w}}_{m+1}-\mathbf{\bar{w}}_{m,\tau}\right \rangle +\frac{L}{2}||\mathbf{\hat{w}}_{m+1}-\mathbf{\bar{w}}_{m,\tau}||^2.
\label{92again}
\end{align}

We then take the expectation of both sides of (\ref{92again}) and since $\mathbb{E}_{Q}\hat{\mathbf{w}}_{m+1}=\bar{\mathbf{w}}_{m,\tau}$ (See (\ref{30})), we have
\begin{align}
\mathbb{E}f(\mathbf{\hat{w}}_{m+1}) \leq \mathbb{E}f(\mathbf{\bar{w}}_{m,\tau})+\frac{L}{2}\mathbb{E}||\mathbf{\hat{w}}_{m+1}-\mathbf{\bar{w}}_{m,\tau}||^2. 
\label{94}
\end{align}
Combining (\ref{93}) and (\ref{94}), we get
\begin{align}
\mathbb{E}f(\mathbf{w}_{m+1})&\leq \mathbb{E}f(\bar{\mathbf{w}}_{m,\tau})+\frac{L}{2}\mathbb{E}\Vert \hat{\mathbf{w}}_{m+1}-\bar{\mathbf{w}}_{m,\tau}\Vert^2 +\frac{L}{2}\mathbb{E}\Vert \hat{\mathbf{w}}_{m+1}-\mathbf{w}_{m+1}\Vert^2 ,
\label{lemma1get}
\end{align}
which concludes Lemma 1.

\subsection{Proof of Lemma 2}
\label{proof of lemma 2}
According to \cite{reisizadeh2020fedpaq} Section 8.3, for perfect downlink transmission, we have
\begin{align}
\mathbb{E}f(\bar{\mathbf{w}}_{m,\tau}) &\leq \mathbb{E}f(\mathbf{w}_m)-\frac{\eta}{2}\sum_{t=0}^{\tau -1}\mathbb{E}\Vert \nabla f(\bar{\mathbf{w}}_{m,t})\Vert^2 \nonumber\\
&-\eta(\frac{1}{2n}-\frac{L\eta}{2n}-\frac{L^2\eta^2\tau(\tau-1)}{n})\sum_{t=0}^{\tau -1}\sum_{i \in [n]}\mathbb{E}\Vert \nabla f(\mathbf{w}_{m,t}^i)\Vert^2 \nonumber\\
&+\frac{L\tau\eta^2\sigma^2}{2n}+\frac{L^2\eta^3\sigma^2(n+1)\tau(\tau-1)}{2n}.
\label{perfect downlink}
\end{align}
When the downlink model is quantized, (\ref{perfect downlink}) will become
\begin{align}
\mathbb{E}f(\bar{\mathbf{w}}_{m,\tau}) &\leq \mathbb{E}f(Q_2(\mathbf{w}_m))-\frac{\eta}{2}\sum_{t=0}^{\tau -1}\mathbb{E}\Vert \nabla f(\bar{\mathbf{w}}_{m,t})\Vert^2 \nonumber\\
&-\eta(\frac{1}{2n}-\frac{L\eta}{2n}-\frac{L^2\eta^2\tau(\tau-1)}{n})\sum_{t=0}^{\tau -1}\sum_{i \in [n]}\mathbb{E}\Vert \nabla f(\mathbf{w}_{m,t}^i)\Vert^2 \nonumber\\
&+\frac{L\tau\eta^2\sigma^2}{2n}+\frac{L^2\eta^3\sigma^2(n+1)\tau(\tau-1)}{2n}.
\label{quantized downlink}
\end{align}
Since loss function $f$ is $L$-smooth, we have
\begin{align}
f(Q_2(\mathbf{w}_{m}))&=f(\mathbf{w}_{m}+Q_2(\mathbf{w}_{m})-\mathbf{w}_{m})  \nonumber\\
&\leq f(\mathbf{w}_{m}) + \left \langle\nabla f(\mathbf{w}_{m}), Q_2(\mathbf{w}_{m})-\mathbf{w}_{m}\right \rangle +\frac{L}{2}||Q_2(\mathbf{w}_{m})-\mathbf{w}_{m}||^2.
\label{quantization error}
\end{align}
Then we take the expectation of both sides of (\ref{quantization error}). Since $Q_2(\mathbf{w}_{m})$ is unbiased on $\mathbf{w}_{m}$ (according to Assumption 1), i.e., $\mathbb{E}Q_2(\mathbf{w}_{m})=\mathbb{E}\mathbf{w}_{m}$, we have
\begin{align}
\mathbb{E}f(Q_2(\mathbf{w}_{m}))\leq \mathbb{E}f(\mathbf{w}_{m}) +\frac{L}{2}\mathbb{E}||Q_2(\mathbf{w}_{m})-\mathbf{w}_{m}||^2 
\leq \mathbb{E}f(\mathbf{w}_{m}) + \frac{Ld\cdot R^2_m(\mathbf{w}_m)}{2s_m^2}.
\label{quantization error1}
\end{align}
Combining (\ref{quantized downlink}) and (\ref{quantization error1}), we get
\begin{align}
\mathbb{E}f(\bar{\mathbf{w}}_{m,\tau}) &\leq \mathbb{E}f(\mathbf{w}_m)+\frac{Ld\cdot R^2_m(\mathbf{w}_m)}{2s_m^2}-\frac{\eta}{2}\sum_{t=0}^{\tau -1}\mathbb{E}\Vert \nabla f(\bar{\mathbf{w}}_{m,t})\Vert^2 \nonumber\\
&-\eta(\frac{1}{2n}-\frac{L\eta}{2n}-\frac{L^2\eta^2\tau(\tau-1)}{n})\sum_{t=0}^{\tau -1}\sum_{i \in [n]}\mathbb{E}\Vert \nabla f(\mathbf{w}_{m,t}^i)\Vert^2 \nonumber\\
&+\frac{L\tau\eta^2\sigma^2}{2n}+\frac{L^2\eta^3\sigma^2(n+1)\tau(\tau-1)}{2n},
\label{lemma2get}
\end{align}
which concludes Lemma 2.

\subsection{Proof of Lemma 3}
\label{proof of lemma 3}
According to the definitions in (\ref{add_notation}), $\hat{\mathbf{w}}_{m+1}$ is the model with the quantization effect on both the uplink and downlink, and $\bar{\mathbf{w}}_{m,\tau}$ is the model without quantization. Here, we calculate the quantization error bound. Using Assumption 1, we have
\begin{align}
\mathbb{E}\Vert &\hat{\mathbf{w}}_{m+1}- \bar{\mathbf{w}}_{m,\tau} \Vert^2 = \mathbb{E}\Vert \mathbf{w}_m+\frac{1}{n}\sum_{i \in [n]}Q_1(\mathbf{w}_{m,\tau}^i-Q_2(\mathbf{w}_m))  - \frac{1}{n}\sum_{i \in [n]}\mathbf{w}_{m,\tau}^i \Vert^2 \nonumber\\
&= \mathbb{E}\Vert \mathbf{w}_m-Q_2(\mathbf{w}_m)+Q_2(\mathbf{w}_m) +\frac{1}{n}\sum_{i \in [n]}Q_1(\mathbf{w}_{m,\tau}^i-Q_2(\mathbf{w}_m))  - \frac{1}{n}\sum_{i \in [n]}\mathbf{w}_{m,\tau}^i \Vert^2 \nonumber\\
&= \mathbb{E}\Vert \mathbf{w}_m-Q_2(\mathbf{w}_m) +\frac{1}{n}\sum_{i \in [n]}Q_1(\mathbf{w}^i_{m,\tau}-Q_2(\mathbf{w}_m))-\frac{1}{n}\sum_{i \in [n]}(\mathbf{w}_{m,\tau}^i - Q_2(\mathbf{w}_m) ) \Vert^2 \nonumber\\
&\leq \mathbb{E}\Vert \mathbf{w}_m-Q_2(\mathbf{w}_m)\Vert^2 +\mathbb{E}\Vert \frac{1}{n}\sum_{i \in [n]}Q_1(\mathbf{w}^i_{m,\tau}-Q_2(\mathbf{w}_m))-\frac{1}{n}\sum_{i \in [n]}(\mathbf{w}_{m,\tau}^i - Q_2(\mathbf{w}_m) ) \Vert^2 \nonumber\\
&\leq \mathbb{E}\Vert \mathbf{w}_m-Q_2(\mathbf{w}_m)\Vert^2 +\frac{1}{n^2}\sum_{i \in [n]}\mathbb{E}\Vert Q_1(\mathbf{w}^i_{m,\tau}-Q_2(\mathbf{w}_m))-(\mathbf{w}_{m,\tau}^i - Q_2(\mathbf{w}_m) ) \Vert^2 \nonumber\\
&\leq d(\frac{R(\mathbf{w}_m)}{s_m})^2 +\frac{1}{n^2}\sum_{i \in [n]}d(\frac{R(\Delta \mathbf{w}_m^i)}{s_m^i})^2,
\end{align}
which concludes Lemma 3.
\bibliographystyle{IEEEbib}
\bibliography{strings,refs}

\begin{thebibliography}{10}

\bibitem{qu2022feddq}
L.~Qu, S.~Song, and C.-Y. Tsui,
\newblock ``{FedDQ: Communication-efficient federated learning with descending quantization},''
\newblock in {\em GLOBECOM 2022-2022 IEEE Global Communications Conference}. IEEE, 2022, pp. 281--286.

\bibitem{konevcny2015federated}
J.~Kone{\v{c}}n{\`y}, B.~McMahan, and D.~Ramage,
\newblock ``Federated optimization: Distributed optimization beyond the datacenter,''
\newblock {\em arXiv preprint arXiv:1511.03575}, 2015.

\bibitem{yang2019federated}
Q.~Yang, Y.~Liu, Y.~Cheng, Y.~Kang, T.~Chen, and H.~Yu,
\newblock ``Federated learning,''
\newblock {\em Synthesis Lectures on Artificial Intelligence and Machine Learning}, vol. 13, no. 3, pp. 1--207, 2019.

\bibitem{li2020federated}
T.~Li, A.~K. Sahu, A.~Talwalkar, and V.~Smith,
\newblock ``Federated learning: Challenges, methods, and future directions,''
\newblock {\em IEEE Signal Processing Magazine}, vol. 37, no. 3, pp. 50--60, 2020.

\bibitem{zhang2021fedpd}
X.~Zhang, M.~Hong, S.~Dhople, W.~Yin, and Y.~Liu,
\newblock ``{FedPD: A federated learning framework with adaptivity to non-iid data},''
\newblock {\em IEEE Transactions on Signal Processing}, vol. 69, pp. 6055--6070, 2021.

\bibitem{zhang2016parallel}
J.~Zhang, C.~De~Sa, I.~Mitliagkas, and C.~R{\'e},
\newblock ``{Parallel SGD: When does averaging help?},''
\newblock {\em arXiv preprint arXiv:1606.07365}, 2016.

\bibitem{konevcny2016federated}
J.~Kone{\v{c}}n{\`y}, H.~B. McMahan, F.~X. Yu, P.~Richt{\'a}rik, A.~T. Suresh, and D.~Bacon,
\newblock ``Federated learning: Strategies for improving communication efficiency,''
\newblock {\em arXiv preprint arXiv:1610.05492}, 2016.

\bibitem{cui2018mqgrad}
G.~Cui, J.~Xu, W.~Zeng, Y.~Lan, J.~Guo, and X.~Cheng,
\newblock ``{MQGrad: Reinforcement learning of gradient quantization in parameter server},''
\newblock in {\em Proceedings of the 2018 ACM SIGIR International Conference on Theory of Information Retrieval}, 2018, pp. 83--90.

\bibitem{kairouz2021advances}
P.~Kairouz, H.~B. McMahan, B.~Avent, A.~Bellet, M.~Bennis, A.~N. Bhagoji, K.~Bonawitz, Z.~Charles, G.~Cormode, R.~Cummings, et~al.,
\newblock ``Advances and open problems in federated learning,''
\newblock {\em Foundations and Trends{\textregistered} in Machine Learning}, vol. 14, no. 1--2, pp. 1--210, 2021.

\bibitem{mcmahan2017communication}
B.~McMahan, E.~Moore, D.~Ramage, S.~Hampson, and B.~A. y~Arcas,
\newblock ``Communication-efficient learning of deep networks from decentralized data,''
\newblock in {\em Artificial Intelligence and Statistics}. PMLR, 2017, pp. 1273--1282.

\bibitem{dean2012large}
J.~Dean, G.~Corrado, R.~Monga, K.~Chen, M.~Devin, M.~Mao, M.~Ranzato, A.~Senior, P.~Tucker, K.~Yang, et~al.,
\newblock ``Large scale distributed deep networks,''
\newblock {\em Advances in Neural Information Processing Systems}, vol. 25, 2012.

\bibitem{mills2019communication}
J.~Mills, J.~Hu, and G.~Min,
\newblock ``{Communication-efficient federated learning for wireless edge intelligence in IoT},''
\newblock {\em IEEE Internet of Things Journal}, vol. 7, no. 7, pp. 5986--5994, 2019.

\bibitem{guha2019one}
N.~Guha, A.~Talwalkar, and V.~Smith,
\newblock ``One-shot federated learning,''
\newblock {\em arXiv preprint arXiv:1902.11175}, 2019.

\bibitem{stich2018local}
S.~U. Stich,
\newblock ``{Local SGD converges fast and communicates little},''
\newblock {\em arXiv preprint arXiv:1805.09767}, 2018.

\bibitem{wang2021cooperative}
J.~Wang and G.~Joshi,
\newblock ``{Cooperative SGD: A unified framework for the design and analysis of local-update SGD algorithms},''
\newblock {\em Journal of Machine Learning Research}, vol. 22, 2021.

\bibitem{yu2019parallel}
H.~Yu, S.~Yang, and S.~Zhu,
\newblock ``{Parallel restarted SGD with faster convergence and less communication: Demystifying why model averaging works for deep learning},''
\newblock in {\em Proceedings of the AAAI Conference on Artificial Intelligence}, 2019, vol.~33, pp. 5693--5700.

\bibitem{haddadpour2019local}
F.~Haddadpour, M.~M. Kamani, M.~Mahdavi, and V.~Cadambe,
\newblock ``{Local SGD with periodic averaging: Tighter analysis and adaptive synchronization},''
\newblock {\em Advances in Neural Information Processing Systems}, vol. 32, 2019.

\bibitem{khaled2020tighter}
A.~Khaled, K.~Mishchenko, and P.~Richt{\'a}rik,
\newblock ``{Tighter theory for local SGD on identical and heterogeneous data},''
\newblock in {\em International Conference on Artificial Intelligence and Statistics}. PMLR, 2020, pp. 4519--4529.

\bibitem{wang2019adaptive}
J.~Wang and G.~Joshi,
\newblock ``{Adaptive communication strategies to achieve the best error-runtime trade-off in local-update SGD},''
\newblock {\em Proceedings of Machine Learning and Systems}, vol. 1, pp. 212--229, 2019.

\bibitem{reisizadeh2020fedpaq}
A.~Reisizadeh, A.~Mokhtari, H.~Hassani, A.~Jadbabaie, and R.~Pedarsani,
\newblock ``{FedPAQ: A communication-efficient federated learning method with periodic averaging and quantization},''
\newblock in {\em International Conference on Artificial Intelligence and Statistics}. PMLR, 2020, pp. 2021--2031.

\bibitem{aji2017sparse}
A.~F. Aji and K.~Heafield,
\newblock ``Sparse communication for distributed gradient descent,''
\newblock in {\em Proceedings of the 2017 Conference on Empirical Methods in Natural Language Processing}, 2017, pp. 440--445.

\bibitem{lindeep}
Y.~Lin, S.~Han, H.~Mao, Y.~Wang, and B.~Dally,
\newblock ``Deep gradient compression: Reducing the communication bandwidth for distributed training,''
\newblock in {\em International Conference on Learning Representations}.

\bibitem{wang2018atomo}
H.~Wang, S.~Sievert, S.~Liu, Z.~Charles, D.~Papailiopoulos, and S.~Wright,
\newblock ``Atomo: Communication-efficient learning via atomic sparsification,''
\newblock {\em Advances in Neural Information Processing Systems}, vol. 31, 2018.

\bibitem{ji2021dynamic}
S.~Ji, W.~Jiang, A.~Walid, and X.~Li,
\newblock ``Dynamic sampling and selective masking for communication-efficient federated learning,''
\newblock {\em IEEE Intelligent Systems}, vol. 37, no. 2, pp. 27--34, 2021.

\bibitem{yan2020dual}
Z.~Yan, D.~Xiao, M.~Chen, J.~Zhou, and W.~Wu,
\newblock ``Dual-way gradient sparsification for asynchronous distributed deep learning,''
\newblock in {\em 49th International Conference on Parallel Processing-ICPP}, 2020, pp. 1--10.

\bibitem{vogels2019powersgd}
T.~Vogels, S.~P. Karimireddy, and M.~Jaggi,
\newblock ``{PowerSGD: Practical low-rank gradient compression for distributed optimization},''
\newblock {\em Advances in Neural Information Processing Systems}, vol. 32, 2019.

\bibitem{li2021lotteryfl}
A.~Li, J.~Sun, B.~Wang, L.~Duan, S.~Li, Y.~Chen, and H.~Li,
\newblock ``{LotteryFL: empower edge intelligence with personalized and communication-efficient federated learning},''
\newblock in {\em 2021 IEEE/ACM Symposium on Edge Computing (SEC)}. IEEE, 2021, pp. 68--79.

\bibitem{lu2018multi}
Q.~Lu, W.~Liu, J.~Han, and J.~Guo,
\newblock ``Multi-stage gradient compression: Overcoming the communication bottleneck in distributed deep learning,''
\newblock in {\em International Conference on Neural Information Processing}. Springer, 2018, pp. 107--119.

\bibitem{phuong2020distributed}
T.~T. Phuong et~al.,
\newblock ``Distributed sgd with flexible gradient compression,''
\newblock {\em IEEE Access}, vol. 8, pp. 64707--64717, 2020.

\bibitem{tsuzuku2018variance}
Y.~Tsuzuku, H.~Imachi, and T.~Akiba,
\newblock ``Variance-based gradient compression for efficient distributed deep learning,''
\newblock {\em arXiv preprint arXiv:1802.06058}, 2018.

\bibitem{chen2020standard}
M.~Chen, Z.~Yan, J.~Ren, and W.~Wu,
\newblock ``Standard deviation based adaptive gradient compression for distributed deep learning,''
\newblock in {\em 2020 20th IEEE/ACM International Symposium on Cluster, Cloud and Internet Computing (CCGRID)}. IEEE, 2020, pp. 529--538.

\bibitem{xiao2021egc}
D.~Xiao, Y.~Mei, D.~Kuang, M.~Chen, B.~Guo, and W.~Wu,
\newblock ``{EGC: Entropy-based gradient compression for distributed deep learning},''
\newblock {\em Information Sciences}, vol. 548, pp. 118--134, 2021.

\bibitem{wangni2018gradient}
J.~Wangni, J.~Wang, J.~Liu, and T.~Zhang,
\newblock ``Gradient sparsification for communication-efficient distributed optimization,''
\newblock {\em Advances in Neural Information Processing Systems}, vol. 31, 2018.

\bibitem{alistarh2018convergence}
D.~Alistarh, T.~Hoefler, M.~Johansson, N.~Konstantinov, S.~Khirirat, and C.~Renggli,
\newblock ``The convergence of sparsified gradient methods,''
\newblock {\em Advances in Neural Information Processing Systems}, vol. 31, 2018.

\bibitem{stich2018sparsified}
S.~U. Stich, J.-B. Cordonnier, and M.~Jaggi,
\newblock ``{Sparsified SGD with memory},''
\newblock {\em Advances in Neural Information Processing Systems}, vol. 31, 2018.

\bibitem{shlezinger2020federated}
N.~Shlezinger, M.~Chen, Y.~C. Eldar, H.~V. Poor, and S.~Cui,
\newblock ``Federated learning with quantization constraints,''
\newblock in {\em ICASSP 2020-2020 IEEE International Conference on Acoustics, Speech and Signal Processing (ICASSP)}. IEEE, 2020, pp. 8851--8855.

\bibitem{alistarh2017qsgd}
D.~Alistarh, D.~Grubic, J.~Li, R.~Tomioka, and M.~Vojnovic,
\newblock ``{QSGD: Communication-efficient SGD via gradient quantization and encoding},''
\newblock {\em Advances in Neural Information Processing Systems}, vol. 30, 2017.

\bibitem{oland2015reducing}
A.~{\O}land and B.~Raj,
\newblock ``Reducing communication overhead in distributed learning by an order of magnitude (almost),''
\newblock in {\em 2015 IEEE International Conference on Acoustics, Speech and Signal Processing (ICASSP)}. IEEE, 2015, pp. 2219--2223.

\bibitem{wen2017terngrad}
W.~Wen, C.~Xu, F.~Yan, C.~Wu, Y.~Wang, Y.~Chen, and H.~Li,
\newblock ``Terngrad: Ternary gradients to reduce communication in distributed deep learning,''
\newblock {\em Advances in Neural Information Processing Systems}, vol. 30, 2017.

\bibitem{xu2020ternary}
J.~Xu, W.~Du, Y.~Jin, W.~He, and R.~Cheng,
\newblock ``Ternary compression for communication-efficient federated learning,''
\newblock {\em IEEE Transactions on Neural Networks and Learning Systems}, 2020.

\bibitem{seide20141}
F.~Seide, H.~Fu, J.~Droppo, G.~Li, and D.~Yu,
\newblock ``1-bit stochastic gradient descent and its application to data-parallel distributed training of speech dnns,''
\newblock in {\em Fifteenth Annual Conference of the International Speech Communication Association}, 2014.

\bibitem{strom2015scalable}
N.~Strom,
\newblock ``{Scalable distributed DNN training using commodity GPU cloud computing},''
\newblock in {\em Sixteenth Annual Conference of the International Speech Communication Association}, 2015.

\bibitem{bernstein2018signsgd}
J.~Bernstein, Y.-X. Wang, K.~Azizzadenesheli, and A.~Anandkumar,
\newblock ``{signSGD: Compressed optimisation for non-convex problems},''
\newblock in {\em International Conference on Machine Learning}. PMLR, 2018, pp. 560--569.

\bibitem{yu2018gradiveq}
M.~Yu, Z.~Lin, K.~Narra, S.~Li, Y.~Li, N.~S. Kim, A.~Schwing, M.~Annavaram, and S.~Avestimehr,
\newblock ``Gradiveq: Vector quantization for bandwidth-efficient gradient aggregation in distributed cnn training,''
\newblock {\em Advances in Neural Information Processing Systems}, vol. 31, 2018.

\bibitem{shlezinger2020uveqfed}
N.~Shlezinger, M.~Chen, Y.~C. Eldar, H.~V. Poor, and S.~Cui,
\newblock ``{UVeQFed: Universal vector quantization for federated learning},''
\newblock {\em IEEE Transactions on Signal Processing}, vol. 69, pp. 500--514, 2020.

\bibitem{gandikota2021vqsgd}
V.~Gandikota, D.~Kane, R.~K. Maity, and A.~Mazumdar,
\newblock ``{vqSGD: Vector quantized stochastic gradient descent},''
\newblock in {\em International Conference on Artificial Intelligence and Statistics}. PMLR, 2021, pp. 2197--2205.

\bibitem{cui2020clustergrad}
L.~Cui, X.~Su, Y.~Zhou, and L.~Zhang,
\newblock ``{ClusterGrad: Adaptive Gradient Compression by Clustering in Federated Learning},''
\newblock in {\em GLOBECOM 2020-2020 IEEE Global Communications Conference}. IEEE, 2020, pp. 1--7.

\bibitem{suresh2017distributed}
A.~T. Suresh, X.~Y. Felix, S.~Kumar, and H.~B. McMahan,
\newblock ``Distributed mean estimation with limited communication,''
\newblock in {\em International Conference on Machine Learning}. PMLR, 2017, pp. 3329--3337.

\bibitem{wu2018error}
J.~Wu, W.~Huang, J.~Huang, and T.~Zhang,
\newblock ``{Error compensated quantized SGD and its applications to large-scale distributed optimization},''
\newblock in {\em International Conference on Machine Learning}. PMLR, 2018, pp. 5325--5333.

\bibitem{karimireddy2019error}
S.~P. Karimireddy, Q.~Rebjock, S.~Stich, and M.~Jaggi,
\newblock ``{Error feedback fixes signSGD and other gradient compression schemes},''
\newblock in {\em International Conference on Machine Learning}. PMLR, 2019, pp. 3252--3261.

\bibitem{sun2019communication}
J.~Sun, T.~Chen, G.~Giannakis, and Z.~Yang,
\newblock ``Communication-efficient distributed learning via lazily aggregated quantized gradients,''
\newblock {\em Advances in Neural Information Processing Systems}, vol. 32, 2019.

\bibitem{tang2018communication}
H.~Tang, S.~Gan, C.~Zhang, T.~Zhang, and J.~Liu,
\newblock ``Communication compression for decentralized training,''
\newblock {\em Advances in Neural Information Processing Systems}, vol. 31, 2018.

\bibitem{jiang2018linear}
P.~Jiang and G.~Agrawal,
\newblock ``A linear speedup analysis of distributed deep learning with sparse and quantized communication,''
\newblock {\em Advances in Neural Information Processing Systems}, vol. 31, 2018.

\bibitem{guo2020accelerating}
J.~Guo, W.~Liu, W.~Wang, J.~Han, R.~Li, Y.~Lu, and S.~Hu,
\newblock ``Accelerating distributed deep learning by adaptive gradient quantization,''
\newblock in {\em ICASSP 2020-2020 IEEE International Conference on Acoustics, Speech and Signal Processing (ICASSP)}. IEEE, 2020, pp. 1603--1607.

\bibitem{jhunjhunwala2021adaptive}
D.~Jhunjhunwala, A.~Gadhikar, G.~Joshi, and Y.~C. Eldar,
\newblock ``Adaptive quantization of model updates for communication-efficient federated learning,''
\newblock in {\em ICASSP 2021-2021 IEEE International Conference on Acoustics, Speech and Signal Processing (ICASSP)}. IEEE, 2021, pp. 3110--3114.

\bibitem{lecun1998mnist}
Y.~LeCun,
\newblock ``{The MNIST database of handwritten digits},''
\newblock {\em http://yann. lecun. com/exdb/mnist/}, 1998.

\bibitem{xiao2017fashion}
H.~Xiao, K.~Rasul, and R.~Vollgraf,
\newblock ``Fashion-mnist: a novel image dataset for benchmarking machine learning algorithms,''
\newblock {\em arXiv preprint arXiv:1708.07747}, 2017.

\bibitem{krizhevsky2010cifar}
A.~Krizhevsky, V.~Nair, and G.~Hinton,
\newblock ``Cifar-10 (canadian institute for advanced research),''
\newblock {\em URL http://www. cs. toronto. edu/kriz/cifar. html}, vol. 5, no. 4, pp. 1, 2010.

\bibitem{han2015deep}
S.~Han, H.~Mao, and W.~J. Dally,
\newblock ``Deep compression: Compressing deep neural networks with pruning, trained quantization and huffman coding,''
\newblock {\em arXiv preprint arXiv:1510.00149}, 2015.

\bibitem{he2016deep}
K.~He, X.~Zhang, S.~Ren, and J.~Sun,
\newblock ``Deep residual learning for image recognition,''
\newblock in {\em Proceedings of the IEEE Conference on Computer Vision and Pattern Recognition}, 2016, pp. 770--778.

\bibitem{haddadpour2021federated}
F.~Haddadpour, M.~M. Kamani, A.~Mokhtari, and M.~Mahdavi,
\newblock ``Federated learning with compression: Unified analysis and sharp guarantees,''
\newblock in {\em International Conference on Artificial Intelligence and Statistics}. PMLR, 2021, pp. 2350--2358.

\end{thebibliography}

\end{document}